\documentclass{article}

\usepackage{microtype}
\usepackage{graphicx}
\usepackage{subfiles}
\usepackage{booktabs} 
\usepackage{mathtools}
\usepackage{amsfonts}
\usepackage{amsthm}
\usepackage{amsmath}
\usepackage{diagbox}
\usepackage{enumitem}
\usepackage{caption}
\usepackage{subcaption}
\usepackage{tikz}
\usetikzlibrary{positioning}

\usepackage{hyperref}

\usepackage[accepted]{icml2021}

\theoremstyle{definition}
\newtheorem{cdef}{Definition}

\providecommand\given{}
\DeclarePairedDelimiterX\AverX[1]{[}{]}{
\renewcommand\given{  \nonscript\:
  \delimsize\vert
  \nonscript\:
  \mathopen{}
  \allowbreak}
#1
}

\icmltitlerunning{Towards Open Ad Hoc Teamwork Using Graph-based Policy Learning}
\begin{document}

\twocolumn[
\icmltitle{Towards Open Ad Hoc Teamwork Using Graph-based Policy Learning}



\icmlsetsymbol{equal}{*}

\begin{icmlauthorlist}
\icmlauthor{Arrasy Rahman}{ed}
\icmlauthor{Niklas H\"opner}{uva}
\icmlauthor{Filippos Christianos}{ed}
\icmlauthor{Stefano V. Albrecht}{ed}
\end{icmlauthorlist}

\icmlaffiliation{ed}{School of Informatics, University of Edinburgh, Edinburgh, United Kingdom}
\icmlaffiliation{uva}{University of Amsterdam, Amsterdam, Netherlands}

\icmlcorrespondingauthor{Arrasy Rahman}{arrasy.rahman@ed.ac.uk}

\icmlkeywords{Machine Learning, ICML}

\vskip 0.3in
]



\printAffiliationsAndNotice{} 


\begin{abstract}
Ad hoc teamwork is the challenging problem of designing an autonomous agent which can adapt quickly to collaborate with teammates without prior coordination mechanisms, including joint training. Prior work in this area has focused on closed teams in which the number of agents is fixed. In this work, we consider \emph{open} teams by allowing agents with different fixed policies to enter and leave the environment without prior notification. Our solution builds on graph neural networks to learn agent models and joint-action value models under varying team compositions. We contribute a novel action-value computation that integrates the agent model and joint-action value model to produce action-value estimates. We empirically demonstrate that our approach successfully models the effects other agents have on the learner, leading to policies that robustly adapt to dynamic team compositions and significantly outperform several alternative methods.
\end{abstract}

\section{Introduction}
Many real-world problems require autonomous agents to perform tasks in the presence of other agents. Recent multi-agent reinforcement learning (MARL) approaches~\citep[e.g.][]{christianos2020shared, foerster2018a, DBLP:conf/icml/RashidSWFFW18, lowe2017multi} solve such problems by jointly training a set of agents with shared learning procedures. However, as agents become capable of long-term autonomy and are used for a growing number of tasks, it is possible that agents may have to interact with previously unknown other agents, without the opportunity for prior joint training. Thus, research in \emph{ad hoc teamwork}~\citep{stone2010ad} aims to design a single autonomous agent, which we refer to as the \textit{learner}, that can interact effectively with other agents without pre-coordination such as joint training.

Prior ad hoc teamwork approaches~\citep{AAAI15-Barrett, albrecht2016belief, barrett2017making, IJCAI19-ravula, chen2020aateam} achieved this aim by combining single-agent RL and agent modeling techniques to learn from direct interaction with other agents. These approaches were designed for \textit{closed teams} in which the number of agents is fixed. In practice, however, many tasks also require the learner to adapt to a changing number of agents in the environment. For example, consider an autonomous car that needs to drive differently depending on the number of nearby vehicles, which may be driven by humans or produced by different manufacturers, and their respective driving styles \citep{albrecht2020igp2}.

We make a step towards the full ad hoc teamwork challenge by considering \emph{open teams} in which agents with various fixed policies may enter and leave the team at any time and without prior notification. \textit{Open ad hoc teamwork} involves three main challenges that must be addressed without pre-coordination. First, the learner must quickly adapt its policy to the unknown policies of other agents. Second, handling openness requires the learner to adapt to changing team sizes in addition to other agents' types, which may affect the policy and role a learner must adopt within the team~\citep{tambe1997towards}. Third, the changing number of agents results in a state vector of variable length, which causes standard RL approaches that require fixed-length state vectors to perform poorly, as we show in our experiments. 

\setcounter{footnote}{2}
We propose a novel algorithm designed for open ad hoc teamwork, called \emph{Graph-based Policy Learning} (GPL) \footnote{Implementation code can be found at \url{https://github.com/uoe-agents/GPL}}, which addresses the aforementioned challenges. GPL adapts to dynamic teams by training a \emph{joint action value model} which allows the learner to disentangle the effect each agent's action has on the learner's returns. To select optimal actions from the joint action value model, we contribute a novel action-value computation method which integrates joint-action value estimates with action predictions learned using an agent model. To handle dynamic team sizes, the joint action value model and agent model are both based on graph neural network (GNN) architectures \citep{DBLP:conf/iclr/TacchettiSMZKRG19,DCG} which have proven useful for dealing with changing input sizes~\citep{hamilton2017inductive,jiang2019graph}. Our computed action values can be used within different value-based single-agent RL algorithms; in our experiments we test two versions of GPL, one based on Q-learning~\citep{mnih2015human} and one based on soft policy iteration~\citep{haarnoja2018soft}.

Our experiments evaluate GPL and various baselines in three multi-agent environments (Level-based foraging \citep{Albrecht2013GameTheoretic}, Wolfpack \citep{Leibo2017SocialDilemmas}, FortAttack \cite{deka2020natural}) for which we use different processes to specify when agents enter or leave the environment and their type assignments. We compare GPL against ablations of GPL that integrate agent models using input concatenation, a common approach used by prior works~\citep{grover2018learning, DBLP:conf/iclr/TacchettiSMZKRG19}; as well as two MARL approaches (MADDPG~\citep{lowe2017multi}, DGN \citep{jiang2019graph}). Our results show that both tested GPL variants achieve significantly higher returns than all other baselines in most learning tasks, and that GPL generalizes more effectively to previously unseen team sizes/compositions. We also provide a detailed analysis of learned concepts within GPL's joint action value models.

\section{Related Work}
\label{sec:relwork}

\textbf{Ad hoc teamwork:}
Ad hoc teamwork is at its core a single-agent learning problem in which the agent must learn a policy that is robust to different teammate types \citep{stone2010ad}. Early approaches focused on matrix games in which the teammate behavior was known \citep{agmon2012leading,stone2009leading}. A predominant approach in ad hoc teamwork is to compute Bayesian posteriors over defined teammate types and utilizing the posteriors in reinforcement learning methods (such as Monte Carlo Tree Search) to obtain optimal responses \cite{barrett2017making,albrecht2016belief}. Recent methods applied deep learning-based techniques to handle switching agent types \cite{IJCAI19-ravula} and to pretrain and select policies for different teammate types \cite{chen2020aateam}. All of these methods were designed for closed teams. In contrast, GPL is the first algorithm designed for open ad hoc teamwork in which agents of different types can dynamically enter and leave the team.

\textbf{Agent modeling:} An agent model takes a history of observations (e.g actions, states) as input and produces a prediction about the modeled agent, such as its goals or future actions \cite{albrecht2018autonomous}. Recent agent modeling frameworks explored in deep reinforcement learning \citep{DBLP:conf/icml/RaileanuDSF18,DBLP:conf/icml/RabinowitzPSZEB18,he2016opponent} are designed for closed environments. In contrast, we consider open multi-agent environments in which the number of active agents and their policies can vary in time. \citet{DBLP:conf/iclr/TacchettiSMZKRG19} proposed to use graph neural networks for modeling agent interactions in closed environments. Unlike GPL, their method uses predicted probabilities of future actions to augment the input into a policy network, which did not lead to higher final returns in their empirical evaluation. Our experiments also show that their method leads to worse generalization to teams with different sizes. 

\textbf{Multi-agent reinforcement learning (MARL):}
MARL algorithms use RL techniques to co-train a set of agents in a multi-agent system \citep{papoudakis2019dealing}. In contrast, ad hoc teamwork focuses on training a single agent to interact with a set of agents of unknown types that are in control over their own actions.
One approach in MARL is to learn factored action values to simplify the computation of optimal joint actions for agents, using agent-wise action values~\citep{DBLP:conf/icml/RashidSWFFW18,DBLP:conf/atal/SunehagLGCZJLSL18} and coordination graphs (CG)~\citep{DCG,zhou2019factorized}. Unlike these methods which use CGs to model joint action values for fully cooperative setups, we use CGs in ad hoc teamwork to model the impact of other agents' actions towards the learning agent's returns. \citet{jiang2019graph} consider MARL in open systems by utilizing GNN-based architectures as value networks. In our experiments we use a baseline following their method and show that it performs significantly worse than GPL.

\section{Problem Formulation}

The goal in open ad hoc teamwork is to train a learner agent to interact with other agents that have unknown behavioural models, called \emph{types}, and which may enter or leave the environment at any timestep. We formalize the problem of open ad hoc teamwork by extending the Stochastic Bayesian Game model~\citep{albrecht2016belief} to allow for openness. The current formulation focuses on fully observable environments and extending the framework to the partially observable case is left as future work.

\subsection{Open Stochastic Bayesian Games (OSBG)}

An OSBG is a tuple $(N,S,A,\Theta,R,P)$, where $N$,$S$,$A$,$\Theta$ represent the set of agents, the state space, the action space, and the type space, respectively. For simplicity we assume a common action space for all agents, but this can be generalised to individual action spaces for each agent. Let $\mathcal{P}$ denote the power set. To define joint actions under variable number of agents, we define a joint agent-action space $\boldsymbol{A_{N}}=\{a|a\in\mathcal{P}(N \times A), \forall{(i,a^{i}),(j,a^{j})} \in a: i=j \Rightarrow a^{i}=a^{j}\}$ and refer to the elements $a\in \boldsymbol{A_{N}}$  as \emph{joint agent-actions}. Similarly, we define a joint agent-type space $\boldsymbol{\Theta_{N}}=\{\theta|\theta\in\mathcal{P}(N \times \Theta), \forall{(i,\theta^{i}),(j,\theta^{j})}\in\theta: i=j\Rightarrow\theta^{i}=\theta^{j}\}$, refer to $\theta \in \boldsymbol{\Theta_{N}}$ as the \emph{joint agent-type} and use $\theta^{i}$ to denote the type of agent $i$ in $\theta$. The conditions in the definition of $\boldsymbol{\Theta_{N}}$ and $\boldsymbol{A_{N}}$ constrain each agent to only select one action while also being assigned to one type.

We assume the learner can observe the current state of the environment and the past actions of other agents but not their types. The learner's reward is determined by $R:S\times \boldsymbol{A_{N}} \mapsto \mathbb{R}$. The transition function $P: S \times \boldsymbol{A_{N}} \mapsto \Delta(S \times \boldsymbol{\Theta_{N}})$ determines the probability of the next state and joint agent-types, given the current state and joint agent-actions, where $\Delta(X)$ is the set of all probability distributions over X. Although the transition function allows agents to have changing types, our current work assumes that an agent's type is fixed between entering and leaving the environment.

Under an OSBG, the game starts by sampling an initial state, $s_{0}$, and an initial set of agents, $N_{0}$, with associated types, $\theta_0^i$ with $i\in N_0$, from a starting distribution $P_{0} \in \Delta(S \times \boldsymbol{\Theta_{N}})$. At state $s_{t}$, agents $N_{t} \subseteq N$ with types $\theta_t^i,i \in N_t$ exist in the environment and choose their actions $a^{i}_{t}$ by sampling from $\pi$. As a consequence of the selected joint agent-actions, the learner receives a reward computed through $R$. Finally, the next state $s_{t+1}$ and the next set of existing agents $N_{t+1}$ and types are sampled from $P$ given $s_t$ and joint agent-actions.

\subsection{Optimal Policy for OSBG}
\label{sec:OptPol}
Assuming that $i$ denotes the learning agent, the learning objective in an OSBG is to estimate the optimal policy defined below: 

\begin{cdef}
Let the joint actions and the joint policy of agents other than $i$ at time $t$ be denoted by $a^{-i}_{t}$ and $\boldsymbol{\pi}^{-i}_{t}$, respectively. Given a discount factor, $0\leq\gamma\leq1$, we define the action-value of policy $\pi^{i}$ , $\bar{Q}_{\pi^{i}}(s,a^{i})$, as :
\begin{equation}\mathbb{E}_{a^{i}_{t} \sim \pi^{i}, a^{-i}_{t}\sim\boldsymbol{\pi}^{-i}_{t}, P}\AverX[\bigg]{\sum_{t=0}^{\infty} \gamma^{t}R(s_{t}, a_{t})\given s_{0}=s, a^{i}_{0}=a^{i}},
\label{ActionReturns}
\end{equation}
which denotes the expected discounted return after the learner executes $a^{i}$ at $s$. A policy, $\pi^{i,*}$, is optimal if :
\begin{equation}
    \forall{\pi^{i}, s, a^{i}}, \bar{Q}_{\pi^{i,*}}(s,a^{i}) \geq \bar{Q}_{\pi^{i}}(s,a^{i}).
\end{equation}
Given $\bar{Q}_{\pi^{i,*}}(s,a)$, an OSBG is solved by always choosing $a^{i}$ with the highest action-value at $s$. 
\end{cdef}

\section{Graph-based Policy Learning}
We introduce the general components of GPL and their respective role for estimating an OSBG's optimal policy. We also describe the neural network architectures and learning procedures used for implementing each component. A general overview of GPL's architecture is provided in Figure~\ref{GPL_architecture} while the complete learning pseudocode is given in Appendix~\ref{GPLalgorithm}.

\begin{figure*}
    \centering
    \includegraphics[width=\textwidth]{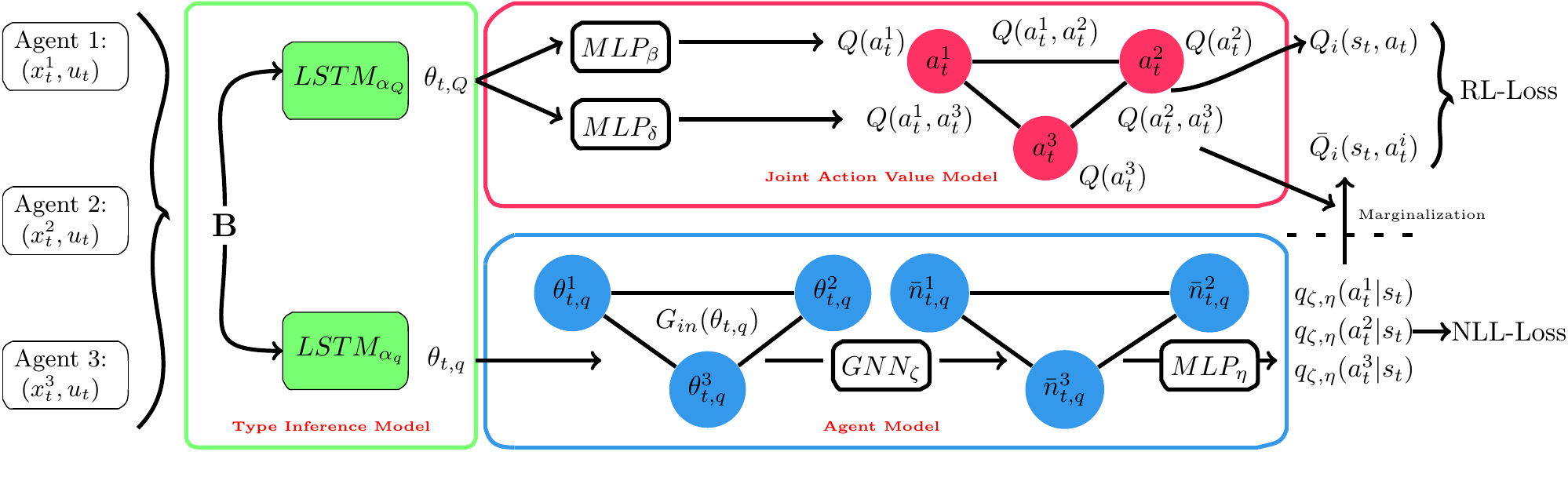}
    \caption{\textbf{Overview of GPL.} The joint action value model (red box) and agent model (blue box) receive type vectors produced by their own type embedding networks (green box), which are parameterized by $\alpha_{Q}$ and $\alpha_{q}$ respectively. These type vectors are processed by the joint-action value and agent model that are parameterized by $(\beta, \delta)$ and $(\mu,\nu)$ respectively. Output from joint action value model and agent model is finally combined using Equation~\ref{SingleActionValue} to compute the action-value function, using which the learner chooses its action.}
    \label{GPL_architecture}
\end{figure*}
\subsection{Method Overview and Motivation}
\label{GeneralFeatures}
In an OSBG, agents' joint actions inherently affect the learner's return through the rewards and next states it experiences. A learner using common value-based RL methods such as Q-Learning~\citep{watkins1992q} will always update the action-values of the learner's previous action, even if that action had minimal impact towards the observed reward from an OSBG. In MARL, a similar credit assignment problem is commonly addressed by using a centralized critic~\citep{lowe2017multi, foerster2018a} that disentangles the effects of other agents' actions to a learner's return by estimating a joint-action value function. Inspired by the importance of joint-action value modeling for credit assignment, GPL includes a component for \textit{joint-action value} estimation. The joint-action value of a policy, $Q_{\pi^{i}}(s, a)$, is defined as:
\begin{equation}\mathbb{E}_{a^{i}_{t} \sim \pi^{i}, a^{-i}_{t}\sim\boldsymbol{\pi}^{-i}, P}\AverX[\bigg]{\sum_{t=0}^{\infty} \gamma^{t}R(s_{t}, a_{t})\given s_{0}=s, a_{0}=a}.
\label{JointActionReturns}
\end{equation}
In contrast to Equation~\eqref{ActionReturns}, this joint-action value denotes the learner's expected return after the joint agent-action $a$ at $s$. Modeling joint-action values prevents the learner from only crediting its own action if it has minimal contribution towards rewards that it experienced.

Unlike centralized training for MARL~\citep{lowe2017multi} where the TD-error for training a joint-action value model can be approximated by using other agents' known policies, in ad hoc teamwork using joint-action value models introduces problems in computing optimal actions and temporal difference errors. This results from the assumption in ad hoc teamwork of not knowing other agents' policies during training and execution. Therefore, the uncertainty in other agents' actions must be accounted for in the action value computation in ad hoc teamwork. 

To perform joint-action value training and action selection for ad hoc teamwork, we approximate the learner's action value function $\bar{Q}_{\pi^{i}}$ by simultaneously learning a factorized joint action value model and an agent model that that approximates $\pi^{-i}_{t}(a^{-i}_{t}|s_{t},\theta^{-i})$ (see Section ~\ref{Sec:GPLArchitecture}). We can then approximate the expectation in Equation ~\ref{ActionReturns} by weighting the action value components with the likelihood of the corresponding agents' actions following :
\begin{equation}
    \bar{Q}_{\pi^{i}}(s_{t},a_{t}^{i}) = \mathbb{E}_{a_{t}^{-i}\sim\boldsymbol{\pi}_{t}^{-i}(.|s_{t},a_{t}^{i},\theta^{-i})}\AverX{Q(s_{t},a)|a^{i}=a_{t}^{i}}.
    \label{Marginalisation}
\end{equation}

Other strategies to deal with the uncertainty in other agents actions are possible. Being optimistic one can choose $a^{i}$ from the joint agent-action with maximum value, which can yield in suboptimal policies for ad hoc teamwork since other agents may choose actions that do not maximize the learners returns. Given an agent model that predicts agents next actions one could also sample other agents actions and compute an average joint action value for each of the learners actions. During learning of the action value model, one can even use information of joint actions taken at previous state to compute the TD-error following SARSA~\citep{rummery1994line}. However, these sampling-based approaches may increase the variance of the action value estimates, which may decrease learner's performance.

GPL's last component is the \textit{type inference} component. In an OSBG, types affect $Q(s,a)$ and $\pi^{-i}(a^{-i}|s,\theta^{-i})$ by determining other agent's immediate and future actions. Estimation of $Q(s,a)$ and $\pi^{-i}(a^{-i}|s,\theta^{-i})$ must therefore take agent types as input. However, agent types are unknown and must be inferred from agents' observed behavior, which we detail in Section 4.2.

The aforementioned GPL components are finally implemented using neural networks that facilitate an efficient computation of $\bar{Q}^{\pi^{*,i}}(s,a^{i})$ by imposing a simple factorization of the joint action value based on coordination graphs ~\citep{guestrin2002coordinated}. Furthermore, environment openness is handled by implementing the joint-action value and agent modeling components with GNNs, which we describe in the next section. This produces a flexible way of computing $\bar{Q}^{\pi^{*,i}}(s,a^{i})$ for any team sizes.

\subsection{The GPL Architecture}
\label{Sec:GPLArchitecture}
In this section, we outline the neural network architectures used for implementing GPL's three components (cf. Figure \ref{GPL_architecture}). This is followed by a description of the loss functions for training the neural networks representing each component.

\textbf{Type inference:} Without knowledge of the type space of an OSBG, GPL assumes types can be represented as fixed-length vectors. Since type inference requires reasoning over agent's behavior over an extended period of time, we use LSTMs~\citep{hochreiter1997long} for type inference. Our usage of LSTMs for type inference aligns with previous approaches for creating fixed-length embeddings of agents~\citep{grover2018learning, DBLP:conf/icml/RabinowitzPSZEB18}.

The LSTM takes the observation ($u_{t}$) and agent-specific information ($x^{i}_{t}$), which are both derived from $s_{t}$, to produce a hidden-state vector as an agent's type embedding. GPL then uses the type embeddings as input for the joint action value and agent modeling network. Although the joint action value and agent modeling feature may use the same type inference network, separate networks are used to prevent both model's gradients from interfering against each other during training. Further details on the preprocessing method to derive $u_{t}$ and $x^{i}_{t}$ from $s_{t}$, along with the computation of type vectors are provided in Appendix~\ref{GPLpreprocessing}. 

\textbf{Joint action value estimation:} We implement GPL's joint action value model as fully connected Coordination Graphs~\citep{guestrin2002coordinated} (CGs) for three reasons. First, CGs factorize the joint action value in a way that facilitates efficient action-value computation, which we will elaborate further when discussing GPL's action-value computation. Second, CGs can be implemented as GNNs~\citep{DCG}, which we demonstrate in Section~\ref{sec:exp_adhoc} to be important for handling environment openness. Third, the aforementioned action value factorization also enables CGs to model the contribution of other agents' actions towards the learner's returns, which we show in Section~\ref{GPLUtilAnalysis} to be the main reason behind GPL's superior performance compared to baselines. 

Given a state, a fully connected CG factorizes the learner's joint action value as a sum of individual utility terms, $Q^{j}_{\beta}(a^{j}_{t}|s_{t})$, and pairwise utility terms, $Q^{j,k}_{\delta}(a_{t}^{j},a_{t}^{k}|s_{t})$, under the following Equation:
\begin{equation}\label{JointActionValueComputation}
    Q_{\beta,\delta}(s_{t}, a_{t}) = \sum_{j\in{N_{t}}} Q^{j}_{\beta}(a^{j}_{t}|s_{t}) + \sum_{\substack{j,k\in{N_{t}}\\j\neq{k}}}Q^{j,k}_{\delta}(a_{t}^{j},a_{t}^{k}|s_{t}).
\end{equation}
$Q^{j}_{\beta}(a^{j}|s)$ intuitively represents $j$'s contribution towards the learner's returns by executing $a^{j}$, while $Q^{j,k}_{\delta}(a^{j},a^{k}|s)$ denotes $j$ and $k$'s contribution towards the learner's returns by jointly choosing $a^{j}$ and $a^{k}$. 

We implement $Q^{j}_{\beta}(a^{j}_{t}|s_{t})$ and $Q^{j,k}_{\delta}(a_{t}^{j},a_{t}^{k}|s_{t})$ as multilayer perceptrons (MLPs) parameterized by $\beta$ and $\delta$ to enable generalization across states. We then enable $\text{MLP}_{\beta}$ and $\text{MLP}_{\delta}$ to model the contribution of agents' individual and pairwise joint actions towards the learner's returns by providing type vectors of agents associated to each utility term, along with the the learner's type vector as input to $\text{MLP}_{\beta}$ and $\text{MLP}_{\delta}$. Given type vectors, $\theta^{i}_{t}$ and $\theta^{j}_{t}$, $\text{MLP}_{\beta}$ outputs a vector with a length of $|A|$ that estimates $Q^{j}_{\beta}(a^{j}|s_{t})$ for each possible actions of $j$ following:
\begin{equation}
    Q^{j}_{\beta}(a^{j}|s_{t}) = \text{MLP}_{\beta}(\theta^{j}_{t},\theta^{i}_{t})(a^{j}).
\end{equation}
On the other hand, instead of outputting the pairwise utility for the $|A|\times|A|$ possible pairwise actions of agent $j$ and $k$, $\text{MLP}_{\delta}$ outputs an $K\times{|A|}$ matrix ($K\ll{|A|}$) given its type vector inputs. Assuming a low-rank factorization of the pairwise utility terms, the output of $\text{MLP}_{\delta}$ is subsequently used to compute $Q^{j,k}_{\delta}(a_{t}^{j},a_{t}^{k}|s_{t})$ following: 
\begin{equation}
    Q^{j,k}_{\delta}(a_{t}^{j},a_{t}^{k}|s_{t}) =  (\text{MLP}_{\delta}(\theta^{j}_{t}, \theta^{i}_{t})^{\mathsf{T}}  \text{MLP}_{\delta}(\theta^{k}_{t},\theta^{i}_{t}))(a^{j}_{t},a^{k}_{t}).
\end{equation}
Previous work from~\citet{zhou2019factorized} demonstrated that low-rank factorization enables scalable pairwise utility computation even under thousands of possible pairwise actions. Finally note that $\text{MLP}_{\beta}$ and $\text{MLP}_{\delta}$ are shared between agents to encourage knowledge reuse for utility term computation, which importance to GPL's superior performance in open ad hoc teamwork is demonstrated in Section~\ref{GPLUtilAnalysis}.

\textbf{Agent modeling:} GPL's agent model assumes that other agents choose their actions independently, but models the effect other agents have on an agent's actions by using the Relational Forward Model (RFM) architecture~\citep{DBLP:conf/iclr/TacchettiSMZKRG19}. RFMs are a class of recurrent graph neural networks that have demonstrated high accuracy in predicting agents' next actions ~\citep{DBLP:conf/iclr/TacchettiSMZKRG19}. The RFM receives agent type embeddings, $\theta_{q}$, as its node input to compute a fixed-length embedding, $\bar{n}$, for each agent and is parameterized by $\zeta$. Let $a^{j}$ be the action taken by agent $j$ in the joint other agent-action $a^{-i}$, we use each agent's updated embedding to approximate $\pi^{-i}(a^{-i}|s,\theta_{-i})$ as:
\begin{align}\label{GNNequations}
\begin{split}
q_{\zeta,\eta}(a^{-i}|s) = \prod_{j\in{-i}} q_{\zeta,\eta,}(a^{j}|s), \\
q_{\zeta,\eta,}(a^{j}|s) = \textrm{Softmax}(\text{MLP}_{\eta}(\bar{n}_{j}))(a^{j}),
\end{split}
\end{align}with $\eta$ being the parameter of an MLP that transforms the updated agent embeddings. 

\textbf{Action value computation:} Evaluating Equation~\eqref{Marginalisation} can be inefficient in larger teams. For instance, a team of $k$ agents which may choose from $n$ possible actions requires the evaluation of $n^{k}$ joint-action terms, which number grows exponentially with the increase in team size. By contrast, a more efficient action-value computation arises from factorizing the joint action value network and using RFM-based agent modeling networks. By substituting the joint-action value and agent models from Equation~\eqref{JointActionValueComputation} and~\eqref{GNNequations} into  Equation~\eqref{Marginalisation}, we obtain Equation ~\eqref{SingleActionValue} as our action-value estimate, which we prove in Appendix~\ref{sec:GPLActValComputation}:
\begin{align}\label{SingleActionValue}
\begin{split}
\bar{Q}(s_{t}, a^{i}) & = Q^{i}_{\beta}(a^{i}|s_{t})  \\
& + \sum_{ \mathclap{\substack{a^{j}\in{A_{j}},  j\neq{i}}}}\big(Q^{j}_{\beta}(a^{j}|s_{t})+Q^{i,j}_{\delta}(a^{i}, a^{j}|s_{t})\big)q_{\zeta,\eta}(a^{j}|s_{t}) \\
&+ \sum_{\mathclap{\substack{a^{j}\in{A_{j}},a^{k}\in{A_{k}}, j, k\neq{i}}}}Q^{j,k}_{\delta}(a^{j}, a^{k}|s_{t})q_{\zeta,\eta}(a^{j}|s_{t})q_{\zeta,\eta}(a^{k}|s_{t}).
\end{split}
\end{align}Unlike Equation~\eqref{Marginalisation}, Equation~\eqref{SingleActionValue} is defined in terms of singular and pairwise action terms. In this case, the number of terms that need to be computed only increases quadratically as the team size increases. The computation of the required terms can be efficiently done in parallel with existing GNN libraries~\citep{wang2019dgl}.

\textbf{Model optimization:} As the learner interacts with teammates during learning, it stores a dataset of states, agents' actions, and rewards that it observed. Given a dataset of other agents' actions it collected at different states, $\{(s_{t}, a_{t})\}_{t=1}^{D}$, the agent modeling network is trained to estimate $\pi(a^{-i}_{t}|s_{t},a^{i}_{t})$ through supervised learning by minimizing the negative log likelihood loss defined below:
\begin{align}\label{ActionModelLoss}
\begin{split}
    L_{\zeta, \eta} &= -\text{log}(q_{\zeta,\eta}(a^{-i}_{t}|s_{t})).
\end{split}
\end{align}On the other hand, the collected dataset is also used to update GPL's joint-action value network using value-based reinforcement learning. Unlike standard value-based approaches ~\citep{mnih2015human}, we use the joint action value as the predicted value. The loss function for the joint action value network is then defined as:
\begin{equation}
    L_{\beta,\delta} = \dfrac{1}{2}\left(Q_{\beta,\delta}\left(s_{t}, a_{t}\right)-y\left(r_{t}, s_{t+1}\right)\right)^2,
    \label{ValueLoss}
\end{equation}
with $y(r_{t}, s_{t+1})$ being a target value which computation depends on the algorithm being used. We subsequently train GPL with Q-Learning (GPL-Q) ~\citep{watkins1992q} and Soft-Policy Iteration (GPL-SPI)~\citep{haarnoja2018soft}, which produces a greedy and stochastic policy respectively. The target value computations of both methods are defined as the following:
\begin{align*}\label{TargetValues}
    y_{\mathrm{QL}}\left(r_{t}, s_{t+1}\right) &= r_{t} + \gamma \text{max}_{a^{i}}\bar{Q}\left(s_{t+1},a^{i}\right),\\
    y_{\mathrm{SPI}}\left(r_{t}, s_{t+1}\right) &= r_{t} + \gamma \sum_{a^{i}}p_{\mathrm{SPI}}(a^{i}|s_{t+1})\bar{Q}\left(s_{t+1},a^{i}\right),
\end{align*}where GPL-SPI's policy uses the Boltzmann distribution,
\begin{equation}\label{Boltzmann}
    p_{\mathrm{SPI}}(a^{i}_{t}|s_{t}) \propto \mathrm{exp}\left(\dfrac{\bar{Q}(s_{t},a^{i})}{\tau}\right),
\end{equation}with $\tau$ being the temperature parameter. 

\section{Experimental Evaluation}
\label{ExperimentalResults}

In this section, we describe our open ad hoc teamwork experiments and demonstrate GPL's performance in them. This is followed by a detailed analysis of concepts learned by GPL's joint action value model. 

\subsection{Multi-Agent Environments} \label{sec:envs}

We conduct experiments in three fully observable multi-agent environments with different game complexity:

\textbf{Level-based foraging (LBF):} In LBF~\citep{Albrecht2013GameTheoretic}, agents and objects with levels $l\in\{1, 2, 3\}$ are spread in a $8\times8$ grid world. The agents' goal is to collect all objects. Agent actions include actions to move along the four cardinal directions, stay still, or to collect objects in adjacent grid locations. An object is collected if the sum of the levels of all agents involved in collecting at the same time is equal to or higher than the level of the object. Upon collecting an object, every agent that collects an object is given a reward equal to the level of the object. An episode finishes if all available objects are collected or after 50 timesteps.

\textbf{Wolfpack:} In Wolfpack ~\citep{Leibo2017SocialDilemmas}, a team of hunter agents must capture moving prey in a $10\times10$ grid world. Episodes consist of 200 timesteps and prey are trained to avoid capture using DQN \cite{mnih2015human}. While the agents have full observability of the environment, prey only observe a limited patch of grid cells ahead of them. Agents in this environment can move along the four cardinal directions or stay still at their current location. To capture a prey, at least two hunters must form a pack by where every pack members is located next to a prey's grid location. Every hunter in a pack that captured a prey is given a reward of two times the size of the capturing pack. However, we penalize agents by -0.5 for positioning themselves next to a prey without teammates positioned in other adjacent grids from the prey. Prey are respawned after they are captured.

\textbf{FortAttack:} FortAttack~\citep{deka2020natural} is situated on a two-dimensional plane where a team of attackers aim to reach a region, which we refer to as the fort, defended by defenders whose aim is to prevent any attackers from reaching the fort. Our learning agent assumes the role of a defender. Agents are equipped with actions to move along the four cardinal directions, rotate, and shoot any opposing team members located in a triangular shooting range defined by the agent's angular orientation and location. An episode ends when either an attacker reaches the fort, the learner is shot by attackers, or 200 timesteps have elapsed. The learner receives a reward of -3 for getting destroyed and 3 for destroying an attacker. On the other hand, a reward of -10 is given when an attacker reaches the fort and a reward of 10 when guards manage to defend the fort for 200 timesteps. A cost of -0.1 is also given for shooting. 

\subsection{Baselines}
\label{sec:BaselineDescription}
We design different learners, which can be categorized into single-agent value-based RL and MARL-based learners, to compare against GPL. Note that baselines that do not use GNNs require fixed-length inputs. To enable these approaches to handle changing number of agents in their observations, we impose an upper limit on the number of agents in the environment and preprocess the observation to ensure a fixed-length input by adding placeholder values. Details of this preprocessing method is provided in Appendix~\ref{sec:InputPreprocessingLSTM}.

\textbf{Single-agent RL baselines:} In line with GPL, all baselines are trained with synchronous Q-learning ~\citep{mnih2016asynchronous} and take as input the type vectors from the type inference network, but differ in their action-value computation and their use of an agent model. \textbf{QL} takes the concatenation of type vectors as input into a feedforward network to estimate action values. \textbf{GNN} applies multi-head attention ~\citep{jiang2019graph} to the type vectors and predicts action values based on the learner's node embedding. The agent model used by \textbf{QL-AM} and \textbf{GNN-AM} is identical in architecture and training procedure to GPL's agent model. However, the predicted action probabilities are concatenated to the individual agent representations $x_{t}$ as explored by prior methods ~\citep{DBLP:conf/iclr/TacchettiSMZKRG19}. An overview of baselines and their components can be found in Table \ref{t:basedef}. These baselines will allow us to investigate what advantage GNNs provide in training and generalization performance, when action probabilities help learning, which method of integrating action probabilities is most useful, and how GPL's approach to computing action values compares to prior methods.

\textbf{MARL baselines:} While in principle our ad hoc teamwork setting precludes joint training of agents via MARL (we only control a single learner agent and may also not know rewards of other agents), we use MARL approaches by assuming that (during training) we control all teammates and all teammates are using the same reward function as the learner. We compare with two MARL algorithms: MADDPG~\citep{lowe2017multi} and DGN~\cite{jiang2019graph}.

MADDPG is a MARL algorithm for closed environments, while DGN is a GNN-based MARL approach designed for joint training in open environments. For evaluation in open ad hoc teamwork, after MARL training completes, we select one of the jointly trained agents and measure its performance when interacting with the teammate types used in our ad hoc teamwork settings (see Sec. \ref{Sec:ExpSetting}).

\begin{table}[!htb]
    \small
    \center
    \caption{Types of ablations based on value network architecture and their use of agent modelling.}
    \label{t:basedef}
    \begin{tabular}{|c|c|c|c|}
        \hline
        Models & GNN  & Agent Model & Joint Action-Value \\
        \hline
        QL  & & & \\ \hline
        QL-AM & & \checkmark &  \\ \hline                 
        GNN & \checkmark & &  \\ \hline
        GNN-AM & \checkmark & \checkmark & \\ \hline
        GPL-Q & \checkmark & \checkmark & \checkmark \\ \hline
        GPL-SPI & \checkmark & \checkmark & \checkmark \\ 
        \hline
    \end{tabular}
\end{table}

\subsection{Data Collection for Training}
While type inference requires reasoning over consecutive observations, we want to avoid increasing the storage and computational resources required for training GPL and baselines. Therefore, we collect experiences across multiple environments in parallel instead of using a single environment and storing experiences in an experience replay. This resembles the data collection process of A3C~\citep{mnih2016asynchronous}. However, we collect experiences synchronously as opposed to A3C's asynchronous data collection process. 

\subsection{Experimental Setup}
\label{Sec:ExpSetting}
We construct environments for open ad hoc teamwork by creating a diverse set of teammate types for each environment. Agent types are designed such that a learner must adapt its policy to achieve optimal return when interacting with the type. In LBF and Wolfpack, each type's policy is implemented either via different heuristics or reinforcement learning-based policies. We vary the teammate policies in terms of their efficiency in executing a task and their roles in a team. Further details of the teammate policies and diversity analysis for Wolfpack and LBF are provided in Appendix~\ref{AdditionalPol}. We use pretrained policies provided by~\citet{deka2020natural} for FortAttack.

We simulate openness by creating an open process that determines how agents enter and leave during episodes, for both training and testing. In LBF and Wolfpack, the number of timesteps an agent can exist in the environment is determined by uniformly sampling from a certain range of integers. After staying for the predetermined number of timesteps, agents are removed from the environment. For FortAttack, agents are removed once they are shot by opponents. After being removed, agents can reenter the environment after a specific period of waiting time. The type of an agent entering an environment is uniformly sampled from all available types. Further details of the open process for each environments are provided in Appendix~\ref{sec:EnvOpenness}.

After every 160000 global training timesteps, GPL and baselines' are stored and evaluated based on their achieved return under the open process. To evaluate generalization capability in terms of number of agents, we impose different limits to the maximum team size resulting from the open process for training and testing. For testing, we increase the upper limit on team size to expose the learner against team configurations it has never encountered before. In all three environments, we specifically limit the team size to three agents during training time and increase this limit to five agents for testing. Since FortAttack has two opposing teams in its environment, these team size restriction is imposed on both teams.

\subsection{Open Ad Hoc Teamwork Results} \label{sec:exp_adhoc}
\begin{figure*}[t]
\begin{subfigure}{0.33\textwidth}
  \centering
  \includegraphics[scale=.33]{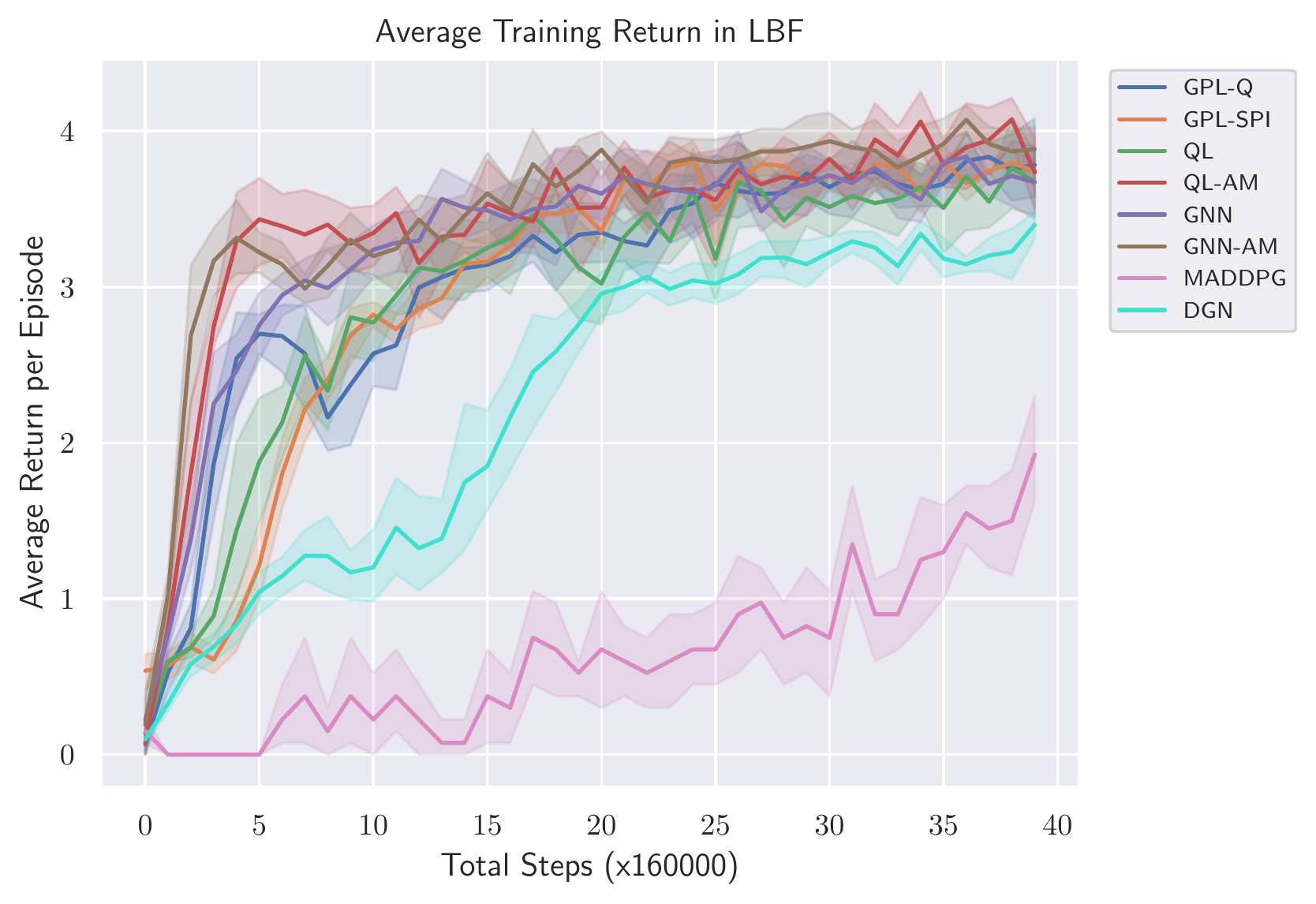}
  \caption{Results in Level-based foraging}
  \label{lbforagingopen}
\end{subfigure}
\begin{subfigure}{0.33\textwidth}
  \centering
  \includegraphics[scale=.33]{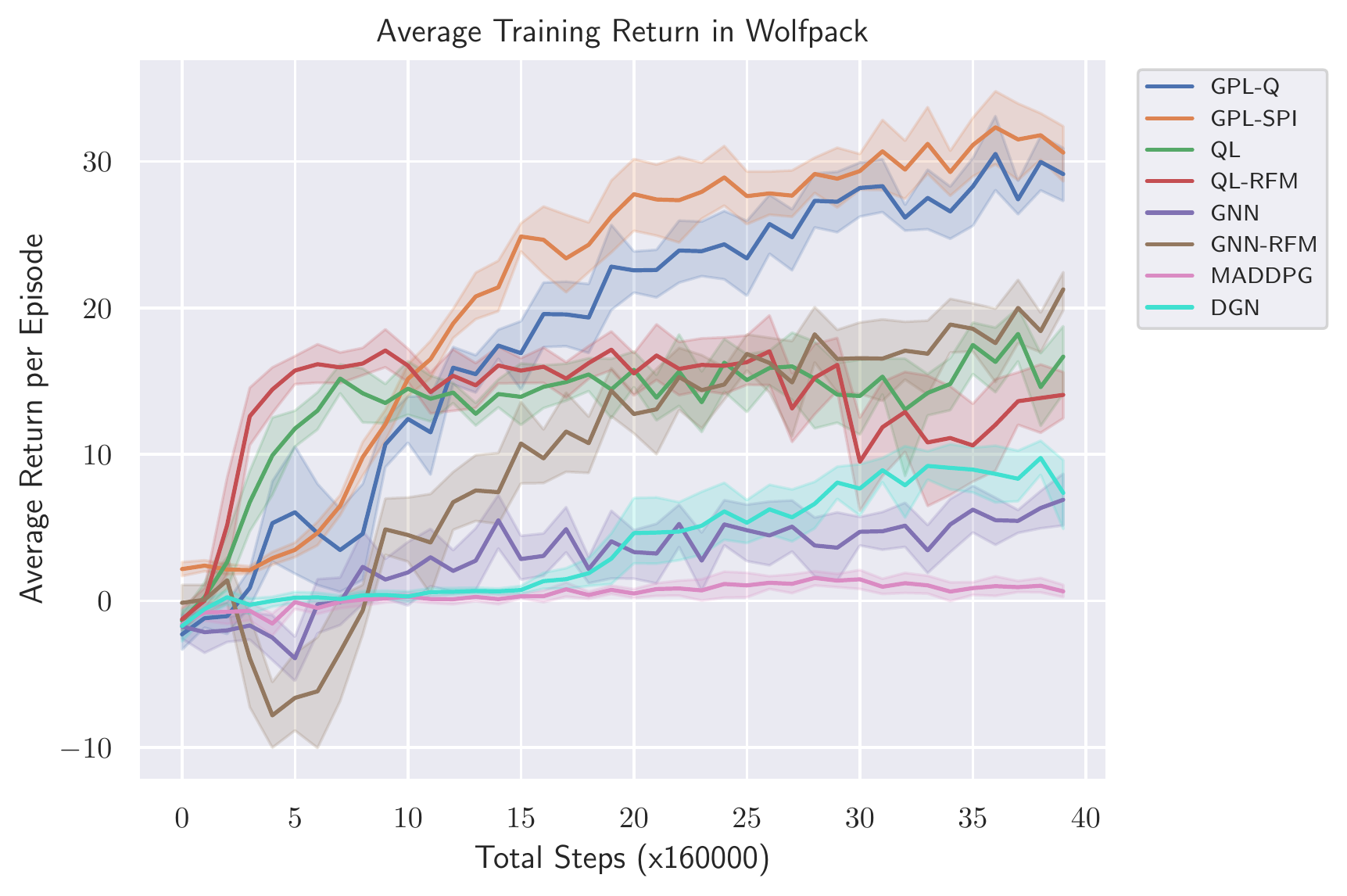}  
  \caption{Results in Wolfpack}
  \label{wolf05open}
\end{subfigure}
\begin{subfigure}{0.33\textwidth}
  \centering
  \includegraphics[scale=.33]{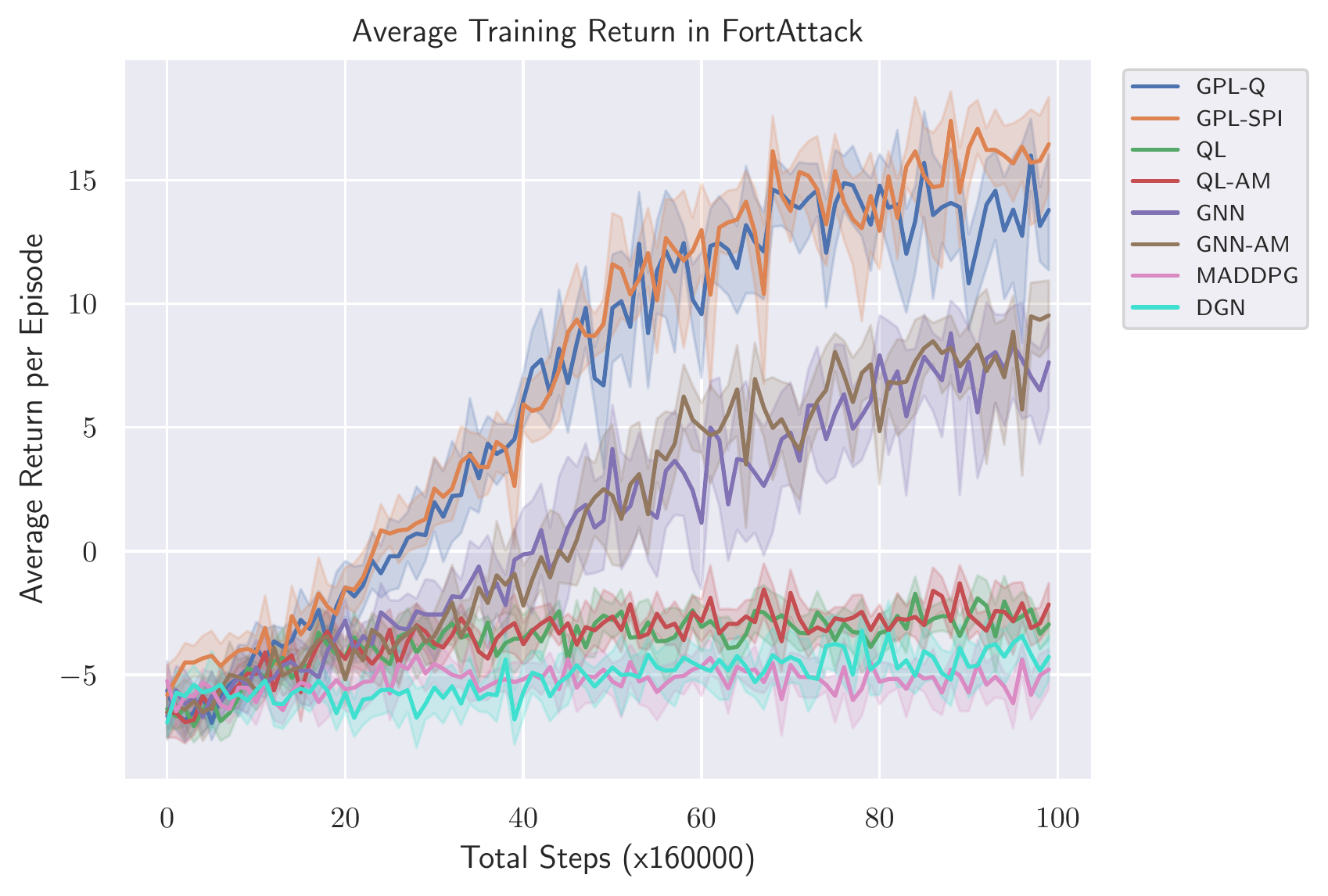}
  \caption{Results in FortAttack}
  \label{fortattackopen}
\end{subfigure}
\caption{\textbf{Open ad hoc teamwork results (training):} Average and 95\% confidence bounds of GPL \& baseline returns during training (up to 3 agents in a team for LBF, Wolfpack, and attacker \& defender teams in FortAttack). For each algorithm, training is done using eight different seeds and the resulting models are saved and evaluated every 160000 global steps.}
\label{OpenTrainingRes}
\end{figure*}

\begin{table*}[t]
    \centering
    \small
    \caption{\textbf{Open ad hoc teamwork results (testing):} Average and 95\% confidence bounds of GPL and baselines during testing (up to 5 agents in a team for LBF, Wolfpack, and attacker \& defender teams in FortAttack). For each algorithm, data was gathered by running the greedy policy resulting from the eight value networks stored at the checkpoint which achieved the highest average performance during training. The asterisk indicates significant difference in returns compared to the single-agent RL baselines.}
    \label{eval-table}
    \begin{tabular}{|l|c|c|c|c|c|c|c|c|}
    \hline 
    Env. & GPL-Q    & GPL-SPI                 & QL             & QL-AM         & GNN       & GNN-AM &DGN & MADDPG     \\ \hline
    LBF                   & \textbf{2.32$\pm$0.22}  & \textbf{2.40$\pm$0.16*} & 1.41$\pm$0.14  & 1.22$\pm$0.29  & 2.07$\pm$0.13  & 1.80$\pm$0.11 & 0.64 $\pm$ 0.9 & 0.91 $\pm$ 0.10\\ 
    Wolf. & \textbf{36.36$\pm$1.71*} & \textbf{37.61$\pm$1.69*} & 20.57$\pm$1.95 & 14.24$\pm$2.65 & 8.88$\pm$1.57 & 30.87$\pm$0.95 & 2.18 $\pm$ 0.66 & 19.20 $\pm$ 2.22 \\ 
    Fort.            & \textbf{14.20$\pm$2.42*} & \textbf{16.82$\pm$1.92*} & -3.51$\pm$0.60 & -3.51$\pm$1.51 & 7.01$\pm$1.63 & 8.12$\pm$0.74 & -5.98 $\pm$ 0.82 & -4.83 $\pm$ 1.24 \\ 
    \hline
    \end{tabular}
\end{table*}

Figure~\ref{OpenTrainingRes} shows the training performance of GPL-based approaches and the baselines. It shows that MARL-based approaches produce similar or worse performance than our worst performing single-agent RL baseline during training. While MARL policies performs better alongside other jointly trained agents, it generalizes poorly against the ad hoc teamwork teammates that cannot be jointly trained with MARL. For completeness, we show MARL learner's improved performance when interacting with other jointly trained agents in Appendix~\ref{sec:MARLJointTrainRes}.

Since results between GPL-Q and GPL-SPI are similar, we use GPL to refer to both in comparison with baselines. Figure~\ref{OpenTrainingRes} also shows that GPL significantly outperforms other baselines that use agent models, such as QL-AM and GNN-AM, in terms of training performance. Despite both being based on GNNs, GPL outperforming GNN-AM highlights GPL's action-value computation method over GNN-AM. As further indicated by the similarity in performance between QL/QL-AM or GNN/GNN-AM, concatenating action probabilities towards observations also does not improve training performance in most cases, which aligns with previous results from~\citet{grover2018learning} and~\citet{DBLP:conf/iclr/TacchettiSMZKRG19}. The reason GPL significantly outperforms others in training is because the joint-action model learns to disentangle the effects of other agents' actions. We further elaborate on GPL's superiority over other methods in terms of training performance through an analysis over its resulting joint action value estimates which we provide in Section~\ref{GPLUtilAnalysis}.

The generalization performance of GPL and the baselines is provided in Table~\ref{eval-table}. The way GPL, GNN and GNN-AM outperform single-agent RL baselines in generalization for LBF despite having similar training performances shows that GNNs are important components for generalizing between different open processes. Furthermore, GPL outperforming GNN-AM's generalization capability shows that using agent models for action-value computation using Equation~\eqref{SingleActionValue} also plays a role in improving generalization capability between open processes. In QL-AM and GNN-AM, the value networks must learn a model that integrates the predicted action probabilities to compute good action-value estimates, which may not generalize well to teams with previously unseen sizes. By contrast, GPL uses predicted action probabilities as weights in its action-value computation following Equation~\eqref{SingleActionValue}, which is proven in Appendix~\ref{sec:GPLActValComputation} to be correct for any team size. Finally, the low generalization performance of MARL-based baselines naturally follows from their low performance during training.

\subsection{Joint Action Value Analysis}
\label{GPLUtilAnalysis}
\begin{figure*}[!h]
\begin{subfigure}{0.44\textwidth}
  \centering
  \includegraphics[scale=.35]{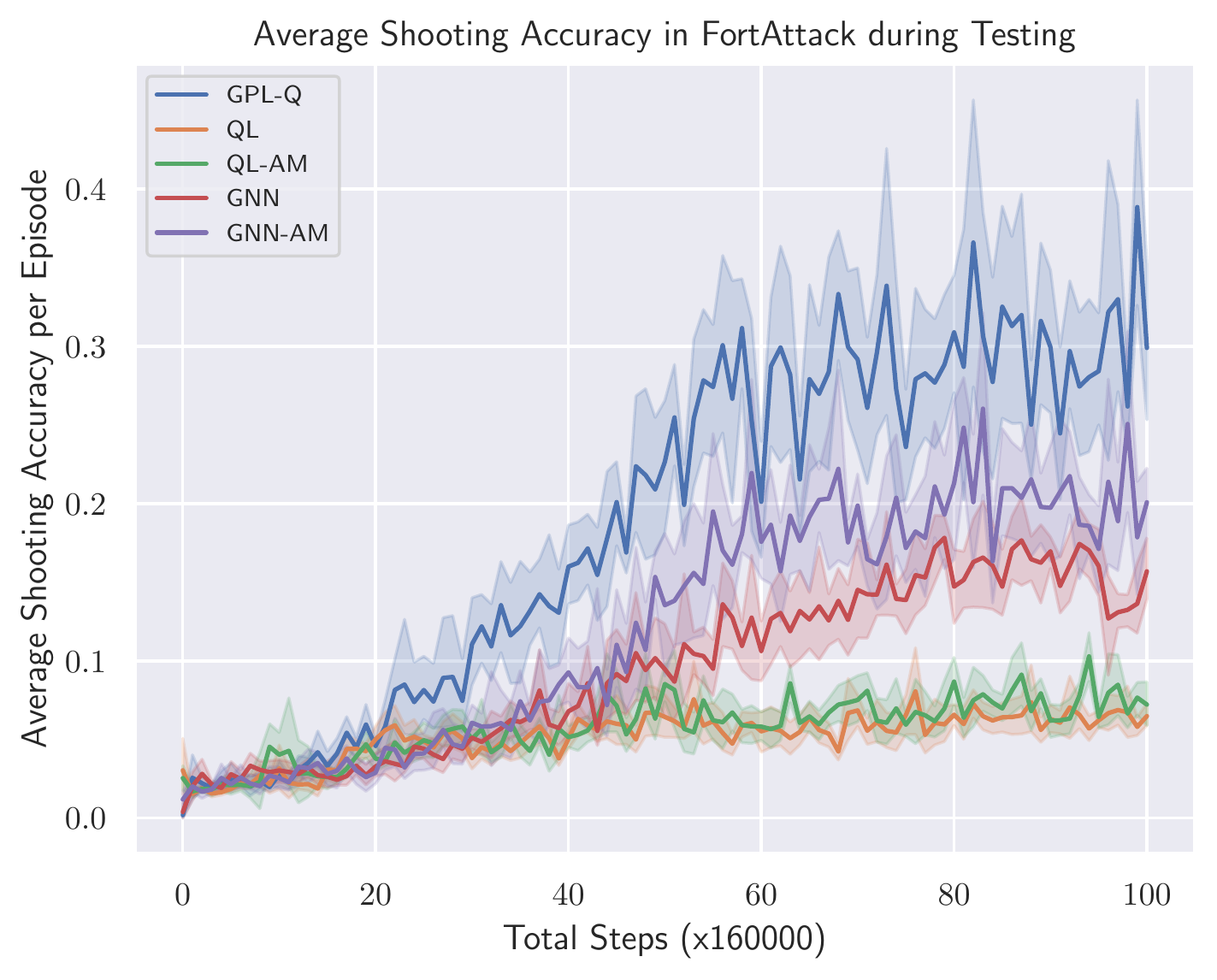}
  \caption{Shooting accuracy in FortAttack}
  \label{Fig:ShootingAccuracy}
 \end{subfigure}
\begin{subfigure}{0.55\textwidth}
\begin{tikzpicture}
  \pgfdeclareimage[height=3.85cm,width=4.7cm]{graph}{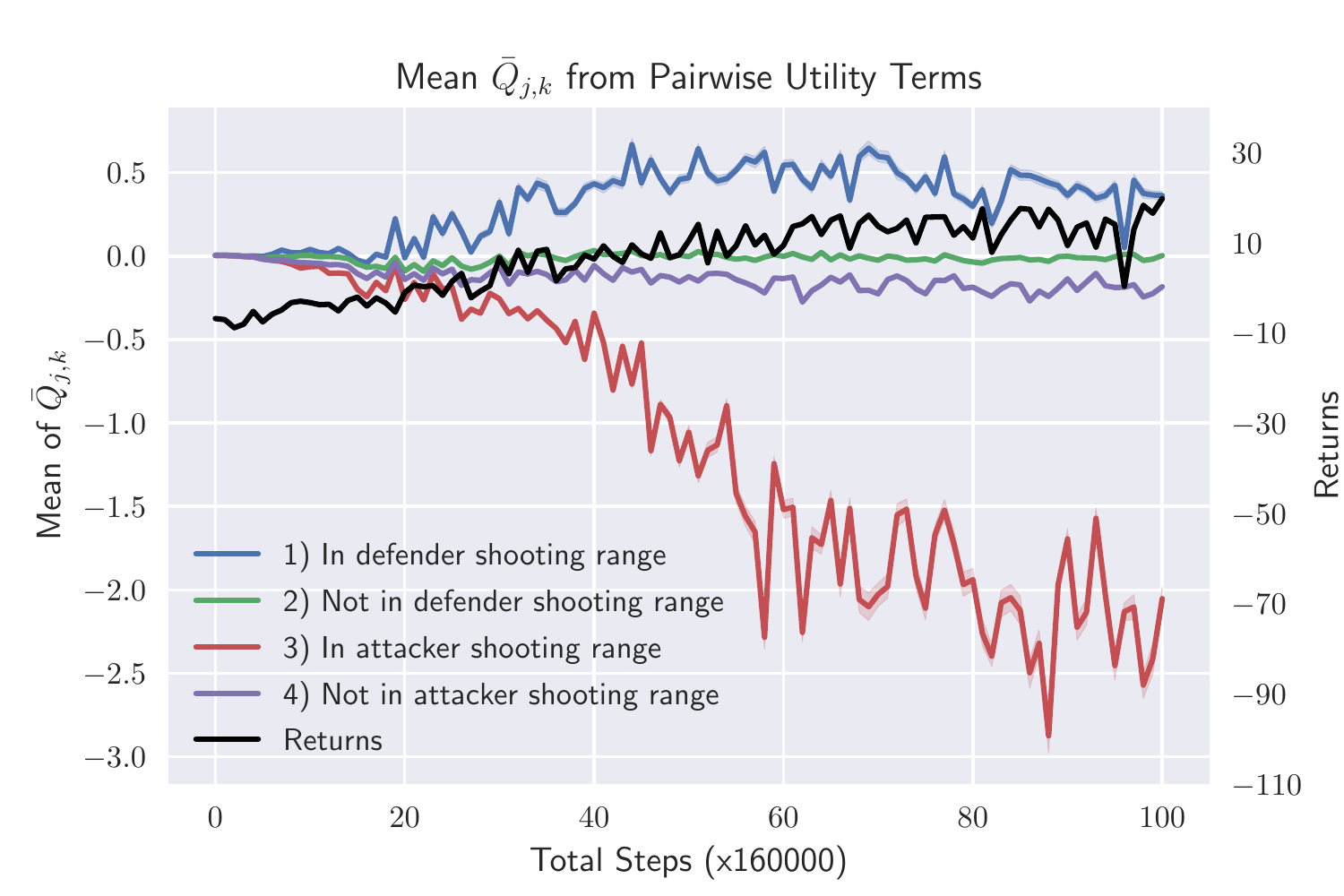};
  \pgfdeclareimage[height=3.85cm,width=2cm]{env1}{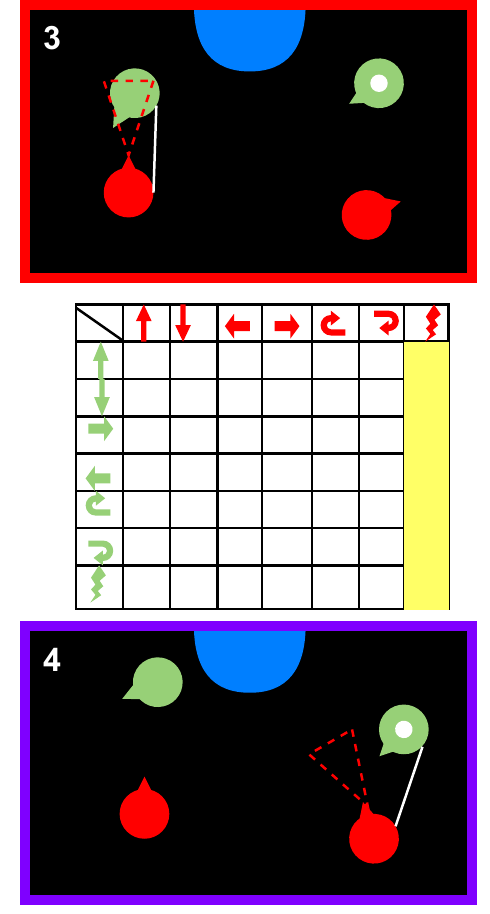};
  \pgfdeclareimage[height=3.85cm,width=2cm]{env2}{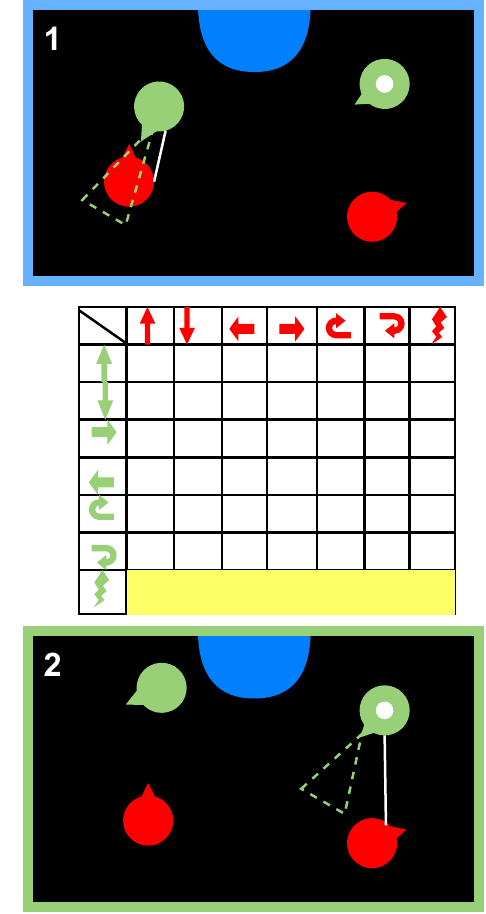};
  \node[] at (2.4,0) (graph){\pgfuseimage{graph}};
  \node[] at (0,0) (uppercircle) [left=0.cm]{\pgfuseimage{env1}};
  \node[] at (4.8,0) (rightsquare)[right=0.1cm]{\pgfuseimage{env2}};
\end{tikzpicture}
  \caption{Evolution of shooting metrics derived from GPL's pairwise utilites.}
  \label{Fig:JointUtilAnalysis}
\end{subfigure}
\caption{\textbf{Shooting-related metrics for FortAttack:} {\bf (a)} For GPL-Q and the baselines, we measure the percentage of times the learner successfully shot an attacker at each checkpoint during FortAttack training (cf. Section~\ref{Sec:ExpSetting}). {\bf (b)} We measure $\bar{Q}_{j,k}$, which is a metric derived from GPL-Q's pairwise utility terms that represents GPL-Q's estimate of the contribution towards the returns resulting from agent $j$ shooting an opponent agent, $k$. Each line in the plot corresponds to a different scenario in which the pairwise utility term is measured. Lines 1 and 2 represent $\bar{Q}_{j,k}$ when $j$ is a defender and $k$ is an attacker inside (1) or outside (2) $j$'s shooting range. Lines 3 and 4 contrast the value of $\bar{Q}_{j,k}$ when $j$ is an attacker and $k$ is a defender inside (3) or outside (4) $j$'s shooting range. To provide an example pairwise interaction where $\bar{Q}_{j,k}$ is computed from for each line, we visualize four sample pairwise interactions in FortAttack (white line in black boxes). Each black box is numbered after the line plot it corresponds to. The fort is represented by the blue half circle, attackers by red circles, defenders by green circles, the learner is marked with a white dot, and shooting ranges are indicated with dashed view cones. The matrices represent the joint action space for an attacker and defender, where the yellow marked fields refer to the actions that are averaged over to compute $\bar{Q}_{j,k}$ shown in the middle plot. This figure shows that the learner becomes increasingly aware of the benefits of shooting attackers inside a defender's shooting range and the negative consequences of a defender approaching an attacker's shooting range as training progresses, which causes GPL-Q's strong performance.}
\label{JointActionValueAnalysis}
\end{figure*}

We investigate how the joint action value model enables GPL-Q to significantly outperform the single-agent RL baselines during training in FortAttack, which is our most complex environment. For completeness, a similar analysis for Wolfpack is provided in Appendix~\ref{sec:WolfpackAVAnalysis}.

When comparing the resulting behavior from learning with GPL-Q and baselines, Figure~\ref{Fig:ShootingAccuracy} shows that GPL-Q's shooting accuracy improves at a faster rate than baselines and eventually converges at a higher value. Investigating the way GPL components encourage faster and better shooting performance may therefore highlight why GPL-based approaches outperform the baselines. We specifically investigate several shooting-related metrics derived from GPL's component, average their values over 480000 sample states gathered at different training checkpoints, and measure their correlation coefficient with GPL's average return. Among all metrics, the highest Pearson correlation coefficient of 0.85 is attained by $\bar{Q}_{j,k}$ when $j$ is a defender and $k$ is an attacker in $j$'s shooting range. $\bar{Q}_{j,k}$ is specifically defined as:
\begin{equation}
    \bar{Q}_{j,k} = \dfrac{\sum_{a^{k}}Q^{j,k}_{\delta}(a^{j}=\textrm{shoot}, a^{k}|s)}{|A^{k}|}.
    \label{utilmetrics}
\end{equation}
$\bar{Q}_{j,k}$ is derived from GPL's pairwise utility terms, $Q^{j,k}_{\delta}(a^{j}, a^{k}|s)$, and can be viewed as GPL's estimate of agent $j$'s average contribution towards the learner when $j$ decides to shoot $k$, averaged over all possible $a^k$. Therefore, this shows that GPL-Q's return strongly correlates with the pairwise utility terms assigned by the joint-action value model when a defender chooses to shoot an attacker inside its shooting range.

Rather than merely being strongly correlated to GPL's returns, we now elaborate why the pairwise utility terms produced by the  joint-action value model is the main reason behind GPL-based learner's higher final returns. Consider when $MLP_{\delta}$ increases its pairwise utility terms associated with shooting attackers inside a defender's shooting range. Since the learner itself is a defender, $MLP_{\delta}$ will also increase values of shooting-related pairwise utility terms when an attacker is inside the learner's shooting range. This encourages the learner to get attackers inside its shooting range and shoot more. Eventually, the learner achieves a higher return and $\bar{Q}_{j,k}$ becomes strongly correlated with the learner's return when $j$ is a defender and $k$ is an attacker inside $j$'s shooting range. Furthermore, Figure~\ref{Fig:JointUtilAnalysis} also shows that $MLP_{\delta}$ learns to associate negative values when defenders enter an attacker's shooting range, which enables the learner to learn to avoid the shooting range of attackers.

Unlike GPL, we show in Appendix~\ref{sec:BaselineActValAnalysis} that baselines that are not equipped with $MLP_{\delta}$ will not be able to learn these important concepts, which lead to their significantly worse performances. Specifically, learning to shoot with the single-agent RL baselines requires increasing the value network's estimate on shooting when an attacker ventures inside the learner's shooting range. In turn, agents can only increase the action value estimate of shooting after experiencing the positive rewards from successfully shooting an opposing team's agent. However, this can be especially difficult during exploration since it requires the learner to position itself at the right distance and orientation from an attacker. Even if the learner manages to get to the right distance from an attacker, a trained attacker can shoot a suboptimal learner instead if it does not orient itself properly or if it does not shoot the attacker first. 

\section{Conclusion and Future Work}

This work addresses the challenging problem of open ad hoc teamwork, in which the goal is to design an autonomous agent capable of robust teamwork under dynamically changing team composition without pre-coordination mechanisms such as joint training. Our proposed algorithm GPL uses coordination graphs to learn joint action-value functions that model the effects of other agents' actions towards the learning agent's returns, along with a GNN-based model trained to predict actions of other teammates. We empirically tested our approach in three multi-agent environments showing that our learned policies can robustly adapt to dynamically changing teams. We empirically show that GPL's success can be attributed to its ability to learn meaningful concepts to explain the effects of other agents' actions on the learning agent's returns. This enables GPL to produce action-values that lead to significantly better training and generalization performances than various baselines. 

An interesting direction for future work is related to extensions towards environments with partial observability and/or continuous action spaces. On the other hand, GPL's restrictive assumption that the learner's joint action value function must factorize following a fully connected CG can degrade the learner's returns for environments with certain reward functions~\cite{castellini2019representational}. Thus, automatically learning the most appropriate joint action value factorization through CG graph structure learning can potentially improve GPL's computational efficiency and return estimates.

\section*{Acknowledgement}
This research was financially supported in part by: the University of Edinburgh Enlightenment Scholarship (A.R.); the Hybrid Intelligence Center, funded by the Netherlands Organisation for Scientific Research under grant number 024.004.022 (N.H.); the UK EPSRC Centre for Doctoral Training in Robotics and Autonomous Systems (F.C.); and the US Office of Naval Research (ONR) grant N00014-20-1-2390 (S.A).

\bibliography{example_paper}

\begin{thebibliography}{41}
\providecommand{\natexlab}[1]{#1}
\providecommand{\url}[1]{\texttt{#1}}
\expandafter\ifx\csname urlstyle\endcsname\relax
  \providecommand{\doi}[1]{doi: #1}\else
  \providecommand{\doi}{doi: \begingroup \urlstyle{rm}\Url}\fi

\bibitem[Agmon \& Stone(2012)Agmon and Stone]{agmon2012leading}
Agmon, N. and Stone, P.
\newblock Leading ad hoc agents in joint action settings with multiple
  teammates.
\newblock In \emph{Proc. of 12th Int. Conf. on Autonomous Agents and Multiagent
  Systems (AAMAS 2012)}, June 2012.

\bibitem[Albrecht \& Ramamoorthy(2013)Albrecht and
  Ramamoorthy]{Albrecht2013GameTheoretic}
Albrecht, S.~V. and Ramamoorthy, S.
\newblock A game-theoretic model and best-response learning method for ad hoc
  coordination in multiagent systems.
\newblock In \emph{Proceedings of the 2013 International Conference on
  Autonomous Agents and Multi-Agent Systems}, pp.\  1155--1156, 2013.

\bibitem[Albrecht \& Stone(2017)Albrecht and Stone]{as2017}
Albrecht, S.~V. and Stone, P.
\newblock Reasoning about hypothetical agent behaviours and their parameters.
\newblock In \emph{Proceedings of the 16th International Conference on
  Autonomous Agents and Multiagent Systems}, pp.\  547--555, 2017.

\bibitem[Albrecht \& Stone(2018)Albrecht and Stone]{albrecht2018autonomous}
Albrecht, S.~V. and Stone, P.
\newblock Autonomous agents modelling other agents: A comprehensive survey and
  open problems.
\newblock \emph{Artificial Intelligence}, 258:\penalty0 66--95, 2018.

\bibitem[Albrecht et~al.(2016)Albrecht, Crandall, and
  Ramamoorthy]{albrecht2016belief}
Albrecht, S.~V., Crandall, J.~W., and Ramamoorthy, S.
\newblock Belief and truth in hypothesised behaviours.
\newblock \emph{Artificial Intelligence}, 235:\penalty0 63--94, 2016.

\bibitem[Albrecht et~al.(2021)Albrecht, Brewitt, Wilhelm, Gyevnar, Eiras,
  Dobre, and Ramamoorthy]{albrecht2020igp2}
Albrecht, S.~V., Brewitt, C., Wilhelm, J., Gyevnar, B., Eiras, F., Dobre, M.,
  and Ramamoorthy, S.
\newblock Interpretable goal-based prediction and planning for autonomous
  driving.
\newblock In \emph{IEEE International Conference on Robotics and Automation
  (ICRA)}, 2021.

\bibitem[Barrett \& Stone(2015)Barrett and Stone]{AAAI15-Barrett}
Barrett, S. and Stone, P.
\newblock Cooperating with unknown teammates in complex domains: A robot soccer
  case study of ad hoc teamwork.
\newblock In \emph{Proceedings of the Twenty-Ninth AAAI Conference on
  Artificial Intelligence}, January 2015.

\bibitem[Barrett et~al.(2011)Barrett, Stone, and Kraus]{barrett2011empirical}
Barrett, S., Stone, P., and Kraus, S.
\newblock Empirical evaluation of ad hoc teamwork in the pursuit domain.
\newblock In \emph{Proc. of 11th Int. Conf. on Autonomous Agents and Multiagent
  Systems (AAMAS)}, May 2011.

\bibitem[Barrett et~al.(2017)Barrett, Rosenfeld, Kraus, and
  Stone]{barrett2017making}
Barrett, S., Rosenfeld, A., Kraus, S., and Stone, P.
\newblock Making friends on the fly: Cooperating with new teammates.
\newblock \emph{Artificial Intelligence}, 242:\penalty0 132--171, 2017.

\bibitem[B{\"o}hmer et~al.(2020)B{\"o}hmer, Kurin, and Whiteson]{DCG}
B{\"o}hmer, W., Kurin, V., and Whiteson, S.
\newblock Deep coordination graphs.
\newblock In \emph{International Conference on Machine Learning}, pp.\
  980--991. PMLR, 2020.

\bibitem[Castellini et~al.(2019)Castellini, Oliehoek, Savani, and
  Whiteson]{castellini2019representational}
Castellini, J., Oliehoek, F.~A., Savani, R., and Whiteson, S.
\newblock The representational capacity of action-value networks for
  multi-agent reinforcement learning.
\newblock In \emph{Proceedings of the 18th International Conference on
  Autonomous Agents and MultiAgent Systems}, pp.\  1862--1864, 2019.

\bibitem[Chen et~al.(2020)Chen, Andrejczuk, Cao, and Zhang]{chen2020aateam}
Chen, S., Andrejczuk, E., Cao, Z., and Zhang, J.
\newblock Aateam: Achieving the ad hoc teamwork by employing the attention
  mechanism.
\newblock In \emph{Proceedings of the AAAI Conference on Artificial
  Intelligence}, volume~34, pp.\  7095--7102, 2020.

\bibitem[Christianos et~al.(2020)Christianos, Schäfer, and
  Albrecht]{christianos2020shared}
Christianos, F., Schäfer, L., and Albrecht, S.~V.
\newblock Shared experience actor-critic for multi-agent reinforcement
  learning.
\newblock In \emph{34th Conference on Neural Information Processing Systems},
  2020.

\bibitem[Deka \& Sycara(2020)Deka and Sycara]{deka2020natural}
Deka, A. and Sycara, K.
\newblock Natural emergence of heterogeneous strategies in artificially
  intelligent competitive teams.
\newblock \emph{arXiv preprint arXiv:2007.03102}, 2020.

\bibitem[Foerster et~al.(2018)Foerster, Farquhar, Afouras, Nardelli, and
  Whiteson]{foerster2018a}
Foerster, J., Farquhar, G., Afouras, T., Nardelli, N., and Whiteson, S.
\newblock Counterfactual multi-agent policy gradients.
\newblock In \emph{Proceedings of the AAAI Conference on Artificial
  Intelligence}, volume~32, 2018.

\bibitem[Grover et~al.(2018)Grover, Al-Shedivat, Gupta, Burda, and
  Edwards]{grover2018learning}
Grover, A., Al-Shedivat, M., Gupta, J., Burda, Y., and Edwards, H.
\newblock Learning policy representations in multiagent systems.
\newblock In \emph{International Conference on Machine Learning}, pp.\
  1802--1811, 2018.

\bibitem[Guestrin et~al.(2002)Guestrin, Lagoudakis, and
  Parr]{guestrin2002coordinated}
Guestrin, C., Lagoudakis, M.~G., and Parr, R.
\newblock Coordinated reinforcement learning.
\newblock In \emph{Proceedings of the Nineteenth International Conference on
  Machine Learning}, pp.\  227--234, 2002.

\bibitem[Haarnoja et~al.(2018)Haarnoja, Zhou, Abbeel, and
  Levine]{haarnoja2018soft}
Haarnoja, T., Zhou, A., Abbeel, P., and Levine, S.
\newblock Soft actor-critic: Off-policy maximum entropy deep reinforcement
  learning with a stochastic actor.
\newblock In \emph{International Conference on Machine Learning}, pp.\
  1861--1870. PMLR, 2018.

\bibitem[Hamilton et~al.(2017)Hamilton, Ying, and
  Leskovec]{hamilton2017inductive}
Hamilton, W., Ying, Z., and Leskovec, J.
\newblock Inductive representation learning on large graphs.
\newblock In \emph{Advances in neural information processing systems}, pp.\
  1024--1034, 2017.

\bibitem[He et~al.(2016)He, Boyd-Graber, Kwok, and
  Daum{\'e}~III]{he2016opponent}
He, H., Boyd-Graber, J., Kwok, K., and Daum{\'e}~III, H.
\newblock Opponent modeling in deep reinforcement learning.
\newblock In \emph{International Conference on Machine Learning}, pp.\
  1804--1813, 2016.

\bibitem[Hochreiter \& Schmidhuber(1997)Hochreiter and
  Schmidhuber]{hochreiter1997long}
Hochreiter, S. and Schmidhuber, J.
\newblock Long short-term memory.
\newblock \emph{Neural computation}, 9\penalty0 (8):\penalty0 1735--1780, 1997.

\bibitem[Jiang et~al.(2019)Jiang, Dun, Huang, and Lu]{jiang2019graph}
Jiang, J., Dun, C., Huang, T., and Lu, Z.
\newblock Graph convolutional reinforcement learning.
\newblock In \emph{International Conference on Learning Representations}, 2019.

\bibitem[Leibo et~al.(2017)Leibo, Zambaldi, Lanctot, Marecki, and
  Graepel]{Leibo2017SocialDilemmas}
Leibo, J.~Z., Zambaldi, V., Lanctot, M., Marecki, J., and Graepel, T.
\newblock Multi-agent reinforcement learning in sequential social dilemmas.
\newblock In \emph{Proceedings of the 16th Conference on Autonomous Agents and
  MultiAgent Systems}, pp.\  464--473, 2017.

\bibitem[Lowe et~al.(2017)Lowe, Wu, Tamar, Harb, Abbeel, and
  Mordatch]{lowe2017multi}
Lowe, R., Wu, Y.~I., Tamar, A., Harb, J., Abbeel, O.~P., and Mordatch, I.
\newblock Multi-agent actor-critic for mixed cooperative-competitive
  environments.
\newblock In \emph{Advances in neural information processing systems}, pp.\
  6379--6390, 2017.

\bibitem[Mnih et~al.(2015)Mnih, Kavukcuoglu, Silver, Rusu, Veness, Bellemare,
  Graves, Riedmiller, Fidjeland, Ostrovski, et~al.]{mnih2015human}
Mnih, V., Kavukcuoglu, K., Silver, D., Rusu, A.~A., Veness, J., Bellemare,
  M.~G., Graves, A., Riedmiller, M., Fidjeland, A.~K., Ostrovski, G., et~al.
\newblock Human-level control through deep reinforcement learning.
\newblock \emph{Nature}, 518\penalty0 (7540):\penalty0 529--533, 2015.

\bibitem[Mnih et~al.(2016)Mnih, Badia, Mirza, Graves, Lillicrap, Harley,
  Silver, and Kavukcuoglu]{mnih2016asynchronous}
Mnih, V., Badia, A.~P., Mirza, M., Graves, A., Lillicrap, T., Harley, T.,
  Silver, D., and Kavukcuoglu, K.
\newblock Asynchronous methods for deep reinforcement learning.
\newblock In \emph{International conference on machine learning}, pp.\
  1928--1937, 2016.

\bibitem[Papoudakis et~al.(2019)Papoudakis, Christianos, Rahman, and
  Albrecht]{papoudakis2019dealing}
Papoudakis, G., Christianos, F., Rahman, A., and Albrecht, S.~V.
\newblock Dealing with non-stationarity in multi-agent deep reinforcement
  learning.
\newblock \emph{arXiv preprint arXiv:1906.04737}, 2019.

\bibitem[Rabinowitz et~al.(2018)Rabinowitz, Perbet, Song, Zhang, Eslami, and
  Botvinick]{DBLP:conf/icml/RabinowitzPSZEB18}
Rabinowitz, N.~C., Perbet, F., Song, H.~F., Zhang, C., Eslami, S. M.~A., and
  Botvinick, M.
\newblock Machine theory of mind.
\newblock In \emph{Proceedings of the 35th International Conference on Machine
  Learning}, pp.\  4215--4224, 2018.

\bibitem[Raileanu et~al.(2018)Raileanu, Denton, Szlam, and
  Fergus]{DBLP:conf/icml/RaileanuDSF18}
Raileanu, R., Denton, E., Szlam, A., and Fergus, R.
\newblock Modeling others using oneself in multi-agent reinforcement learning.
\newblock In \emph{Proceedings of the 35th International Conference on Machine
  Learning}, pp.\  4254--4263, 2018.

\bibitem[Rashid et~al.(2018)Rashid, Samvelyan, de~Witt, Farquhar, Foerster, and
  Whiteson]{DBLP:conf/icml/RashidSWFFW18}
Rashid, T., Samvelyan, M., de~Witt, C.~S., Farquhar, G., Foerster, J.~N., and
  Whiteson, S.
\newblock {QMIX:} monotonic value function factorisation for deep multi-agent
  reinforcement learning.
\newblock In \emph{Proceedings of the 35th International Conference on Machine
  Learning}, pp.\  4292--4301, 2018.

\bibitem[Ravula et~al.(2019)Ravula, Alkobi, and Stone]{IJCAI19-ravula}
Ravula, M., Alkobi, S., and Stone, P.
\newblock Ad hoc teamwork with behavior switching agents.
\newblock In \emph{International Joint Conference on Artificial Intelligence
  (IJCAI)}, August 2019.

\bibitem[Rummery \& Niranjan(1994)Rummery and Niranjan]{rummery1994line}
Rummery, G.~A. and Niranjan, M.
\newblock \emph{On-line Q-learning using connectionist systems}, volume~37.
\newblock University of Cambridge, Department of Engineering Cambridge, UK,
  1994.

\bibitem[Stone et~al.(2009)Stone, Kaminka, and Rosenschein]{stone2009leading}
Stone, P., Kaminka, G.~A., and Rosenschein, J.~S.
\newblock Leading a best-response teammate in an ad hoc team.
\newblock In \emph{Agent-mediated electronic commerce. Designing trading
  strategies and mechanisms for electronic markets}, pp.\  132--146. Springer,
  2009.

\bibitem[Stone et~al.(2010)Stone, Kaminka, Kraus, and Rosenschein]{stone2010ad}
Stone, P., Kaminka, G., Kraus, S., and Rosenschein, J.
\newblock Ad hoc autonomous agent teams: Collaboration without
  pre-coordination.
\newblock In \emph{Proceedings of the AAAI Conference on Artificial
  Intelligence}, volume~24, 2010.

\bibitem[Sunehag et~al.(2018)Sunehag, Lever, Gruslys, Czarnecki, Zambaldi,
  Jaderberg, Lanctot, Sonnerat, Leibo, Tuyls, and
  Graepel]{DBLP:conf/atal/SunehagLGCZJLSL18}
Sunehag, P., Lever, G., Gruslys, A., Czarnecki, W.~M., Zambaldi, V.~F.,
  Jaderberg, M., Lanctot, M., Sonnerat, N., Leibo, J.~Z., Tuyls, K., and
  Graepel, T.
\newblock Value-decomposition networks for cooperative multi-agent learning
  based on team reward.
\newblock In \emph{Proceedings of the 17th International Conference on
  Autonomous Agents and MultiAgent Systems}, pp.\  2085--2087, 2018.

\bibitem[Tacchetti et~al.(2019)Tacchetti, Song, Mediano, Zambaldi,
  Kram{\'{a}}r, Rabinowitz, Graepel, Botvinick, and
  Battaglia]{DBLP:conf/iclr/TacchettiSMZKRG19}
Tacchetti, A., Song, H.~F., Mediano, P. A.~M., Zambaldi, V.~F., Kram{\'{a}}r,
  J., Rabinowitz, N.~C., Graepel, T., Botvinick, M., and Battaglia, P.~W.
\newblock Relational forward models for multi-agent learning.
\newblock In \emph{7th International Conference on Learning Representations},
  2019.

\bibitem[Tambe(1997)]{tambe1997towards}
Tambe, M.
\newblock Towards flexible teamwork.
\newblock \emph{Journal of Artificial Intelligence Research}, 7:\penalty0
  83--124, 1997.

\bibitem[Tampuu et~al.(2017)Tampuu, Matiisen, Kodelja, Kuzovkin, Korjus, Aru,
  Aru, and Vicente]{tampuu2017multiagent}
Tampuu, A., Matiisen, T., Kodelja, D., Kuzovkin, I., Korjus, K., Aru, J., Aru,
  J., and Vicente, R.
\newblock Multiagent cooperation and competition with deep reinforcement
  learning.
\newblock \emph{PloS one}, 12\penalty0 (4), 2017.

\bibitem[Wang et~al.(2019)Wang, Yu, Zheng, Gan, Gai, Ye, Li, Zhou, Huang, Ma,
  Huang, Guo, Zhang, Lin, Zhao, Li, Smola, and Zhang]{wang2019dgl}
Wang, M., Yu, L., Zheng, D., Gan, Q., Gai, Y., Ye, Z., Li, M., Zhou, J., Huang,
  Q., Ma, C., Huang, Z., Guo, Q., Zhang, H., Lin, H., Zhao, J., Li, J., Smola,
  A.~J., and Zhang, Z.
\newblock Deep graph library: Towards efficient and scalable deep learning on
  graphs.
\newblock \emph{ICLR Workshop on Representation Learning on Graphs and
  Manifolds}, 2019.

\bibitem[Watkins \& Dayan(1992)Watkins and Dayan]{watkins1992q}
Watkins, C.~J. and Dayan, P.
\newblock Q-learning.
\newblock \emph{Machine learning}, 8\penalty0 (3-4):\penalty0 279--292, 1992.

\bibitem[Zhou et~al.(2019)Zhou, Chen, Wen, Yang, Su, Zhang, Zhang, and
  Wang]{zhou2019factorized}
Zhou, M., Chen, Y., Wen, Y., Yang, Y., Su, Y., Zhang, W., Zhang, D., and Wang,
  J.
\newblock Factorized q-learning for large-scale multi-agent systems.
\newblock In \emph{Proceedings of the First International Conference on
  Distributed Artificial Intelligence}, pp.\  1--7, 2019.

\end{thebibliography}
\bibliographystyle{icml2021}
\newpage
\appendix

\section{GPL action-value computation}
\label{sec:GPLActValComputation}
Here, we show for completeness that the expression for the learner's action value in Equation ~\ref{SingleActionValue} is exactly the expectation introduced in Equation ~\ref{Marginalisation} when other agents policy is approximated by the agent model from Equation ~\ref{GNNequations} and the joint action value factorizes as in Equation ~\ref{JointActionValueComputation}. The indices indicating the parameters of the action value function and the agent model are left out for brevity.

\begin{align}
\begin{split}
\label{SubbedModels}
  \bar{Q}(s_{t},a^{i}) 
  &=\mathbb{E}_{a^{-i}\sim\boldsymbol{q}(.|s_{t})}\AverX{Q(s_{t},a)|a^{i}=a_{t}^{i}} \\
  &= \sum_{a^{-i}\in{A^{-i}}} Q(s_{t}, a) q(a^{-i}|s_{t},a^{i}) \\
  &= \sum_{a^{-i}\in{A^{-i}}} (\sum_{ \mathclap{\substack{a^{j}\in{A_{j}}}}} Q^{j}(a^{j}|s_{t}) \\
  &+ \sum_{ \mathclap{\substack{a^{j}\in{A_{j}}, a^{k} \in{A_{k}}}}}Q^{j,k}(a^{j}, a^{k}|s_{t})) q(a^{-i}|s_{t},a^{i}) \\
  &= Q^{i}(a^{i}|s_{t})  \\
&+ \sum_{ \mathclap{\substack{a^{j}\in{A_{j}},  j\neq{i}}}}\big(Q^{j}(a^{j}|s_{t})+Q^{i,j}(a^{i}, a^{j}|s_{t})\big)q(a^{j}|s_{t}) \\
&+ \sum_{\mathclap{\substack{a^{j}\in{A_{j}},a^{k}\in{A_{k}}, j, k\neq{i}}}}Q^{j,k}(a^{j}, a^{k}|s_{t})q(a^{j}|s_{t})q(a^{k}|s_{t}),
\end{split}
\end{align}
where in the first step we used the definition of the expectation and in the second step we substituted in Equation ~\ref{JointActionValueComputation} for the joint action value model. In the third step we substituted in the agent model from Equation ~\ref{GNNequations} and marginalized out actions from the agent model $q$ that are not part of the corresponding action value factor. Note that Equation ~\ref{SubbedModels} is valid regardless of the number of teammates in the environment.

\section{Environment details}
\label{env_details}

We provide additional details to reproduce the environments used in the experiments. Source code and installation instructions for these environments are included as part of the supplementary materials.
\begin{figure*}[h]
\centering
\begin{subfigure}{.30\textwidth}
  \centering
  \includegraphics[trim={2.5cm 13.5cm 4.5cm 2.5cm},clip,scale=.235]{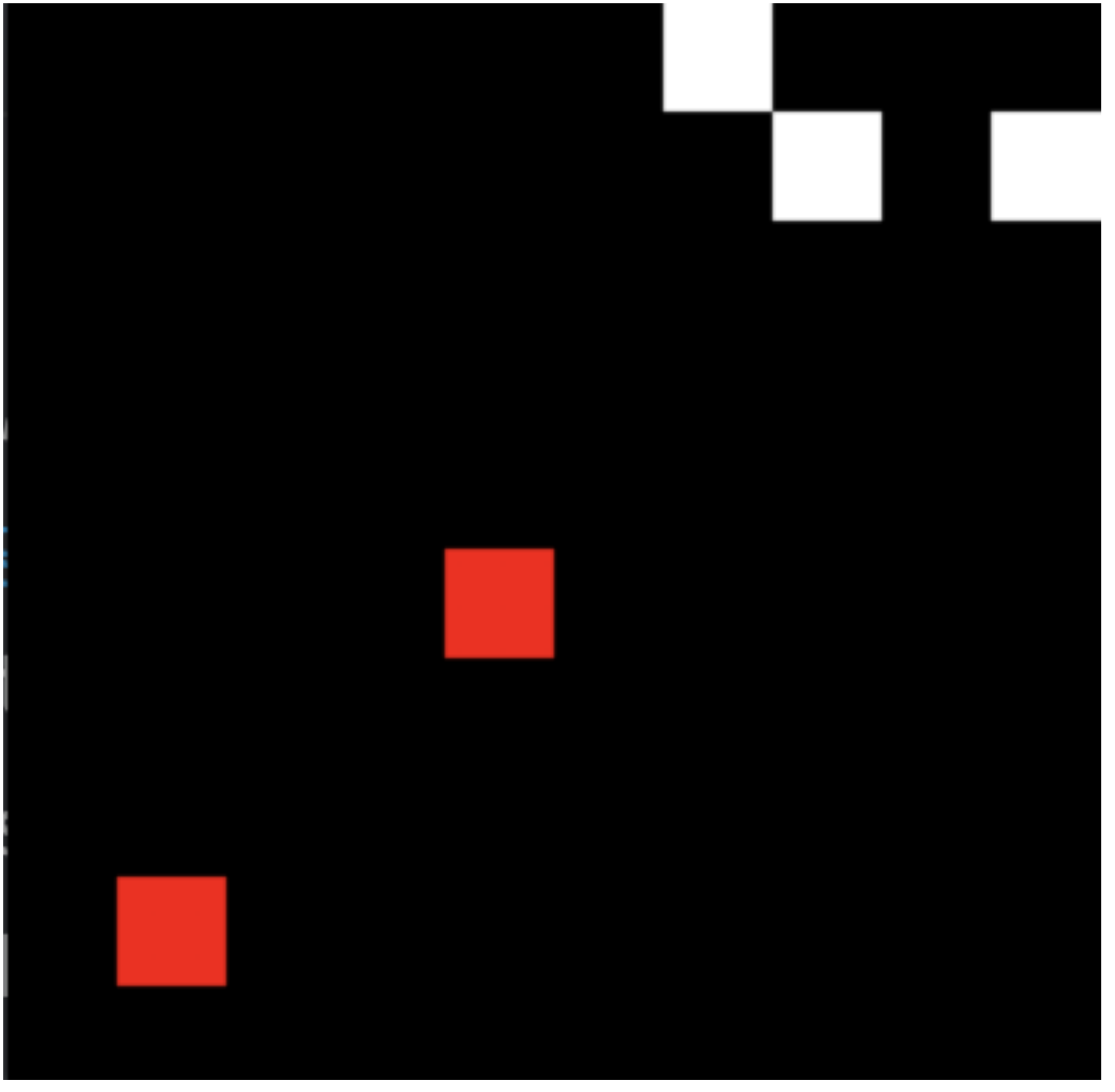}
  \caption{Wolfpack}
  \label{wolfpackscreen}
\end{subfigure}%
\hfill
\begin{subfigure}{.30\textwidth}
  \centering
  \includegraphics[trim={0cm 12.5cm 0cm 2.cm},clip,scale=.22]{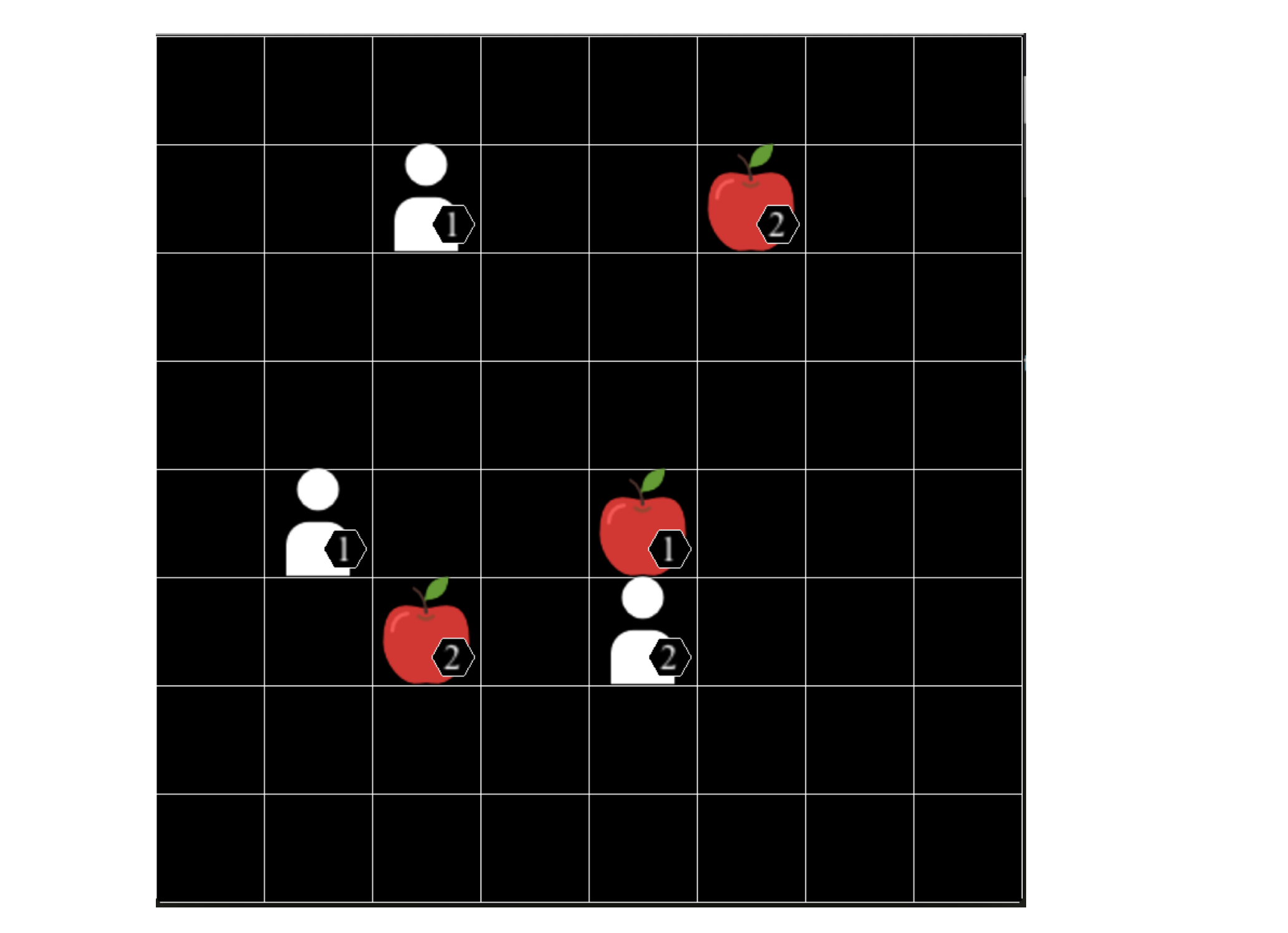}
  \caption{Level-based foraging}
  \label{lb-foragingscreen}
\end{subfigure}
\hfill
\begin{subfigure}{.35\textwidth}
  \centering
  \includegraphics[trim={1cm 20cm 0cm 2.5cm},clip,scale=.43]{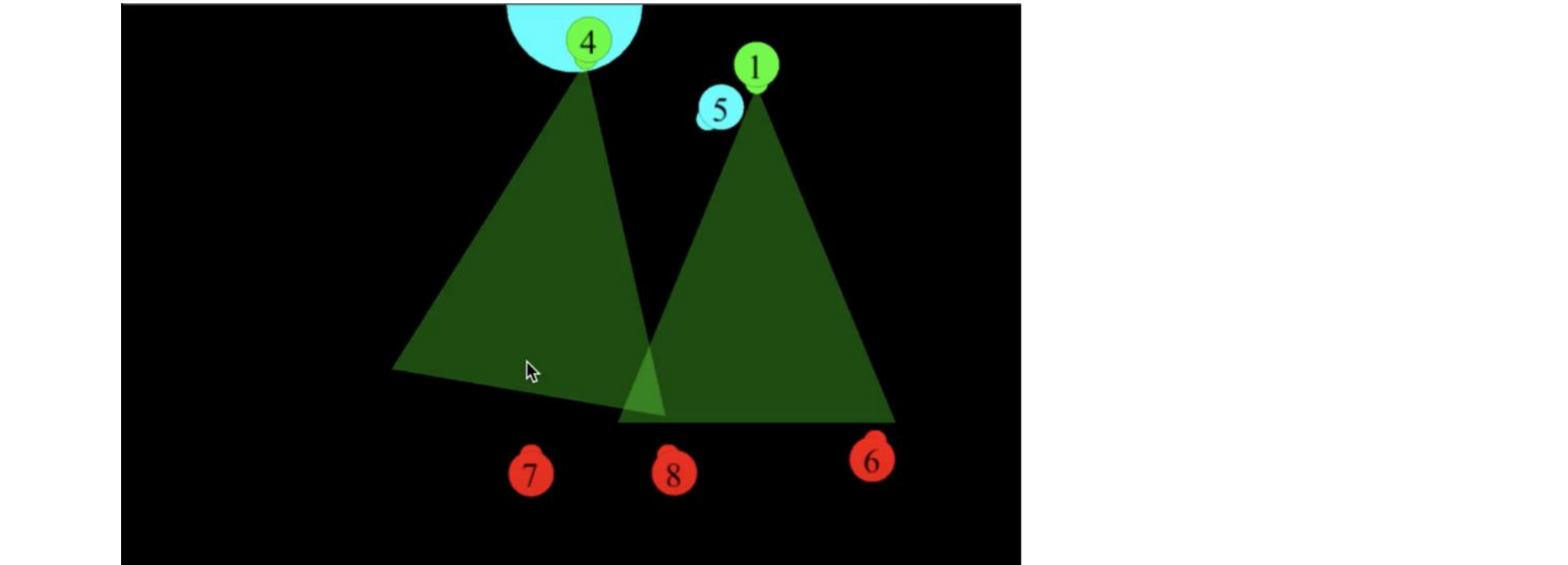}
  \caption{FortAttack}
  \label{fortattackscreen}
\end{subfigure}
\caption{An example screen capture of  (a) Wolfpack, (b) level-based foraging, and (c) FortAttack. In Wolfpack, the white colored grid cells represent the players while the red ones represent the prey. With level-based foraging, the apple icons represent the location of objects to remove while the white icons represent the players that attempt to remove these objects. Levels of objects and players are visualized next to their respective icons. With FortAttack, The blue semicircle on the northern part of the 2D world is the fort. The blue, green, and red dots represents the learning agent, defenders, and attackers respectively. Agent 1 and 4 are visualized in the middle of executing the shooting action and the green triangles represents their shooting range.}
\end{figure*}
\subsection{Wolfpack}
\label{wolfpack-env-description}
In the Wolfpack environment, we train preys to avoid capture via the DQN algorithm~\citep{mnih2015human}. Each prey receives an RGB image of the 17$\times$25 grid centered around the its location. If the 17$\times$25 image patch exceeds the boundaries of the grid world, we draw blue grid cells representing the grid world boundaries while also padding black colored grid cells outside the visualized boundaries to create the input image. The input image is subsequently used by a convolutional neural network to compute the action value estimates. We then train six convolutional neural networks using Independent DQN~\cite{tampuu2017multiagent} to control two preys and four wolves. During training prey are given a penalty of -1 each time they are captured while the wolves follow the same reward structure used to train GPL agents in Wolfpack. 

At the end of training value networks of the wolves are discarded and the value networks of the prey are used for our experiment. We finally provide Wolfpack as a single-agent reinforcement learning environment which can readily be used for open teamwork. An example image representing a state of the Wolfpack environment is provided in Figure~\ref{wolfpackscreen}. In our Wolfpack implementation, other teammates and prey are treated as non-playable characters and models or heuristics controlling them are provided as part of the environment source code.

\subsection{Level-based foraging}
In the level-based foraging environment, levels of players and objects are sampled uniformly from the set $L=\{1,2,3\}$. The number of objects in the environment is set to three for each episode. Furthermore, initial locations of agents and objects are sampled uniformly from the available locations in the grid. An episode either terminates after 50 timesteps or after all objects have been collected. An image representing an example state of the level-based foraging game is provided in Figure~\ref{lb-foragingscreen}.

\subsection{FortAttack}
The FortAttack environment limits the agents to have a position between -0.8 and 0.8 for the horizontal axis coordinate, while the vertical axis coordinate is limited between -1 and 1. The fort is a semicircle centered on (0.0,1.0) with a radius of 0.3. When being initialized, the vertical axis coordinate of the attackers is initialized between -1 and -0.8 to make attackers start from a location that is far from the fort. On the other hand, defender's vertical axis coordinate is initialized between 0.8 and 1.
\label{lb-foraging-env-description}
\subsection{Teammate policies}
\label{AdditionalPol}
We implement a diverse set of heuristics to control the teammates in Wolfpack and LBF. Further details of the heuristics used for both environments are provided in the following section. We provide empirical evidence showing that the set of heuristics is diverse and requires significant adaptation to achieve optimal performance by training an agent using the QL baseline against a specific type of teammate and evaluated the resulting policy against different types of teammates. We found that neither policies trained against a specific teammate nor a policy resulting from training against all possible types of teammates could reach the optimal performance against every teammate type. 

\begin{figure*}[h]
\centering
\begin{subfigure}{.5\textwidth}
  \centering
  \includegraphics[scale=.35]{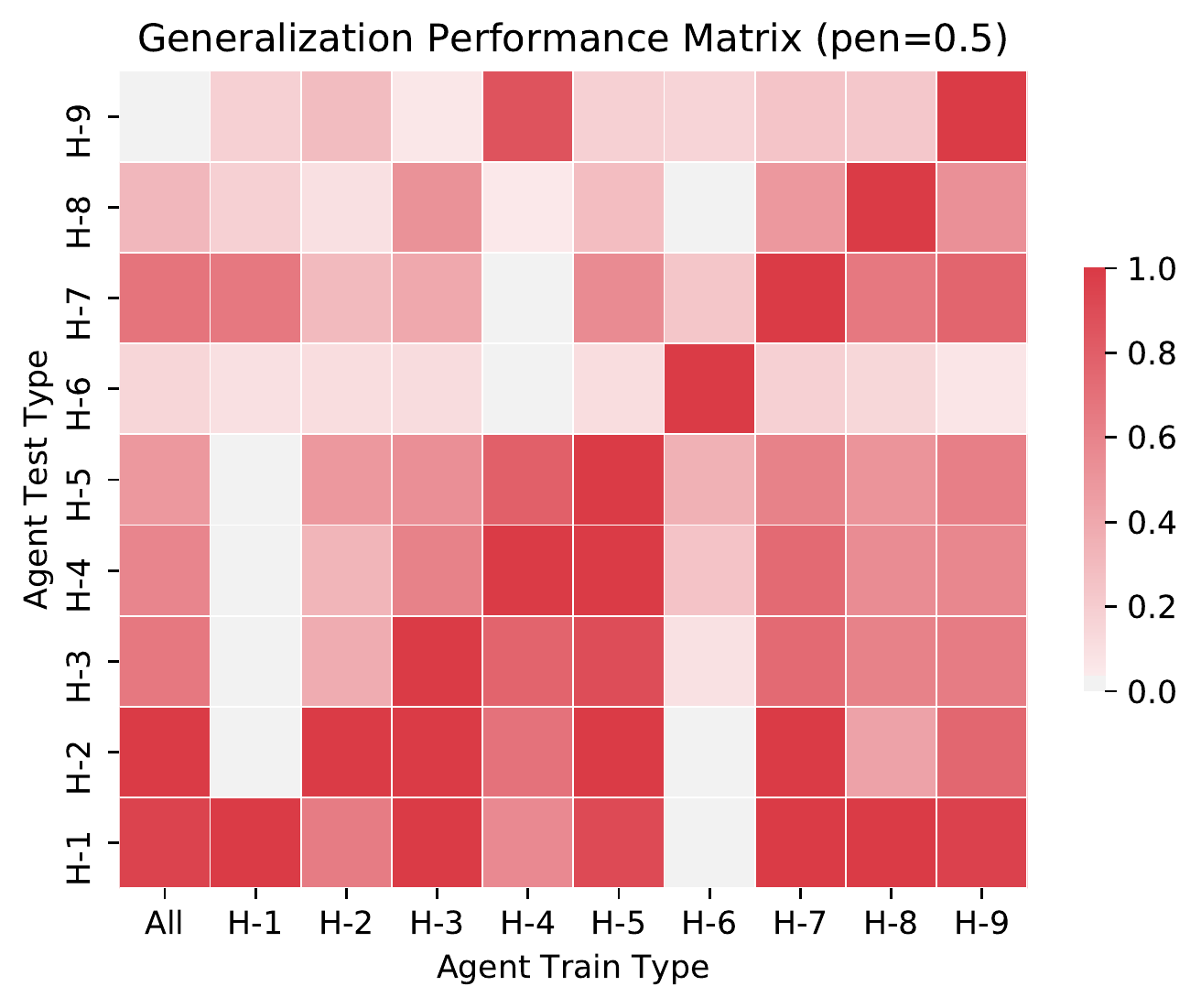}
  \caption{Wolfpack.}
  \label{wolfpack_teammate_gen}
\end{subfigure}%
\begin{subfigure}{.5\textwidth}
  \centering
  \includegraphics[trim={0cm 0.3cm 0cm 1cm},clip,scale=.35]{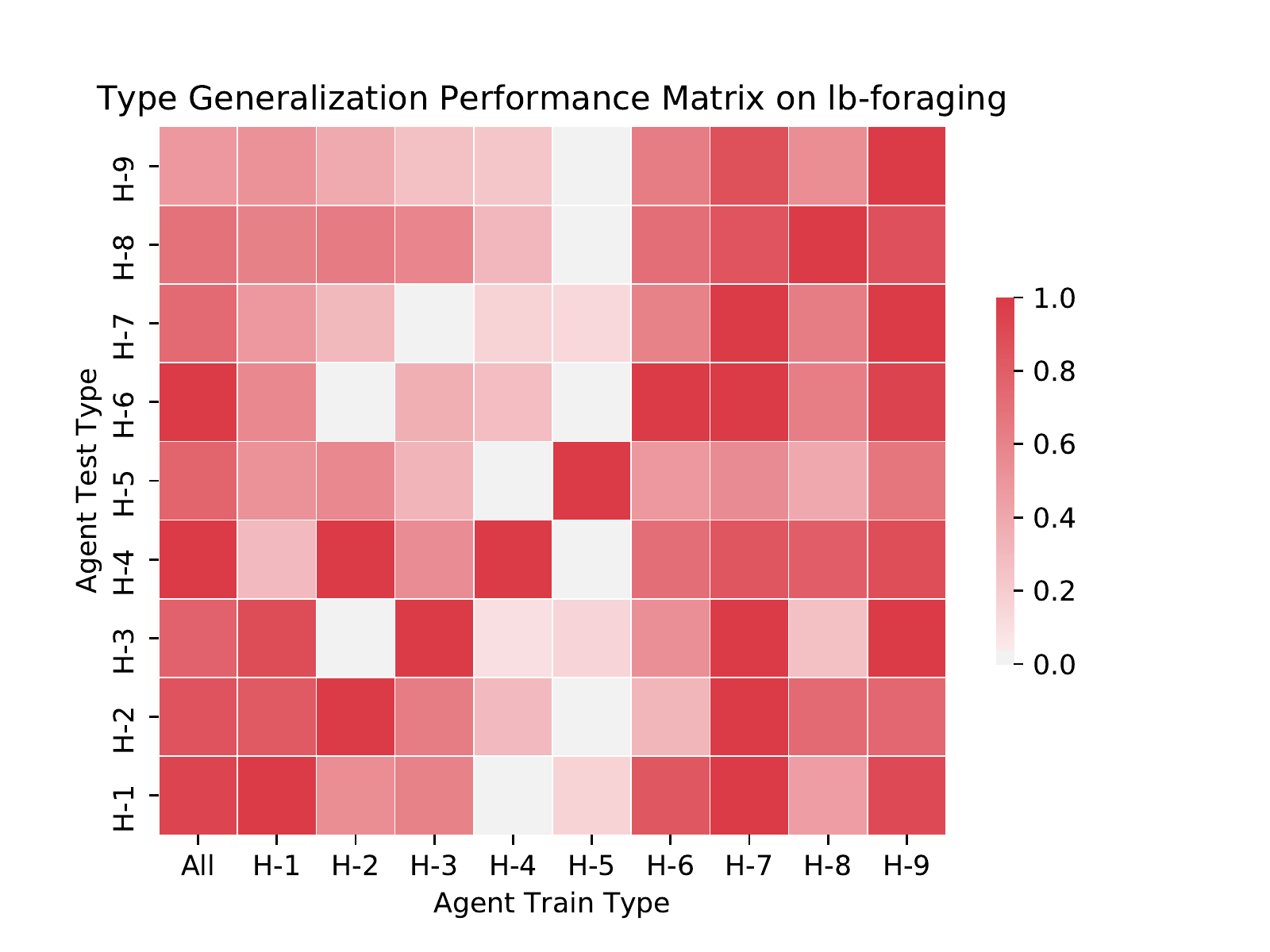}
  \caption{Level-based foraging.}
  \label{lbforaging-type}
\end{subfigure}
\caption{Generalization performance of a QL Agent trained to interact with a single teammate with a fixed type in both  Wolfpack with a penalty of 0.5 (a) and level-based foraging (b).  All experiments here are conducted using 4 seeds. The horizontal axis denotes the type of teammate encountered during training.  The “All” versions are trained against a random teammate sampled uniformly over all possible types at the start of each episode. The vertical axis denotes the type of agent used to test the trained policies. Performance in each evaluation environment is scaled between zero and one by performing min-max scaling on each row of the heat matrix.}
\label{TeammateTypes}
\end{figure*}

The results of this experiment for both environments are provided in Figure~\ref{TeammateTypes}. As we have done in the team size generalization experiments, for each approach we periodically checkpoint the policies resulting from training and choose the checkpoint with the highest performance in training to be evaluated and reported in the heat matrix visualization. Figure~\ref{TeammateTypes} shows that even for policies trained against all types of agents, none of the resulting policies consistently achieves optimal performance for all teammate types. 

On the other hand, for FortAttack we use the 5 pretrained policies provided by~\citet{deka2020natural}. In the original work that proposed FortAttack, GNN-based networks were trained to create stochastic policies to control attacker and defenders.~\citet{deka2020natural} subsequently visually analyzed the resulting behaviour of the trained policies along different checkpoints and found different adopted by attackers and defenders during the training process. An example policy, which we eventually used as the different agent types for our experiments, was given for each type of strategy adopted by the attackers and defenders. 
\subsubsection{Wolfpack}
To create a diverse set of teammates for open ad hoc teamwork, we used the following mixture between heuristics proposed by~\citet{barrett2011empirical} for the predator prey domain along with RL-based models to control teammates :
\begin{itemize}
\item \textbf{Random agent (H1): } The random agent chooses its action at any timestep by uniformly sampling the set of possible actions.
\item \textbf{Greedy agent (H2): } The greedy agent chooses its action following the greedy predator heuristic provided in~\citep{barrett2011empirical}. Intuitively, it sets the closest grid cell adjacent to the closest prey from its current location as its destination. It then chooses to move closer to the destination by moving along an axis for which it has the largest distance from the prey.
\item \textbf{Greedy probabilistic agent (H3): } The greedy probabilistic agent chooses its action following the greedy probabilistic predator heuristic provided in~\citep{barrett2011empirical}. The way it chooses its destination is the same with greedy agents. However, it randomly chooses one of the two available axis to move closer to the nearest prey. An agent's distance from the prey on each axis is provided as input to a boltzmann distribution to decide which axis should the agent move along.
\item \textbf{Teammate aware agents (H4): }This agent follows the teammate aware predator heuristic from ~\citep{barrett2011empirical}. Intuitively, this heuristics assumes all teammates are using the same heuristic. It subsequently computes a hierarchy between agents based on their distance to their targeted preys. The hierarchy determines the sequence in which agents choose their actions. Agents must take into account the actions of agents higher up the hierarchy to avoid collision. An $\text{A}^*$ planner is subsequently used to compute the action to reach the destination.
\item \textbf{GNN-Based teammate aware agents (H5): } We train an RFM model with supervised learning to predict the actions taken by a group of teammate aware agents. This was done to avoid the possibly slow running time of the $\text{A}^*$ planner in teammate aware agents. During interaction, it assumes that every agent is a teammate aware agent and passes their features along with prey locations as input to the network. Agent of this type subsequently uses the representation of its associated node as input to an MLP which has been trained to imitate the distribution over actions for teammate aware agents. Our agents subsequently samples the resulting distribution to decide their actions. 
\item \textbf{Graph DQN agents (H6): } We train an RFM-based controller trained by DQN to control a team of agents. It parses the state information following the input preprocessing method for GPL provided in Section~\ref{GPLpreprocessing} and provides it as input to an RFM. Node representations produced by the RFM are passed into an MLP to compute the action value function of the player associated to the node. Since this type only controls a single agent during interaction, only the action value associated to the controlled agent is used to take an action.
\item \textbf{Greedy waiting agents (H7): } This heuristic is similar to greedy agents. However, agents are equipped with a waiting radius sampled randomly between three to five. The greedy heuristic is followed when either the Manhattan distance between the agent and closest prey is more than the waiting radius or when there is already another teammate inside the waiting radius of the closest prey. Otherwise, the agent will uniformly sample an action until the prey moves away or another teammate comes close to the prey.
\item \textbf{Greedy probabilistic waiting agents (H8):} Similar to greedy waiting agents, agents of this type are equipped with a waiting radius sampled randomly between three to five. However, it is the greedy-probabilistic heuristic being followed when either the Manhattan distance between the agent and the targeted prey is more than the waiting radius or there is already another teammate inside the waiting radius. 
\item \textbf{Greedy team-aware waiting agents (H9): } Similar to greedy waiting agents, agents of this type are equipped with a waiting radius sampled randomly between three to five. However, it is the teammate aware heuristic being followed when either the Manhattan distance between the agent and the targeted prey is more than the waiting radius or there is already another teammate inside the waiting radius. 
\end{itemize}

\subsubsection{Level-based foraging}
Similar to Wolfpack, we create a diverse set of teammate types for level-based foraging which requires agents to adapt their policies towards their teammates for achieving optimal performance. With level-based foraging, we use a mixture of heuristics~\citep{Albrecht2013GameTheoretic, as2017} and controllers trained using the A2C algorithm~\cite{mnih2016asynchronous} as our teammate policies. With the heuristic-based agents, their observations are limited to a square patch of grid cells with the agent's location being the center of the grid. The size of this observation square is uniformly sampled between $3\times{3}$, $5\times{5}$, or $7\times{7}$. Details of the different types of heuristics used in level-based foraging are provided below: 
\begin{itemize}
    \item \textbf{Heuristic H1:} This type of agent follows heuristic $\theta_{j}^{F2}$ proposed by~\citet{as2017} where agents under this heuristic follow the agent with the highest level if it observes another agent with a higher level than its own. If no agent has a higher level, it follows the farthest observable agent from their locations instead. The controlled agent then computes the object targeted by the leader agent if they follow heuristic H3 provided below and chooses an action that will get itself closer to the target object. If the agent is already next to the targeted object, it will choose to pick up the object. If the agent cannot follow the aforementioned rules due to not observing any objects in their observation square, it chooses the leader's position as its target instead. If no other teammates are observed, it uniformly samples an action from the set of possible actions instead.
    \item \textbf{Heuristic H2:} This type of agent follows heuristic $\theta_{j}^{F1}$ proposed by~\citet{as2017} where the controlled agent chooses a leader agent, assumes they follow certain heuristics to choose their targeted object, and gets itself closer to the object they think is targeted by the leader. Unlike H1, it chooses an observable agent with the farthest distance from itself as its leader. Furthermore, it assumes that the leader follows heuristic H4 provided below in choosing its target object. Otherwise, the way it chooses its actions when it is next to the targeted object is the same as in H1. Furthermore, its action selection method when there are no objects or teammate agents in its observation square follows that of H1.
    \item \textbf{Heuristic H3:} This type of agent follows heuristic $\theta_{j}^{L2}$ proposed by~\citet{as2017} where the controlled agent targets objects that have the highest level below its own level and gets closer to the object. If no objects in the set of visible objects are below its level, it chooses to target the item with the highest level instead. Otherwise, it uniformly samples actions from the set of possible actions when there are no objects observed in its observation square.
    \item \textbf{Heuristic H4:} This type of agent follows heuristic $\theta_{j}^{L1}$ proposed by~\citet{as2017} where the controlled agent targets the farthest visible object from its current location and gets closer to the object. When no objects are visible in its observation square, it samples actions uniformly from the set of possible actions.
    \item \textbf{A2C Agent (H5):} This type of agent is produced by independently training four agents together in level-based foraging using the A2C algorithm~\citep{mnih2016asynchronous}. Among the four agents, we choose the one which has the highest performance compared to others as our A2C-based controller for this type. Also, agents of this type do not have observation squares but receive the whole state of the environment as input to their policy.
    \item \textbf{Heuristic H6:} This type of agent follows one of the heuristics proposed by~\citet{Albrecht2013GameTheoretic} for level-based foraging where agents always take actions that take them closer to the closest object in their observation square. If no objects exists, agents uniformly sample an action from the set of possible actions.
    \item \textbf{Heuristic H7:} This type of agent also follows one of the heuristics proposed by~\citet{Albrecht2013GameTheoretic}. In choosing their actions, agents of this type go to the object inside the observation square which is closest to the center of all observed players. It follows heuristic H6 when no objects are observed in its observation square.
    \item \textbf{Heuristic H8:} This type of agent follows a heuristic proposed by~\citet{Albrecht2013GameTheoretic} where agents choose the closest object with the same level or lower than their own level as their target. If none such objects exists, the agent uniformly samples an action from its action space.
    \item \textbf{Heuristic H9:} This type of agent follows a heuristic proposed by~\citet{Albrecht2013GameTheoretic} where it scans its surrounding for observable target objects that has at most the same level as the sum of all observable agents' levels. It then computes the center of all agents' locations and chooses a target object with the least distance to the center of observable agents' location as its destination. If no possible target exists, it uniformly samples an action from its action space.
\end{itemize}

\subsubsection{FortAttack} \label{sec:add_Details_fortattack}The behavior of attackers as well as defenders not under our control is determined by policies obtained in the experiments from \citet{deka2020natural}. The policies show distinct behavioral patterns summarized below. More information on how these policies were obtained can be found in \citet{deka2020natural}.

\begin{itemize}
    \item \textbf{Guard Type 1 - Random guard:} For this agent, we randomly initialize a neural network and used it as the policy network for the guards. 
    \item \textbf{Guard Type 2 - Flash laser:} Defenders position themselves in front of the fort and flash their lasers continually. This behavior is independent of the movement of the attackers.
    \item \textbf{Guard Type 3 - Spread out Flash laser:} Similar to type 1 but instead of positioning themselves in front of the fort the defenders spread out across the whole width of the environment.  
    \item \textbf{Guard Type 4 - Smart spreading:} Defenders spread smartly across the defensive zone and only shoot to kill attackers.
    \item \textbf{Attacker Type 1 - Sneak:} Attackers spread out across the whole environment to maximize the likelihood of finding an open space in the defensive line. 
    \item \textbf{Attacker Type 2 - Deceive:} Attackers split up their attack. That is if the majority of attackers come from the right, one attacker will try to sneak beyond the defenders from the other side.  
    \item \textbf{Guard Type 6 - Ensemble-trained agents:} Guards of this type are trained by letting it interact against Attackers that are uniformly sampled from Attacker type 1 and type 2.

\end{itemize}

\section{GPL Input preprocessing}
\label{GPLpreprocessing}
\begin{figure*}[t]
    \centering
    \begin{subfigure}[b]{0.35\textwidth}
        \centering
        \includegraphics[trim={3.5cm 14.5cm 4cm 5.5cm},clip, scale=0.45]{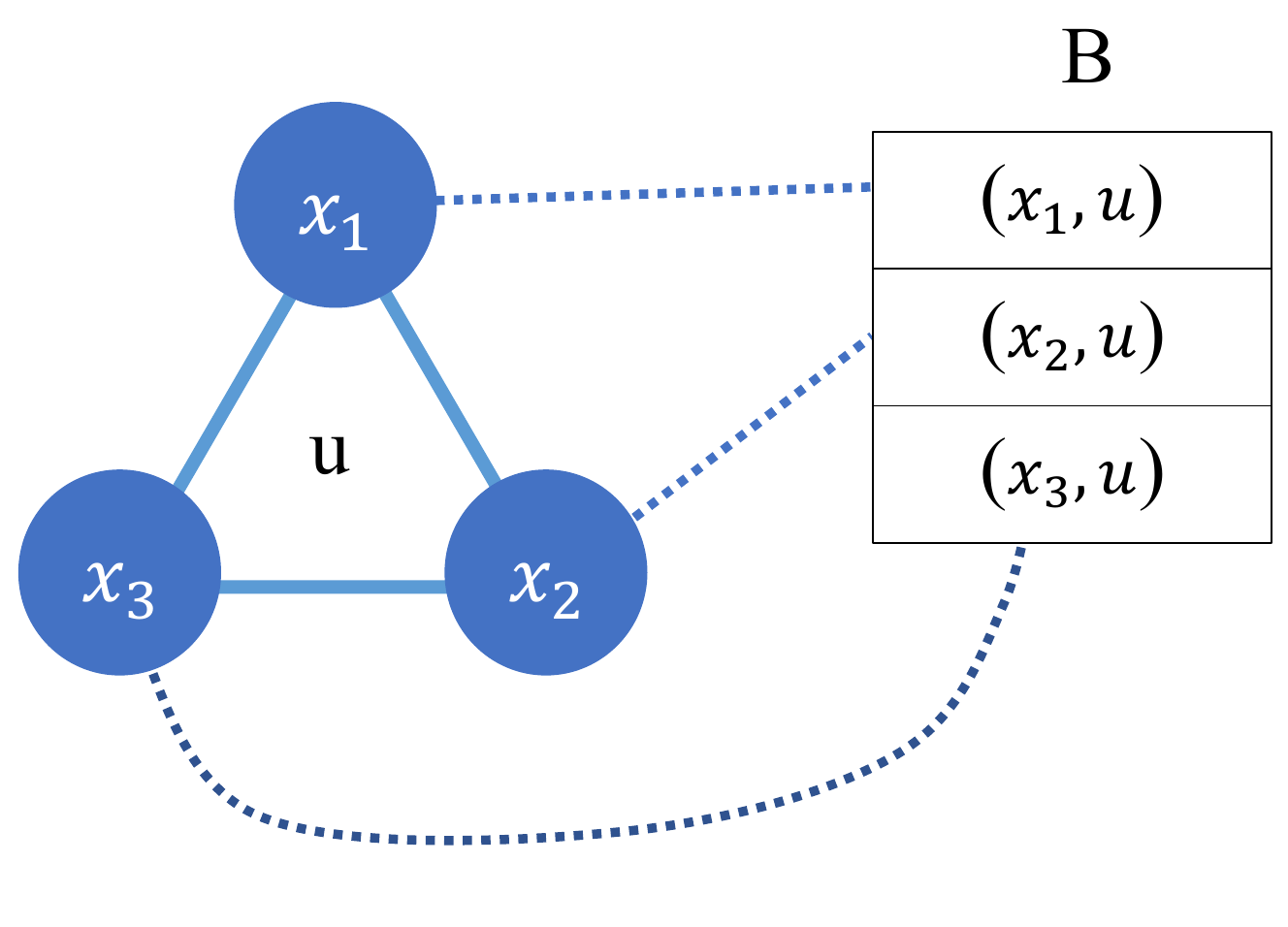}
        \caption{Observation preprocessing.}
        \label{fig:inp-preprocessing}
    \end{subfigure}%
    \begin{subfigure}[b]{0.65\textwidth}
        \centering
        \includegraphics[trim={3.cm 4.5cm 4.2cm 2.5cm},clip, scale=0.40]{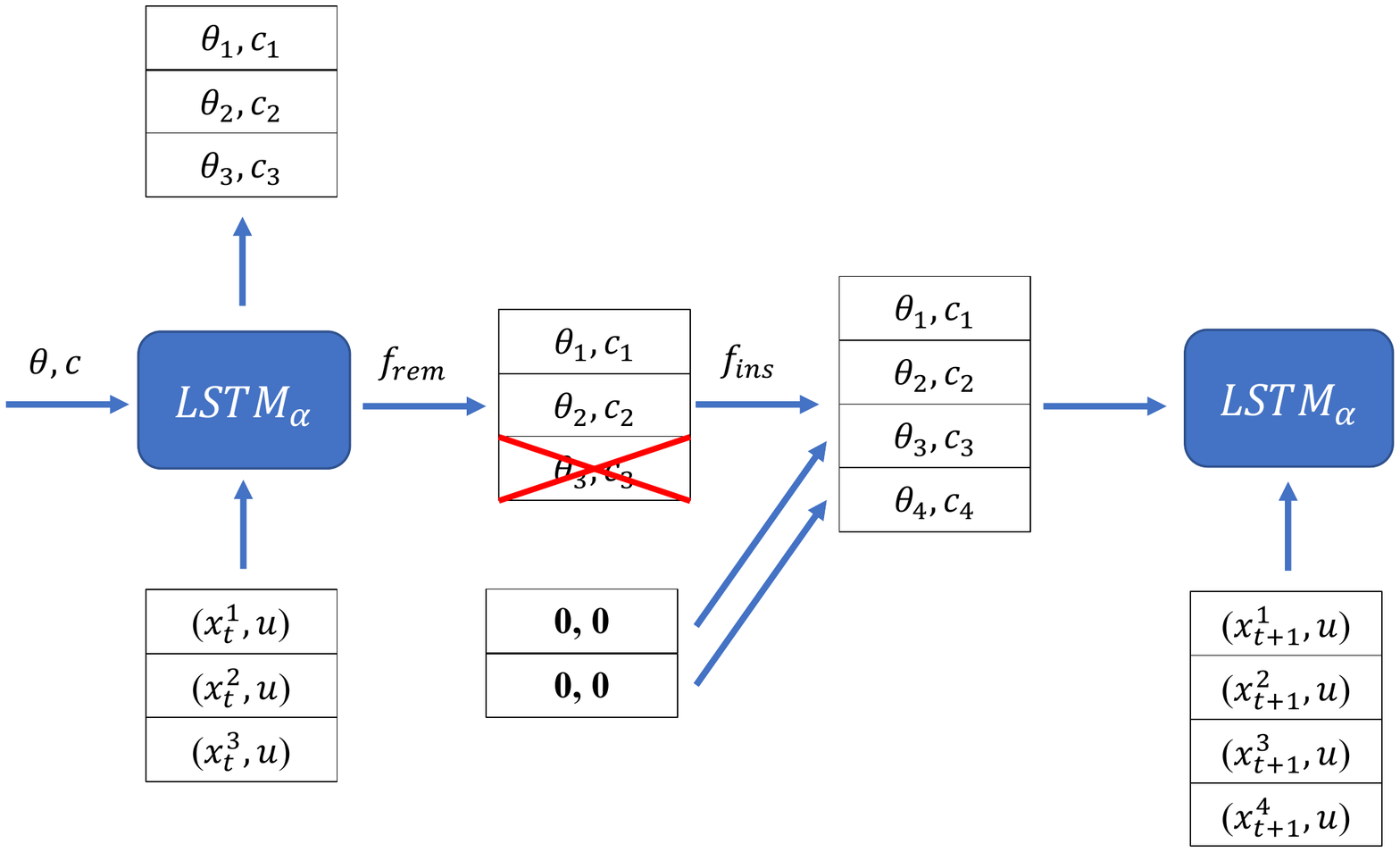}
        \caption{LSTM hidden vector preprocessing.}
        \label{fig:h-postprocessing}
    \end{subfigure}%
    \caption{The figure shows (a) the preprocessing of observation information into input for the GPL algorithm along with (b) the additional processing steps done to the agent embedding vectors to handle environment openness. Part (b) shows an example processing step where agent 3 is removed from the environment and two new agents joins the environment.}  
    \label{fig:pre_post_processing}
\end{figure*}
As input to GPL, for each agent the observation is parsed into a set of vectors containing agent specific information, $x$, concatenated with shared state information, $u$, to create an input batch $B$. The agent specific information generally contains locations of the agents in both environments. In level-based foraging, information about an agent's level is also included in $x$. Other remaining information such as prey locations in Wolfpack or object location and level in level-based foraging are included in $u$. $B$ is then passed to an LSTM along with the previous hidden states, $\theta$, and cell states, $c$, of an LSTM to compute the embedding of each agent required by the models defined in GPL. 

To deal with the changing batch size of $B$ across time as a result of environment openness, the hidden and cell state are further processed inbetween timesteps. Assuming $i_{t}$ and $d_{t}$ correspond to the sets of added and removed agents at time $t$, $f_{rem}$ removes the states associated to agents leaving the environment while $f_{ins}$ inputs a zero vector for the states associated to agents joining the environment. We formally define this hidden and cell state processing following Equation~\ref{PostprocessingEq} while additionally providing an example illustration of this processing method in Figure~\ref{fig:pre_post_processing}. Finally, since we assume that the hidden and cell states associated to all agents including the learning agent should be reset to \textbf{0} at the end of each episode, we define $d_{t}$ as the set of all existing agents and $i_{t}$ as the set of agents at the initial state of the following episode each time an episode ends.
\begin{equation}
    \label{PostprocessingEq}
    Prep(\theta_{t}, c_{t}) = f_{ins}(f_{rem}(\theta_{t}, c_{t}, d_{t}), i_{t})
\end{equation}

\section{GPL pseudocode}
\label{GPLalgorithm}
Before we describe the full GPL pseudocode, we first define important functions that we will use in the pseudocode. First, we denote the observation and hidden vector preprocessing method described in Appendix~\ref{GPLpreprocessing} as the \textbf{PREPROCESS} function. Furthermore, we denote the action-value and joint-action value computation through Equation~\eqref{SingleActionValue} and ~\eqref{JointActionValueComputation} as the \textbf{MARGINALIZE} and \textbf{JOINTACTEVAL} functions respectively. Based on these functions, we define the \textbf{QV} function that preprocesses the input and computes the action-values for given joint-action value and agent networks. The computations in \textbf{QV} is provided in Algorithm~\ref{alg:ActValComputation}.
\begin{algorithm}[!htb]
   \caption{GPL Action Value Computation}
   \label{alg:ActValComputation}
\begin{algorithmic}
   \STATE {\bfseries Input:} state $s$, \\ joint-action value model parameters ($\alpha_{Q}, \beta,\delta$),\\ agent model parameters ($\alpha_{q}, \eta,\zeta$), \\ agent model LSTM hidden vectors  $h_{t-1,q}$, \\ joint-action value model LSTM hidden vectors $ h_{t-1,Q}$
   \FUNCTION{\textbf{QV}($s, \alpha_{Q}, \alpha_{q}, \beta, \delta, \eta, \zeta, h_{t-1,Q}, h_{t-1,q}$)}
   \STATE $\mathrm{B}, \theta_{Q}, c_{Q} \leftarrow \mathrm{\textbf{PREPROCESS}}(s, h_{t-1,Q})$
   \STATE $\mathrm{B}, \theta_{q}, c_{q} \leftarrow \mathrm{\textbf{PREPROCESS}}(s, h_{t-1,q})$
   \STATE $\theta'_{Q}, c'_{Q} \leftarrow \mathrm{LSTM}_{\alpha_{Q}}(B, \theta_{Q}, c_{Q})$
   \STATE $\theta'_{q}, c'_{q} \leftarrow \mathrm{LSTM}_{\alpha_{q}}(B, \theta_{q}, c_{q})$
   \STATE $\forall{j}, \bar{n}_{j} \leftarrow (RFM_{\zeta}(\theta'_{q}, c'_{q}))_{j}$
   \STATE $\forall{j}, q_{\eta,\zeta}(.|s_{t}) \leftarrow \mathrm{Softmax}(\mathrm{MLP}_{\eta}(\bar{n}_{j}))$
   \STATE $\forall{j, a^{j}}, Q^{j}_{\beta}(a^{j}|s_{t}) \leftarrow \textrm{MLP}_{\beta}(\theta_{Q}^{'j}, \theta_{Q}^{'i})(a^{j})$
   \STATE $\forall{j, a^{j}, a^{k}}$,
   $$
   Q^{j,k}_{\delta}(a^{j},a^{k}|s_{t}) \leftarrow \textrm{MLP}_{\delta}(\theta_{Q}^{'j},\theta_{Q}^{'k}, \theta_{Q}^{'i})(a^{j},a^{k})$$
   \STATE Compute $\bar{Q}(s,a^{i})$ using Equation~\eqref{SingleActionValue}
   \STATE $\bar{Q}(s,.) \leftarrow \mathrm{\textbf{MARGINALIZE}}($
   $$q_{\eta,\zeta}(.|s_{t}), Q_{\beta}(.|s_{t})), Q_{\delta}(.,.|s_{t})$$
   )
   \STATE \textbf{return} $\bar{Q}(s,.), (\theta'_{Q}, c'_{Q}), (\theta'_{q}, c'_{q})$
   \ENDFUNCTION
\end{algorithmic}
\end{algorithm}

Aside from these functions, we define \textbf{QJOINT} and \textbf{PTEAM}, which output is required to compute the loss functions, $L_{\beta,\delta}$ and $L_{\eta,\zeta}$, in Equation~\eqref{ValueLoss} and~\eqref{ActionModelLoss}. \textbf{QJOINT} is a function that computes the predicted joint action value for an observed state and joint actions. On the other hand, \textbf{PTEAM} computes the joint teammate action probability at a state. Both \textbf{QJOINT} and \textbf{PTEAM} are further defined in Algorithm~\ref{alg:jointactval} and~\ref{alg:jointactprobcomp}.

\begin{algorithm}[!htb]
   \caption{GPL Joint-Action Value Computation}
   \label{alg:jointactval}
\begin{algorithmic}
   \STATE {\bfseries Input:} state $s$, observed joint action a, \\ joint-action value model parameters ($\alpha_{Q}, \beta,\delta$),\\ joint-action value model LSTM hidden vectors $ h_{t-1,Q}$
   \FUNCTION{\textbf{QJOINT}($s, a, \alpha_{Q}, \beta, \delta, h_{t-1,Q}$)}
   \STATE $\mathrm{B}, \theta_{Q}, c_{Q} \leftarrow \mathrm{\textbf{PREPROCESS}}(s, h_{t-1,Q})$
   \STATE $\theta'_{Q}, c'_{Q} \leftarrow \mathrm{LSTM}_{\alpha_{Q}}(B, \theta_{Q}, c_{Q})$
   
   \STATE $\forall{j, a^{j}}, Q^{j}_{\beta}(a^{j}|s_{t}) \leftarrow \textrm{MLP}_{\beta}(\theta_{Q}^{'j}, \theta_{Q}^{'i})(a^{j})$
   \STATE $\forall{j, a^{j}, a^{k}}$,
   $$
   Q^{j,k}_{\delta}(a^{j},a^{k}|s_{t}) \leftarrow \textrm{MLP}_{\delta}(\theta_{Q}^{'j},\theta_{Q}^{'k}, \theta_{Q}^{'i})(a^{j},a^{k})$$
   \STATE Compute $Q(s,a)$ using Equation~\eqref{JointActionValueComputation}\\ $Q(s,a) \leftarrow \mathrm{\textbf{JOINTACTEVAL}}($\\
   \begin{center}
       $a, q_{\eta,\zeta}(.|s_{t}), Q_{\beta}(.|s_{t})), Q_{\delta}(.,.|s_{t})$
   \end{center}
   )
   \STATE \textbf{return} $Q(s,a)$
   \ENDFUNCTION
\end{algorithmic}
\end{algorithm}

\begin{algorithm}[!htb]
   \caption{GPL Teammate Action Probability Computation}
   \label{alg:jointactprobcomp}
\begin{algorithmic}
   \STATE {\bfseries Input:} state $s$, observed joint actions $a$,\\ agent model parameters ($\alpha_{q}, \eta,\zeta$), \\ agent model LSTM hidden vectors  $h_{t-1,q}$
   \FUNCTION{\textbf{PTEAM}($s, a, \alpha_{q}, \eta, \zeta, h_{t-1,q}$)}
   \STATE $\mathrm{B}, \theta_{q}, c_{q} \leftarrow \mathrm{\textbf{PREPROCESS}}(s, h_{t-1,q})$
   \STATE $\theta'_{q}, c'_{q} \leftarrow \mathrm{LSTM}_{\alpha_{q}}(B, \theta_{q}, c_{q})$
   \STATE $\forall{j}, \bar{n}_{j} \leftarrow (RFM_{\zeta}(\theta'_{q}, c'_{q}))_{j}$
   \STATE $\forall{j}, q^{j}_{\eta,\zeta}(.|s) \leftarrow \mathrm{Softmax}(\mathrm{MLP}_{\eta}(\bar{n}_{j}))$
   \STATE $q_{\eta,\zeta}(a^{-i}|s, a^{i}) \leftarrow \prod_{j\in{-i}}q^{j}_{\eta,\zeta}(a^{j}|s)$
   \STATE \textbf{return} $q_{\eta,\zeta}(a^{-i}|s, a^{i})$
   \ENDFUNCTION
\end{algorithmic}
\end{algorithm}

Using the functions we previously defined, we finally describe GPL's training algorithm. GPL collects experience from parallel environments through the modified Asynchronous Q-Learning framework~\citep{mnih2016asynchronous} where asynchronous data collection is replaced with a synchronous data collection instead. Despite this, it is relatively straightforward to modify the pseudocode to use an experience replay instead of a synchronous process for data collection. As in the case of existing deep value-based RL approaches, we also use a separate target network whose parameters are periodically copied from the joint action value model to compute the target values required for optimizing Equation~\ref{ValueLoss}. We finally optimize the model parameters in the pseudocode to optimize the loss function provided in Section~\ref{Sec:GPLArchitecture} using gradient descent. GPL's training process is finally described in Algorithm~\ref{alg:fullGPL}.

\begin{algorithm*}[!htb]
   \caption{GPL Training}
   \label{alg:fullGPL}
\begin{algorithmic}
 \STATE \textbf{Input: }Number of training steps $T$, time between updates $t_{update}$, time between target network updates $t_{targ\_update}$.
   \STATE Initialize the joint-action value model parameters, $\alpha_{Q}, \beta, \delta$.
   \STATE Initialize the agent model parameters, $\alpha_{q}, \eta, \zeta$.
   \STATE Create target joint-action value networks.\\
   \begin{center}
       $\alpha'_{Q}, \beta', \delta' \leftarrow \alpha_{Q}, \beta, \delta$ 
   \end{center}
   
   \STATE $\theta_{Q}, c_{Q}, \theta^{targ}_{Q}, c^{targ}_{Q} \leftarrow \mathbf{0,0,0,0}$
   \STATE $\theta_{q}, c_{q} \leftarrow \mathbf{0,0}$
   \STATE $d\alpha_{Q}, d\alpha_{q}, d\beta, d\delta, d\eta, d\zeta \leftarrow \mathbf{0, 0, 0, 0, 0, 0}$
   \STATE Observe $s$ from environment
   \FOR{$t=1$ {\bfseries to} $T$}
   \STATE $h_{Q}, h_{q}, h_{Q}^{targ} \leftarrow (\theta_{Q}, c_{Q}),(\theta_{q}, c_{q}), (\theta^{targ}_{Q}, c^{targ}_{Q})$
   \STATE $\bar{Q}(s,.), h'_{Q}, h'_{q} \leftarrow \mathrm{\textbf{QV}}(s, \alpha_{Q}, \alpha_{q},\beta, \delta, \eta, \zeta, h_{Q}, h_{q})$
   \STATE Sample action according to the learning algorithm being used, \\\begin{center}
   $a^{i}_{t} \sim \begin{cases}
    \text{eps-greedy}(\epsilon, \bar{Q}(s,.)),& \text{if Q-Learning}\\
    p_{\text{SPI}}(\bar{Q}(s,.),\tau)             & \text{if SPI}
    \end{cases}
    $
    \end{center}
    \STATE Execute $a^{i}$ and observe $a, r$ and $s'$.
    \STATE Compute predicted joint-action value for $a_{t}$,\\\begin{center}
        $Q_{\beta,\delta}(s,a) \leftarrow \mathrm{\mathbf{QJOINT}}(s, a, \alpha_{Q}, \beta, \delta, h_{Q})$
    \end{center}
    \STATE Compute action-value of next state using target network. \\ \begin{center}
        $\bar{Q}'\left(s,a^{i}\right),h^{targ}_{Q},\_ \leftarrow \mathrm{\textbf{QV}}(s', \alpha'_{Q}, \alpha_{q}, \beta', \delta', \eta, \zeta, h^{targ}_{Q}, h'_{q})$
    \end{center}
    
    \STATE Compute target value for updating the joint-action value model with, \\\begin{center}
    $ y\left(r, s'\right) = 
    r + \gamma \text{max}_{a^{i}}\bar{Q}'\left(s',a^{i}\right),$\\\end{center} if Q-Learning is used, or \\ \begin{center}$
     y\left(r, s'\right) = r + \gamma \sum_{a^{i}}p_{\mathrm{SPI}}(a^{i}|s')\bar{Q}'\left(s',a^{i}\right), 
    $\\\end{center} if using SPI.
    \STATE Compute predicted action probabilities of teammates using the agent models,\\ \begin{center}
        $q_{\eta,\zeta}(a^{-i}|s, a^{i}) \leftarrow \mathrm{\mathbf{PTEAM}}(s, a, \alpha_{q}, \eta, \zeta, h_{q})$
    \end{center}
    
    \STATE Using $Q_{\beta,\delta}(s_{t},a_{t}), y\left(r_{t}, s_{t+1}\right)$, and $q_{\eta,\zeta}(a^{-i}|s, a^{i})$, compute $L_{\zeta,\eta}$ and $L_{\beta,\delta}$ with Equation~\eqref{ActionModelLoss} and~\eqref{ValueLoss}.
    \STATE Accumulate parameter gradients for updates\\\begin{center}
        $d\alpha_{Q} = d\alpha_{Q} + \nabla_{\alpha_{Q}} L_{\beta,\delta}$, $d\alpha_{q} = d\alpha_{q} + \nabla_{\alpha_{q}} L_{\eta,\zeta}$\\
        $d\beta = d\beta + \nabla_{\beta} L_{\beta,\delta}$, $d\delta = d\delta + \nabla_{\delta} L_{\beta,\delta}$\\
        $d\eta = d\eta + \nabla_{\eta} L_{\eta,\zeta}$, $d\zeta = d\zeta + \nabla_{\zeta} L_{\eta,\zeta}$
    \end{center}
    \IF{$t$ mod $t_{\mathrm{update}}$ = 0}
        \STATE Update $\alpha_{Q}, \alpha_{q}, \beta, \delta, \eta, \zeta$ using gradient descent based on $d\alpha_{Q}, d\alpha_{q}, d\beta, d\delta, d\eta, d\zeta$.
        \STATE $d\alpha_{Q}, d\alpha_{q}, d\beta, d\delta, d\eta, d\zeta \leftarrow \mathbf{0, 0, 0, 0, 0, 0}$
    \ENDIF
    \IF{$t$ mod $t_{\mathrm{targ\_update}}$ = 0}
        \STATE $\alpha'_{Q}, \beta', \delta' \leftarrow \alpha_{Q}, \beta, \delta$ 
    \ENDIF
     \STATE $(\theta_{Q}, c_{Q}),(\theta_{q}, c_{q}), s \leftarrow h'_{Q}, h'_{q}, s' $
   \ENDFOR
\end{algorithmic}
\end{algorithm*}

\section{Training Details}
\label{arch_exp_detail_appendix}

In this section we provide additional details about the training setup. First, we describe the way we add and remove agents in our open teamwork experiments. Secondly, we also provide details of the input padding method used by the baseline algorithms. Finally, we provide details of the architecture used in our experiments along with the hyperparameters used for training.

\subsection{Environment openness}
\label{sec:EnvOpenness}
To induce environment openness in Wolfpack and LBF, we sample the duration for which an agent exists in the environment, along with the waiting duration required for an agent which is removed from the environment to get added to the environment. For Wolfpack, the active duration is sampled uniformly between 25 and 35 timesteps while the waiting duration is sampled uniformly between 15 and 25 timesteps. For level-based foraging, the active duration is sampled uniformly between 15 to 25 timesteps while the waiting duration is sampled uniformly between 10 to 20 timesteps. 

In both environments, active duration is designed to be longer than waiting duration to create environments with large team sizes during interaction. On the other hand, both active and waiting duration for level-based foraging is less than its Wolfpack counterpart since the objects collected in level-based foraging remain stationary which causes the task to require less time to solve compared to Wolfpack. As a consequence, an episode might finish early and using larger active and waiting durations might cause agents to not be added or removed from the environment at all.

On the other hand, openness in FortAttack is solely induced by agents getting destroyed or respawned. When an agent is destroyed, its distance to the agent that shot it down determines the waiting time required before it can reenter the environment. Specifically, we measure the distance between the destroyed agent and the shooter agent and linearly interpolate between 0 and 80 timesteps and round it to the closest integer to determine the waiting time. Agents are out for longer as they get closer to the shooter when they die. 

Finally, for each environment, we uniformly sample the type of a teammate from the pool of possible teammate types when they are respawned. To provide different open processes for training and generalization, we vary the upper limit for team sizes in LBF, Wolfpack, and the attacking \& defending teams in FortAttack. During training, the team size is limited to up to 3 agents. On the other hand, the upper limit is increased to 5 agents for each team in the generalization scenario.

\subsection{Baseline input preprocessing}
\label{sec:InputPreprocessingLSTM}
For our baseline approaches, in Section~\ref{sec:BaselineDescription} we mentioned about padding the input vectors. We specifically use a placeholder value of -1 for features associated to inactive agents. Furthermore, since our generalization experiments can have up to five agents in the environment, these placeholders are added to the input until the length of the input vector corresponds to inputs of an environment with five agents. Adding placeholder values of -1 also applies to the predicted action probability concatenated to the input vectors for the QL-AM baseline.

We also need to ensure that a feature is not always assigned a placeholder value during training to prevent its associated model parameters from not being able to generalize when encountering input vectors which do not have placeholder values assigned to the feature. To handle this, we randomly assign teammates an integer index between one and four when they are added to the environment. We prevent different teammates from having the same index and use it to determine the location of their features in the input vector. This index remains the same while an agent is active in the environment. As a result, all features are assigned a non-placeholder value at some point during training.

Aside from concatenating the predicted action probabilities to the input vector, we tried the approach proposed by~\citet{DBLP:conf/iclr/TacchettiSMZKRG19} which maps the output of the agent model into an RGB image which contains information about the probability of an agent being positioned at certain grid cells following the action it might execute at the current state. We use the same convolutional neural network architecture used in their work to produce a fixed-length embedding of the image. This embedding is subsequently concatenated to the input vector and passed as input to a deep RL approach. However, we decided not to use this approach as a baseline due to the following reasons:

\begin{itemize}
    \item Our experiments indicate no improvement in performance in the 6.4 million steps we used to train our approaches. Due to the increased number of parameters introduced by the convolutional neural network, more training steps might be required and we have not found the number of steps that works for this approach yet. 
    \item A fair comparison against GPL might be difficult since GPL does not receive RGB images as input. 
    \item Actions in level-based foraging which lead to the same teammate location in the next state such as staying still and retrieving the object cannot be straightforwardly mapped into an RGB image following the mapping approach proposed by~\citet{DBLP:conf/iclr/TacchettiSMZKRG19}. 
\end{itemize}
Nonetheless, we view the approach that concatenates predicted action probabilities to the input vector as an adequate representative of approaches that augment the input with teammate information and relies on a non-linear function approximation to learn a policy or value estimate for the agent.

\subsection{Hyperparameters and network architecture}
\begin{figure*}[t!] 
\centering
\begin{subfigure}{0.31\textwidth}
\scalebox{1}{\includegraphics[trim={5cm 13.cm 12.cm 4.cm}, clip, width=\linewidth, scale=0.45]{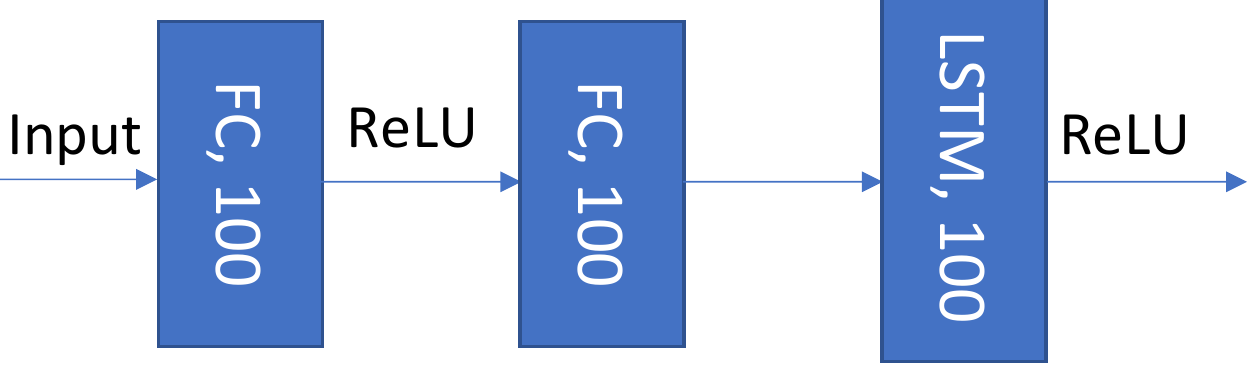}}
\caption{$LSTM_{\alpha_{Q/q/p}}$} \label{fig:a}
\end{subfigure}\hspace*{\fill}
\centering
\begin{subfigure}{0.31\textwidth}
\scalebox{1}{\includegraphics[trim={5cm 8.5cm 12.cm 8.7cm}, clip, width=\linewidth, scale=0.45]{imgs/ArchSizes/ArchSize2.pdf}}
\caption{$MLP_{\beta}$ } \label{fig:b}
\end{subfigure}\hspace*{\fill}
\centering
\begin{subfigure}{0.31\textwidth}
\includegraphics[trim={5cm 4.2cm 12.cm 12.5cm},clip, width=\linewidth, scale=0.45]{imgs/ArchSizes/ArchSize2.pdf}
\caption{$MLP_{\delta}$ } \label{fig:c}
\end{subfigure}

\medskip
\centering
\begin{subfigure}{0.31\textwidth}
\scalebox{1}{\includegraphics[trim={20cm 4.2cm 0.cm 12.5cm}, clip, width=\linewidth, scale=0.35]{imgs/ArchSizes/ArchSize2.pdf}}
\caption{Edge processing ($GNN_{\mu / \zeta}$)} \label{fig:d}
\end{subfigure}\hspace*{\fill}
\centering
\begin{subfigure}{0.31\textwidth}
\scalebox{1}{\includegraphics[trim={20cm 4.2cm 0.cm 12.5cm}, clip, width=\linewidth, scale=0.35]{imgs/ArchSizes/ArchSize2.pdf}}
\caption{Node processing ($GNN_{\mu / \zeta}$)} \label{fig:e}
\end{subfigure}\hspace*{\fill}
\centering
\begin{subfigure}{0.31\textwidth}
\scalebox{1}{\includegraphics[trim={20cm 13cm 0.cm 4cm}, clip, width=\linewidth, scale=0.35]{imgs/ArchSizes/ArchSize2.pdf}}
\caption{$MLP_{\mu}$ and $MLP_{\eta}$ } \label{fig:f}
\end{subfigure}

\medskip
\centering
\begin{subfigure}{0.55\textwidth}
{\hspace{2.5cm}\includegraphics[trim={5cm 7.8cm 6.5cm 17.5cm}, clip, scale=0.60]{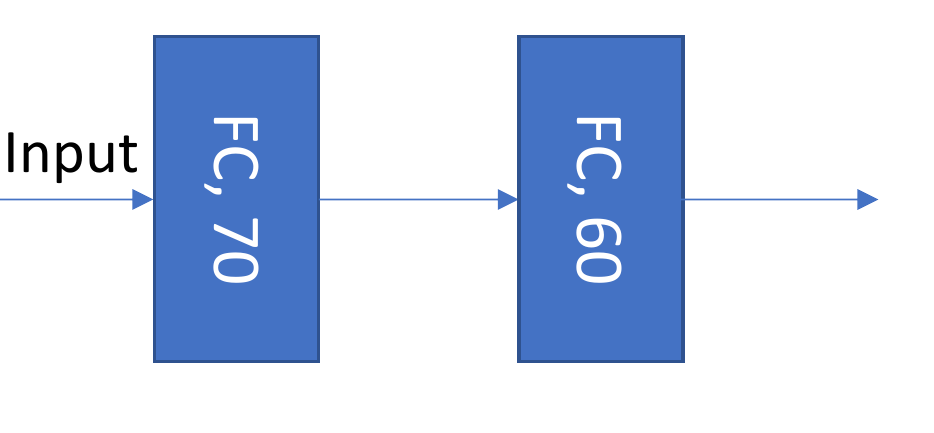}}
\caption{Key, query, and value computation.} \label{fig:g}
\end{subfigure}
\centering
\begin{subfigure}{0.25\textwidth}
\scalebox{1}{\hspace{0.5cm}\includegraphics[trim={5cm 3.3cm 6.5cm 22.cm}, clip, scale=0.60]{imgs/ArchSizes/ArchSizeUpdated.pdf}}
\caption{Action value computation.} \label{fig:h}
\end{subfigure}\hspace*{\fill}
\centering

\caption{\textbf{GPL and baseline architecture details for LBF and Wolfpack:} We provide details of the architecture used in GPL's type embedding network (a), singular utility computation (b), pairwise utility computation (c), edge embedding computation in the agent and auxiliary agent model (d), node embedding computation (e) in the agent and auxiliary agent model, the MLP used by the agent and auxiliary agent model to process the resulting GNN node embeddings (f), the size of the MLPs used for computing the key, query, and value for multihead attention in GNN-QL and GNN-QL-AM (g), and the final layer used in action value computation for GNN-QL \& GNN-QL-AM. In all images, \textbf{FC} denotes a fully connected layer, \textbf{LSTM} denotes an LSTM layer, and the accompanying number denotes the size of the layer. Labels on the arrows indicate the non-linear functions used between the layers while no labels indicate no non-linear functions being applied to the resulting output vectors. With the baselines, we combine the architectures in (a) and (b) to compute the action values while the agent and auxiliary agent model used in some of the baselines follow (d), (e), and (f). With GNN-QL and GNN-QL-AM, the layer defined by (g) and (h) is subsequently used after (a) and (b) to compute the action values.} \label{fig:architecture_detail}
\end{figure*}

Details of the neural network architectures used by GPL in Wolfpack and LBF are provided in Figure~\ref{fig:architecture_detail}. Before being processed by the LSTM, the type embedding network passes the input through two fully connected layers. Results from the embedding network are subsequently passed to the GPL component that has the type embedding as input. For the joint action value computation, the singular and pairwise utility computation utilizes an architecture provided in Figure~\ref{fig:b} and Figure~\ref{fig:c}. The agent and auxiliary agent models follows the architecture provided in Figure~\ref{fig:d}, Figure~\ref{fig:e}, and Figure~\ref{fig:f}.

To allow a fair comparison between the baselines and GPL, the model architecture used by baselines in Wolfpack and LBF follows the architecture used by GPL. Specifically, baselines pass their input vectors to the architecture in Figure~\ref{fig:a} and subsequently passes the output to an architecture following Figure~\ref{fig:b} to compute the action values. For baselines that use agent and auxiliary agent models, the architecture of these models follows the same architecture provided by Figure~\ref{fig:d}, Figure~\ref{fig:e}, and Figure~\ref{fig:f}.

For MADDPG and DGN, we use networks with similar sizes to those used in open ad hoc teamwork experiments. The only difference is we do not use LSTMs in the network since there is no need for type inference in the MARL approaches. As a result, the architecture used by DGN is simply the architecture used by GNN with the LSTM layers replaced with an MLP with the same output size. The decentralized policy for MADDPG is also similar with the value network architecture of QL with the LSTM replaced by an MLP with the same output size.

For training, we use the following hyperparameters for GPL and all proposed baselines:
\begin{itemize}
    \item $K=5$ for the low rank factorization of pairwise utility terms in GPL.
    \item 8 attention heads were used for DGN, GNN-QL and GNN-QL-AM.
    \item Data is collected from 16 parallel environments to collect a total experience of 6.4 million environment steps.
    \item Models are optimized using the Adam optimization algorithm with a learning rate of $2.5\times10^{-4}$. 
    \item Models are updated every 4 steps on the parallel environment.
    \item Instead of updating the target networks by periodically copying the joint action value network, target networks are updated using a weighted average of the parameters of the joint action value network. Assuming that $\phi$ is a parameter of the joint action value network, we update the corresponding parameter in the target network $\phi^{'}$ by using $\phi^{'} \leftarrow (1-\alpha)\phi^{'} + \alpha\phi$, with $\alpha$ set to $10^{-3}$.
    \item $\epsilon$ for the exploration policy is linearly annealed from 1 to 0.05 in the first 4.8 million environment steps and remains the same afterwards.
    \item We also use an attention weight regularization term, $\lambda$, of 0.03 for DGN and a temperature of 0.1 for MADDPG's gumbel softmax function.
\end{itemize}

These hyperparameters and network architectures are obtained by initially searching for a network and hyperparameter configuration which works for QL. After finding an architecture along with a set of hyperparameters which works best for QL, we train a similar sized architecture with the same hyperparameters for the rest of the baselines. Despite our lack of hyperparameter tuning for the rest of the baselines, we still obtain better performance than QL in almost all cases.

For FortAttack, we initially started the training process with the same hyperparameter setup as LBF and Wolfpack. Due to QL not learning, we decided to increase the size of the network along with running the training process for more timesteps to take into account of the increased complexity of the environment compared to Wolfpack and LBF. Nonetheless, we did not find any success in training QL regardless of the different network sizes and hyperparameters we tried.  

We subsequently focused on finding a network architecture along with hyperparameters that worked for GPL. A similar sized network with GPL along with GPL's training hyperparameters were used for training other baselines. This resulted in the following architecture and hyperparameters for FortAttack:
\begin{itemize}
    \item $K=6$ for the low rank factorization of pairwise utility terms in GPL.
    \item 8 attention heads were used for DGN, GNN-QL and GNN-QL-AM.
    \item Data was collected from 16 parallel environments to collect a total experience of 16 million environment steps.
    \item Models were optimized using the Adam optimization algorithm with a learning rate of $1.0\times10^{-4}$. 
    \item Models were updated every 4 steps on the parallel environment.
    \item Instead of updating the target networks by periodically copying the joint action value network, target networks were updated using a weighted average of the parameters of the joint action value network. Assuming that $\phi$ is a parameter of the joint action value network, we updated the corresponding parameter in the target network $\phi^{'}$ by using $\phi^{'} \leftarrow (1-\alpha)\phi^{'} + \alpha\phi$, with $\alpha$ set to $10^{-3}$.
    \item The exploration parameter $\epsilon$ was linearly annealed from 1 to 0.05 in the first 8 million environment steps and remains the same afterwards.
    \item 64 units were used in the fully connected and LSTM layer for $LSTM_{\alpha_{Q/q/p}}$. For comparison, in Wolfpack and LBF we used 100 units.
    \item 128 units were used for both fully connected layers in $MLP_{\beta}$ and $MLP_{\delta}$. For comparison, we used 70 and 60 units respectively in Wolfpack and LBF.
    \item 40 and 70 units were used respectively for the first and second layer in the edge and node processing network ($GNN_{\mu/\zeta}$) of the agent model. For comparison, we used 30 and 70 units respectively for Wolfpack and LBF.
    \item We used 128 units in $MLP_{\mu}$ and $MLP_{\nu}$. In comparison, we used 20 units for Wolfpack and LBF.
    \item Finally, we used 128 units for both fully connected layers involved in the key, query, and value computation under GNN-QL and GNN-QL-AM. For comparison, we used 70 and 60 units respectively for Wolfpack and LBF.

\end{itemize}

Finally the way these components were assembled into the architectures of the baselines was still the same between FortAttack and the other environments.

\section{Additional MARL Training Results}
\label{sec:MARLJointTrainRes}
In this section, we provide additional results from the MARL algorithms we used in our open ad hoc training when interacting against teammates that are jointly trained with itself. 

\subsection{MARL training performance against jointly trained teammates}
\label{subsec:MARLTrain}
\begin{figure*}[!htb]
\begin{subfigure}{0.33\textwidth}
  \centering
  \includegraphics[scale=.33]{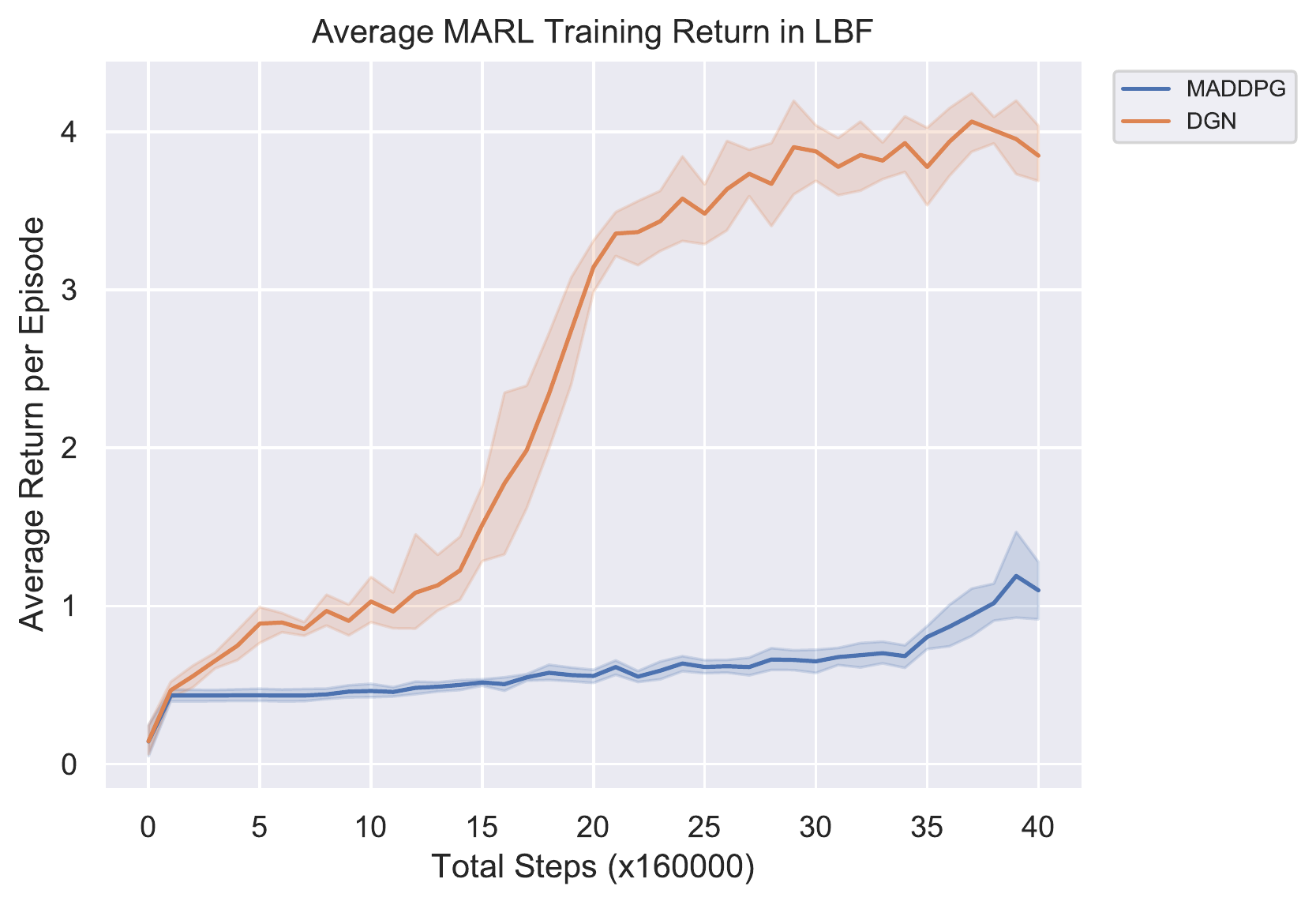}
  \caption{Results in Level-based foraging}
  \label{fig:MARLlbforagingopen}
\end{subfigure}
\begin{subfigure}{0.33\textwidth}
  \centering
  \includegraphics[scale=.33]{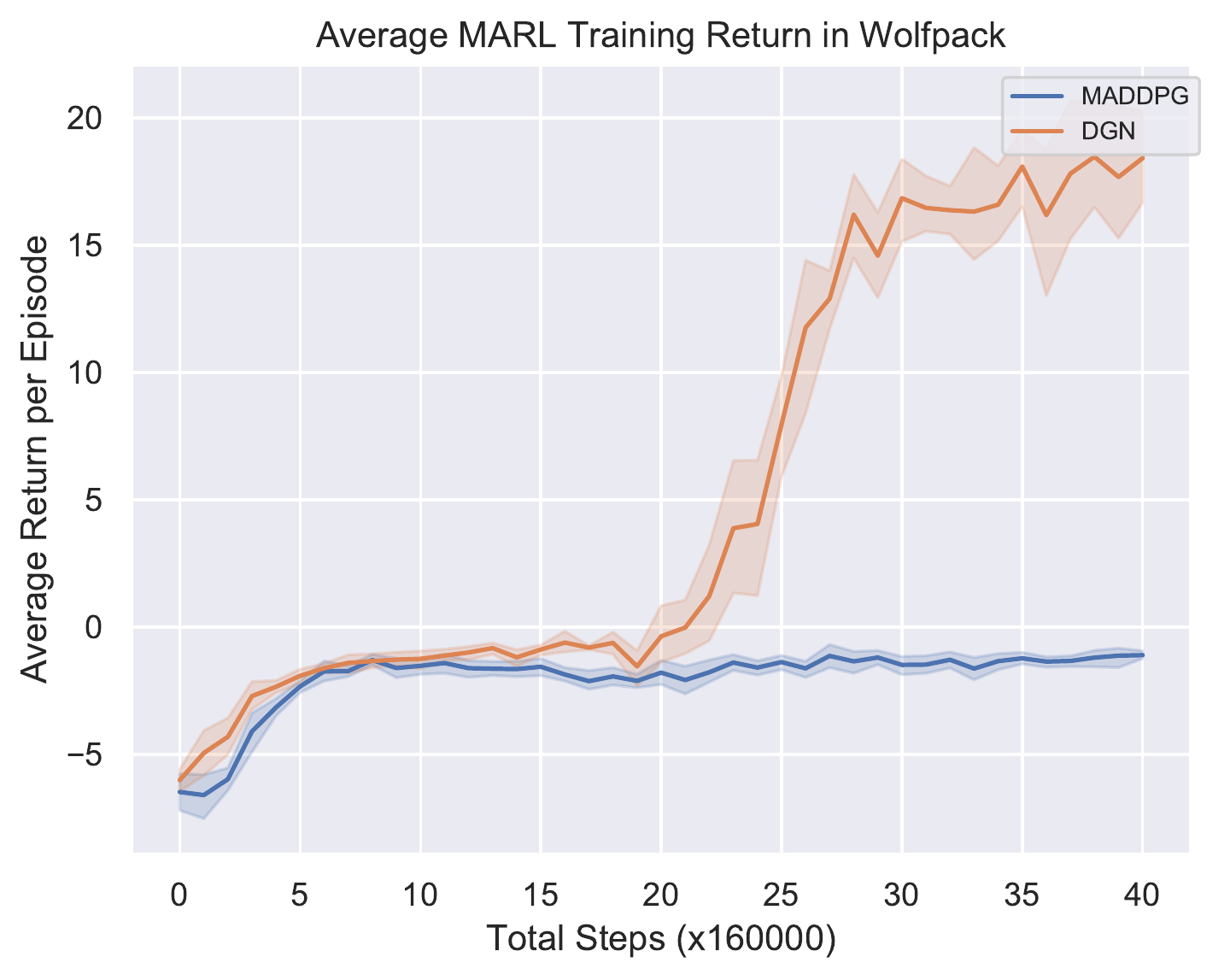}  
  \caption{Results in Wolfpack}
  \label{fig:MARLwolf05open}
\end{subfigure}
\begin{subfigure}{0.33\textwidth}
  \centering
  \includegraphics[scale=.33]{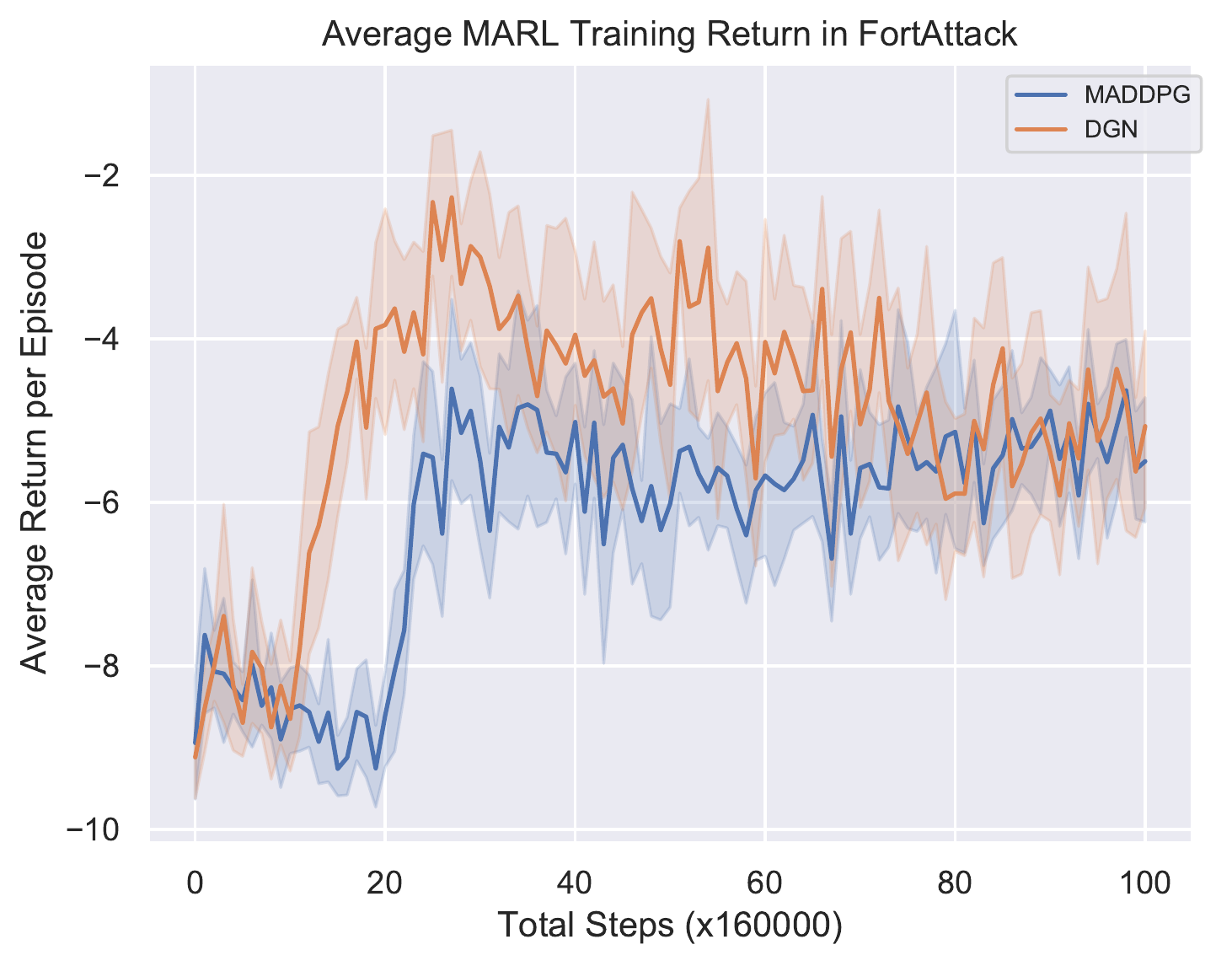}
  \caption{Results in FortAttack}
  \label{fig:MARLfortattackopen}
\end{subfigure}
\caption{\textbf{Open MARL results (training) against jointly trained agents:} Average and 95\% confidence bounds of MADDPG and DGN returns during training (up to 3 agents in a team for LBF, Wolfpack, and attacker \& defender teams in FortAttack). For each algorithm, training is done using eight different seeds and the resulting models are saved and evaluated every 160000 global steps. Compared to Figure~\ref{OpenTrainingRes}, we see that the MARL baselines achieved better performance when interacting against jointly trained teammates than when they interacted with the unknown policies during open ad hoc teamwork.}
\label{fig:MARLOpenTrainingRes}
\end{figure*}

Following training that we did to create the MARL policies we evaluated in Section~\ref{sec:exp_adhoc}, we evaluate the MARL policies in an open process that is similar to the open process used in the training setup in Section~\ref{sec:exp_adhoc}. The only difference with the open process used in the training setup in Section~\ref{sec:exp_adhoc} is that the sampled teammates are always agents that are jointly trained with the MARL agent. From the results provided in Figure~\ref{fig:MARLOpenTrainingRes}, we see that the MARL baselines achieved better performances than when they interact against our ad hoc teamwork teammates.

\subsection{MARL generalization performance against jointly trained teammates}
In this section, we describe the performance of the MARL policies when the open process in Section~\ref{subsec:MARLTrain} is changed such that the upper limit on team size is five agents for LBF, Wolfpack, and attacker \& defender teams in FortAttack. The results are provided in Table~\ref{t:MARLEvalJointTrain}.
\begin{table}[!htb]
    \small
    \center
    \begin{tabular}{|c|c|c|}
        \hline
        Environment & MADDPG  & DGN \\
        \hline
        LBF  & 0.64 $\pm$ 0.15 & 1.65 $\pm$ 0.25 \\ \hline
        Wolfpack & -1.15 $\pm$ 0.20 & 9.36 $\pm$ 2.46 \\ \hline  
        FortAttack & 0.03 $\pm$ 1.03 & -8.13 $\pm$ 0.98  \\ \hline  
    \end{tabular}
    \caption{\textbf{Open MARL results (generalization) against jointly trained agents:} Average and 95\% confidence bounds of MADDPG and DGN returns during generalization (up to 5 agents in a team for LBF, Wolfpack, and attacker \& defender teams in FortAttack). For each algorithm, training is done using eight different seeds and the resulting models are saved and evaluated every 160000 global steps. Among all saved policies, we choose the checkpoint that has the highest average performance during training (Section~\ref{subsec:MARLTrain}) and report the mean and 95\% confidence bounds of the performance of the policies at that checkpoint.}
    \label{t:MARLEvalJointTrain}
    \vspace{-10pt}
\end{table}

\section{Team size \& Fixed type generalization results}
\label{TSgeneral}
To further demonstrate that GPL outperforms the single-agent baselines in open ad hoc teamwork, we train GPL under different open processes than what we used in our main experiment in Section~\ref{sec:exp_adhoc}. We exclude MARL baselines from these experiments since our main experiment shows that they consistently perform worse than our worst performing single-agent RL baselines. For the first experiment, we train agents in LBF, Wolfpack, and FortAttack under an open process where the learner's team consists of two agents during training. 

The other experiment trains agents in FortAttack under two open processes where the types of teammates remains fixed during interaction. In the first open process, defender teammate type is fixed towards guard type 4 mentioned in Section~\ref{sec:add_Details_fortattack}. By contrast, the second open process sets the defender types as guard type 6. The attackers in both open processes are configured to have attacker type 2.

\begin{figure*}[t]
\begin{subfigure}{.33\textwidth}
  \centering
  \includegraphics[scale=.35]{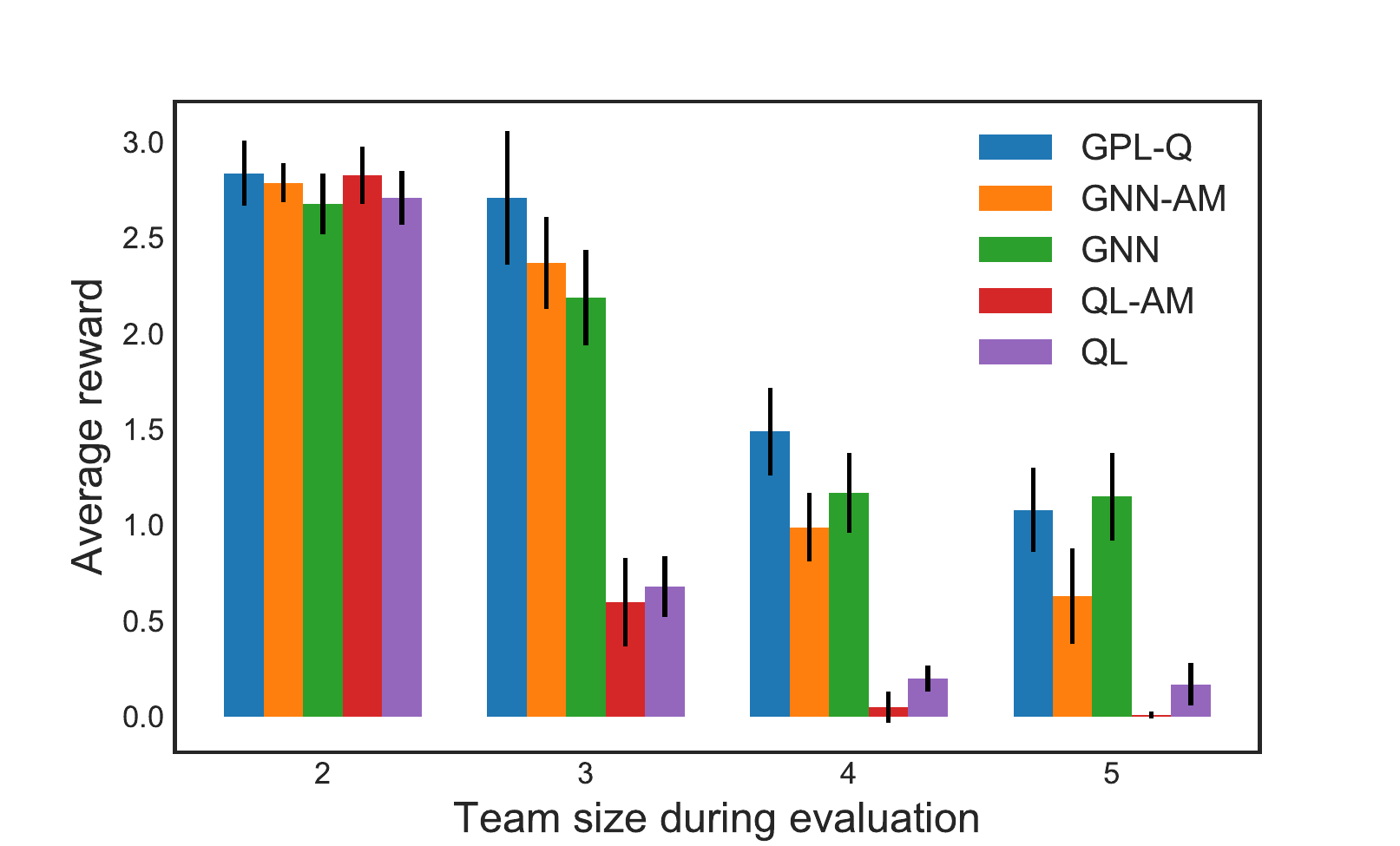}
  \caption{Results in level-based foraging}
  \label{lbforagingopen5}
\end{subfigure}
\begin{subfigure}{.33\textwidth}
  \centering
  \includegraphics[scale=.35]{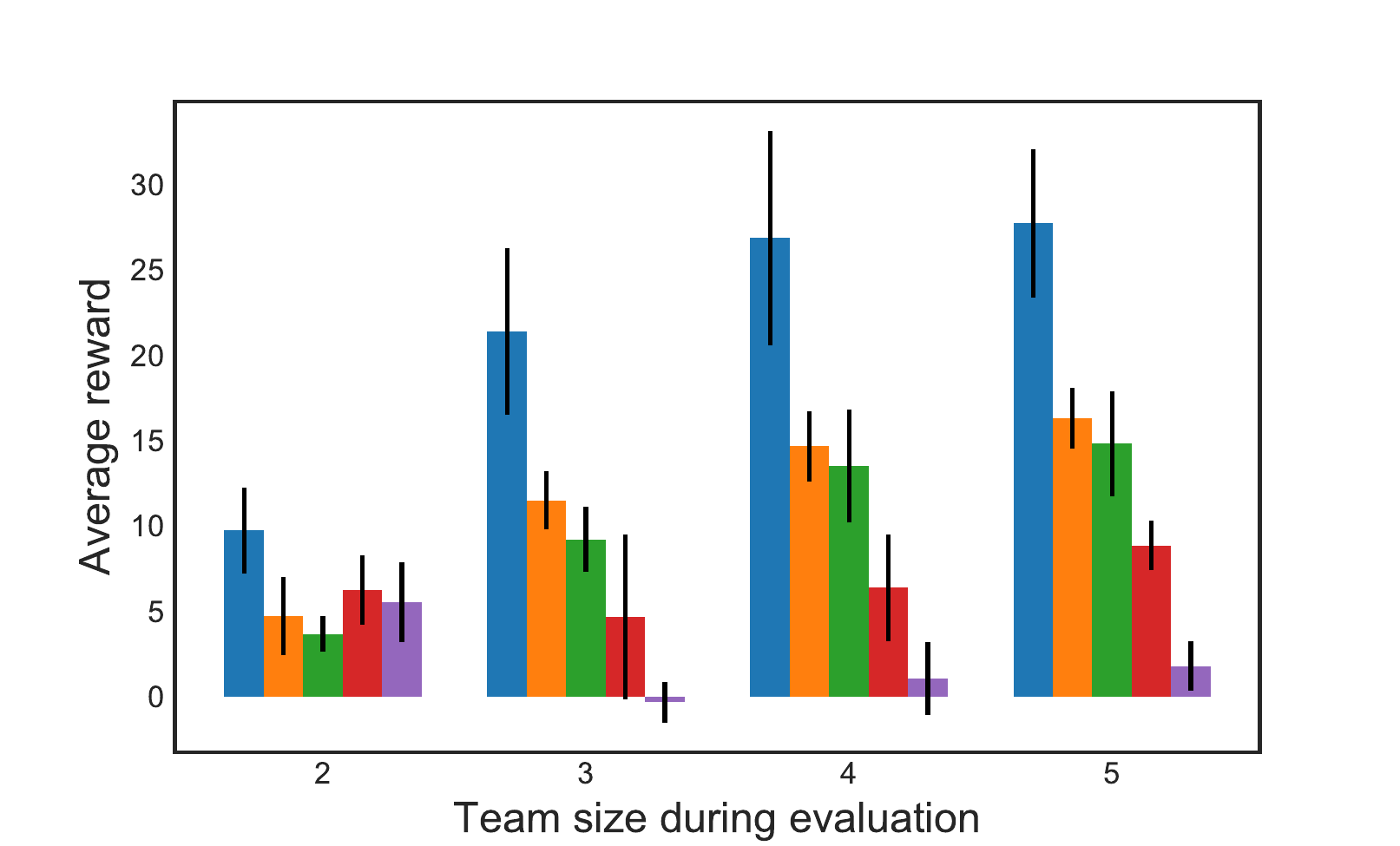}  
  \caption{Results in Wolfpack with 0.5 penalty}
  \label{wolf05open5}
\end{subfigure}
\begin{subfigure}{.33\textwidth}
  \centering
  \includegraphics[scale=.35]{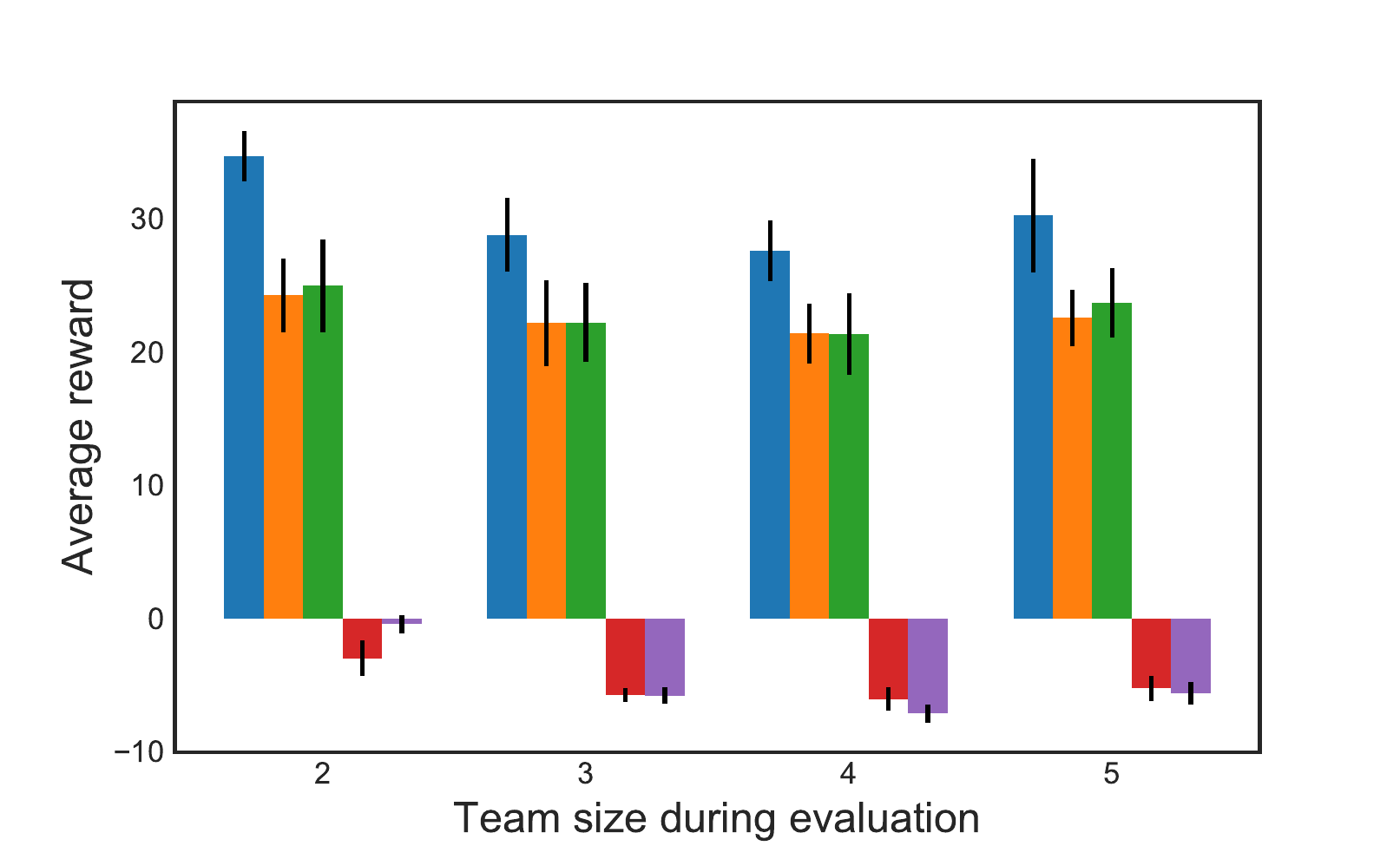}
  \caption{Results in FortAttack}
  \label{FortAttackopen5}
\end{subfigure}
\caption{\textbf{Team size generalization results:} Average rewards and 95\% confidence bounds of GPL and baselines from team size generalization experiments collected over eight seeds. Agent was trained to interact in a team of two agents for LBF, Wolfpack, and FortAttack. With FortAttack, we used a setup where the number of attackers was always equal to the number of teammate defenders. Value networks are stored every 160000 steps and the performance of greedy policies from value networks stored at the checkpoint with the highest average performance during training were used to compute the average returns and their 95\% confidence interval. This result shows that GPL-Q still outperforms other single-agent RL baselines, except for LBF where its performance is not significantly different from GNN and GNN-AM.}
\label{ClosedTrainingRes5}
\end{figure*}

\begin{figure*}[t]
\begin{subfigure}{.5\textwidth}
  \centering
  \includegraphics[scale=.35]{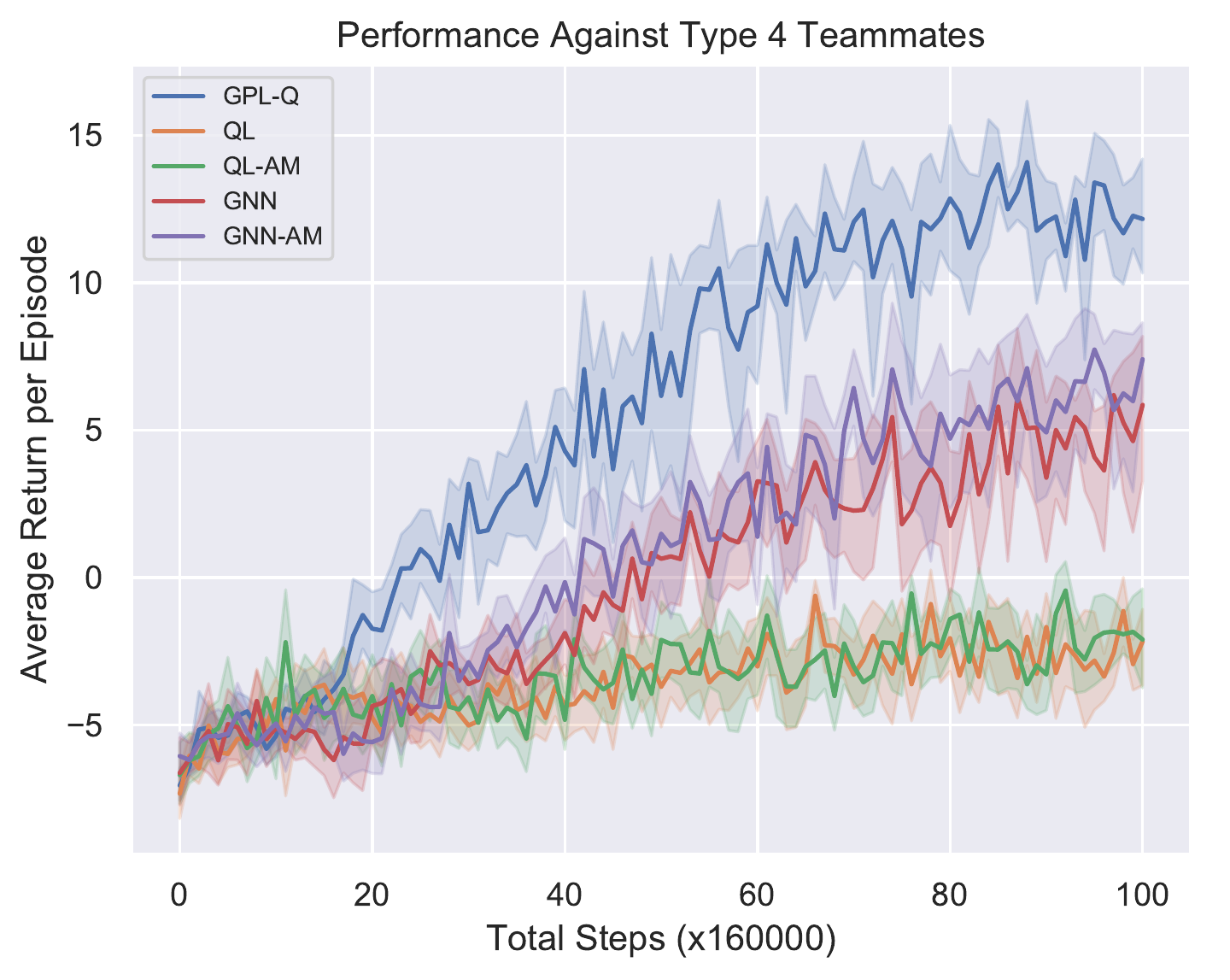}
  \caption{Performance against Type 4 teammates in FortAttack.}
  \label{FortAttackTeammateType5}
\end{subfigure}
\begin{subfigure}{.33\textwidth}
  \centering
  \includegraphics[scale=.35]{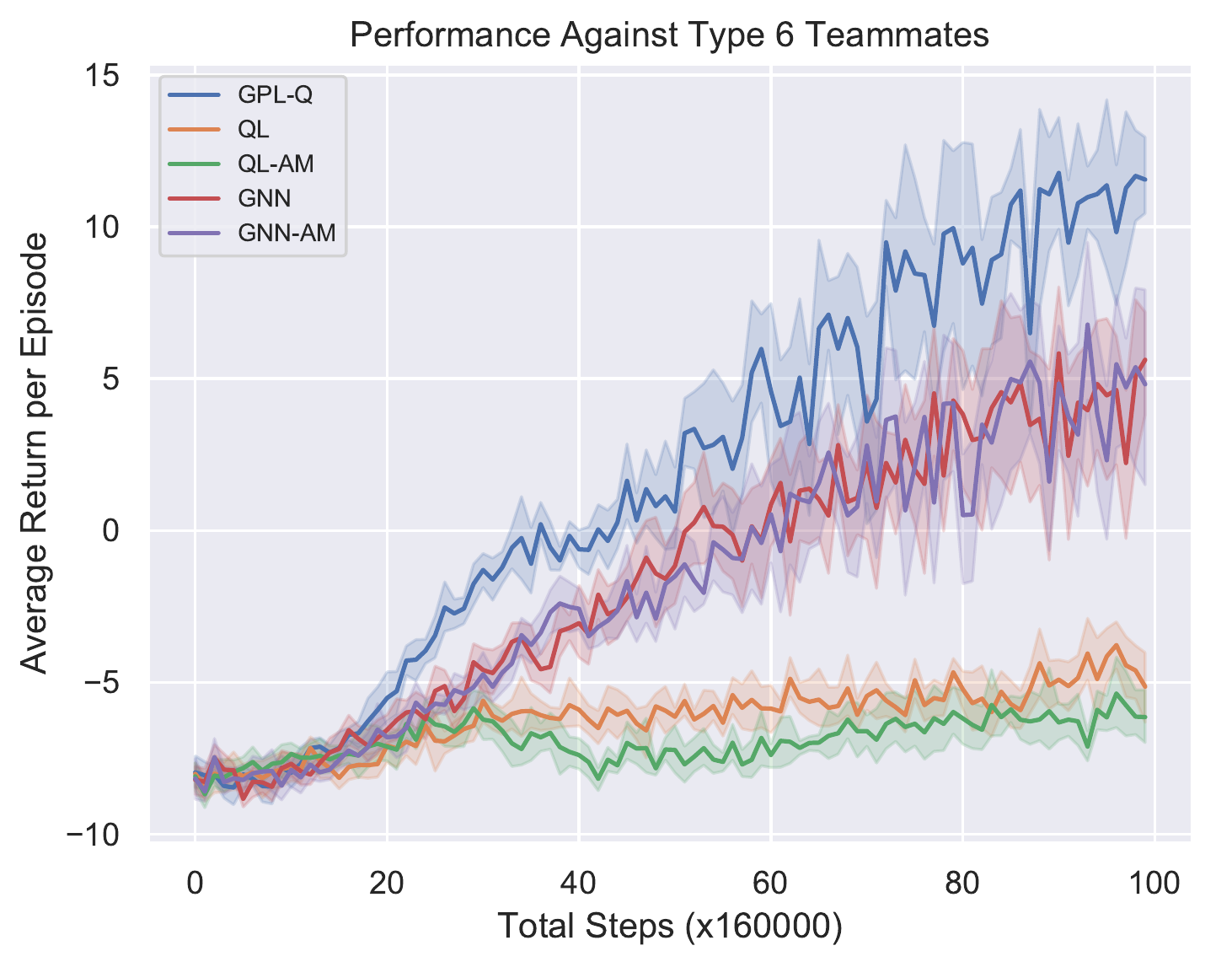}  
  \caption{Performance against type 6 teammates in FortAttack.}
  \label{FortAttackTeammateType6}
\end{subfigure}
\caption{\textbf{Fixed type experiment results:} Average rewards and 95\% confidence bounds of GPL-Q and baselines from fixed type experiments collected over eight seeds in FortAttack. Agent was trained to interact in a team of up to 3 agents with all teammates having a fixed type. Value networks are stored every 160000 steps and the performance of greedy policies from value networks stored at the checkpoint with the highest average performance during training were used to compute the average returns and their 95\% confidence interval. The results illustrates the performance of the methods when trained against (a) type 4 teammates and (b) type 6 teammates.}
\label{FixedTypeTraining}
\end{figure*}

Figure~\ref{ClosedTrainingRes5} shows the results for the fixed team size experiments. On the other hand, Figure~\ref{FixedTypeTraining} provides the resulting performance from training the learner against teammates of fixed types. The results further demonstrates GPL's superior performance to baselines for open ad hoc teamwork.

\section{Wolfpack action value analysis}
\label{sec:WolfpackAVAnalysis}
We provide an analysis of the individual and joint action values learned by GPL in Wolfpack. For this analysis, we still use $\bar{Q}_{j}$ as defined in Section~\ref{GPLUtilAnalysis}. On the other hand, since the Wolfpack environment has no shooting actions, we replace $\bar{Q}_{j,k}$ with $C_{j,k}(a)$ that we define in Equation~\ref{CJK}. 
\begin{align}
    \label{CJK}
    C_{j,k}(a) &= Q^{j,k}_{i}(a^{j}, a^{k}|s) - N_{j,k}(a)\\
    N_{j,k}(a) &= \dfrac{\sum_{x,y\in{A}, x\neq{a_{j}}, y\neq{a_{k}}}Q^{j,k}_{i}(x, y|s)}{|A|^{2}-1}
\end{align}

$C_{j,k}(a)$ denotes the difference between the pairwise action value associated to the action chosen by the agents compared to the average pairwise action value of other pairs of actions it could have taken. In our analysis, we only investigate $C_{j,k}(a)$ for observations that precede the capture of a prey by the learning agent. We further filter out values $C_{j,k}(a)$ for which $j$ and $k$ do not belong to the pack of wolves that successfully captured a prey together with the learning agent. By investigating this, we wanted to see whether GPL assigns a higher pairwise action value towards pairs of actions that lead to a joint prey capture.

\begin{figure*}[h]
\centering
\begin{subfigure}{.45\textwidth}
  \centering
  \includegraphics[trim={0.0cm 0.0cm 0.0 0cm},clip,scale=.4]{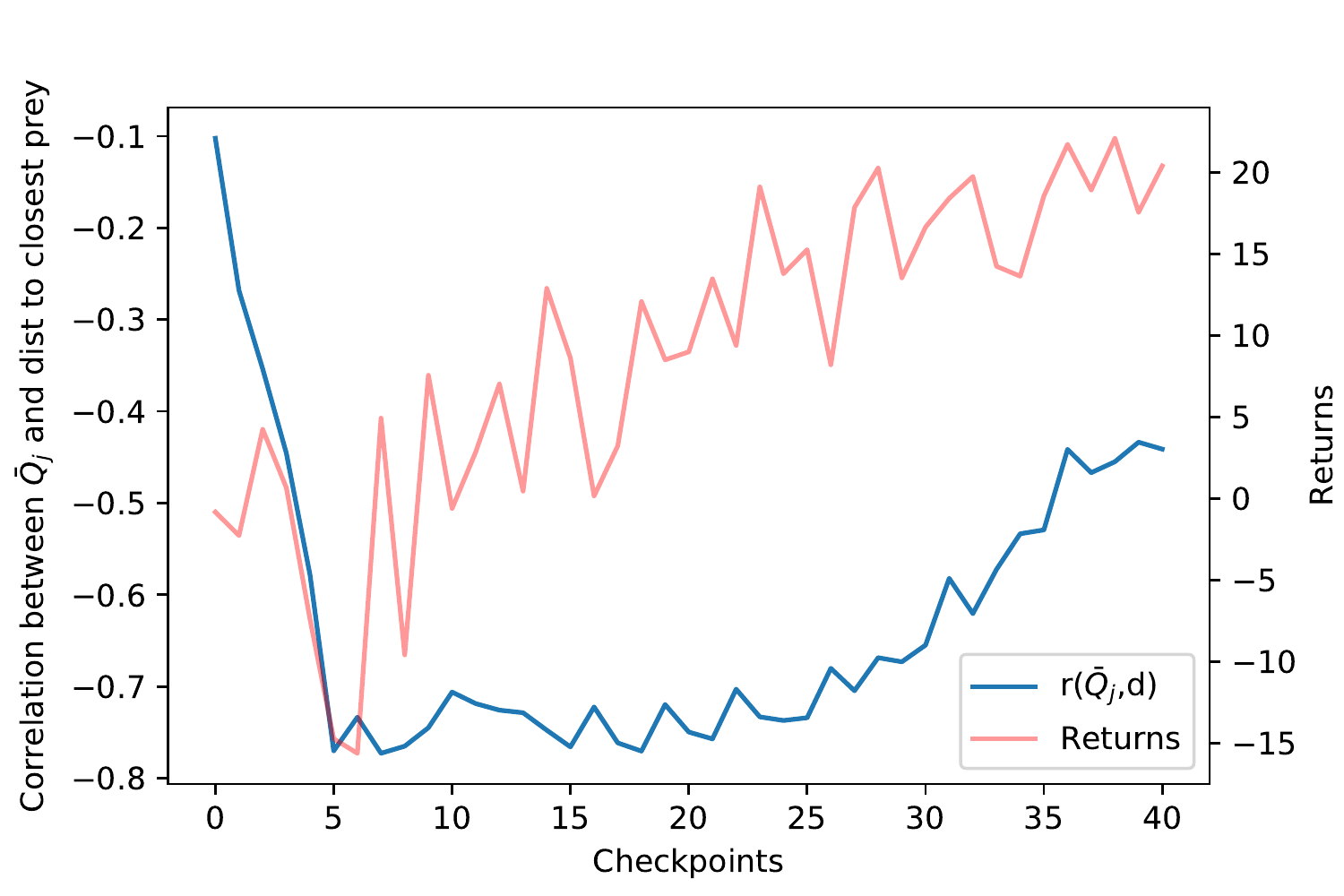}
  \caption{Correlation between a teammate's $\bar{Q}_{j}$ with their distance to the closest prey from the learning agent.}
  \label{ShootPairwiseVis5}
\end{subfigure}
\hfill
\begin{subfigure}{.45\textwidth}
  \centering
  \includegraphics[trim={0.0cm 0.0cm 0 0cm},clip,scale=.4]{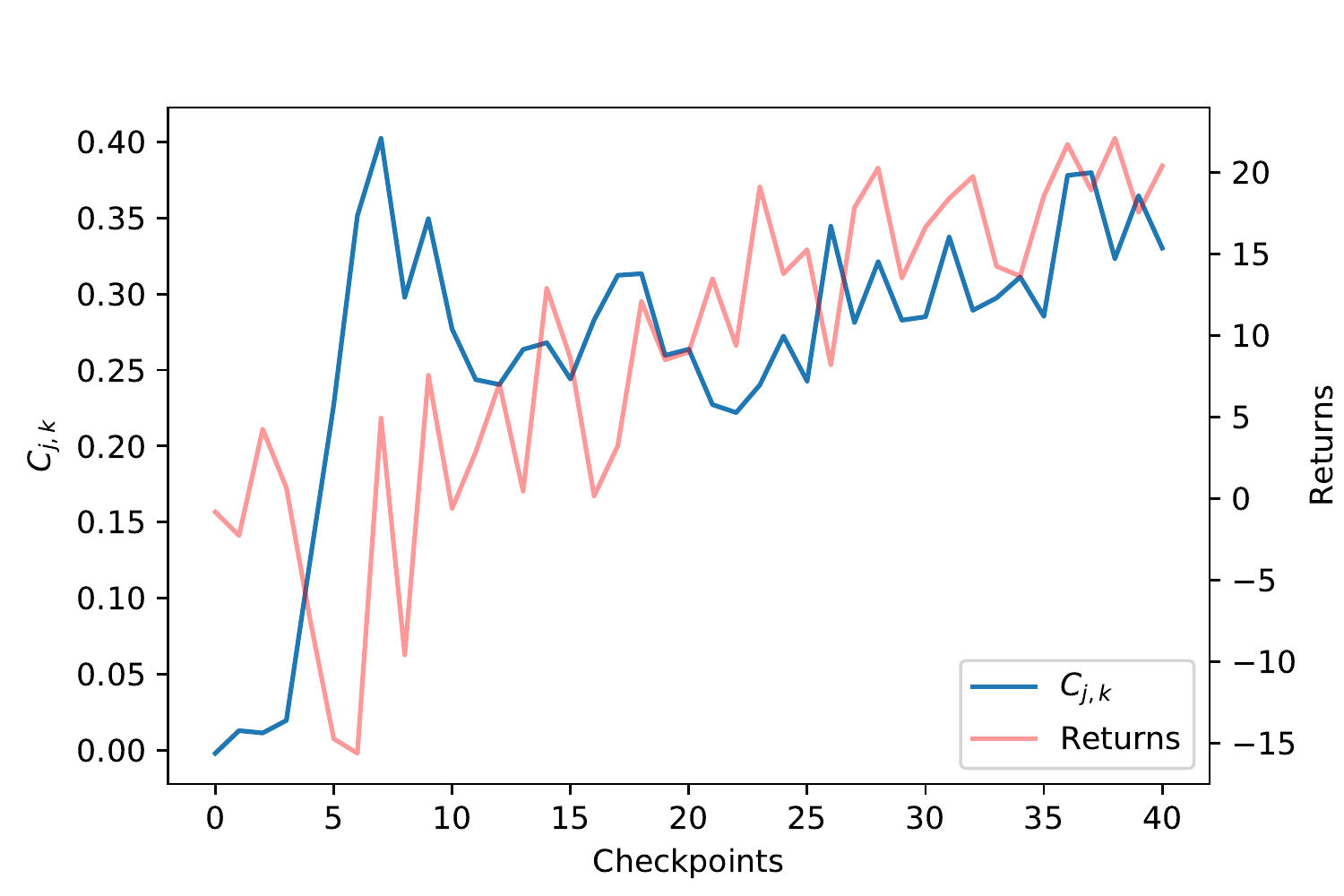}  
  \caption{Average value of $C_{j,k}(a)$ for situations where agent $j$, $k$, and learning agent's actions leads to the capture of a prey.}
  \label{FeatureImportance5}
\end{subfigure}
\caption{\textbf{FortAttack individual and pairwise action value visualizations: } The visualizations show how various metrics evolve across different checkpoints and correlate with agent's learning performance. Data is obtained by executing the greedy policy entailed by the value network for 24000 timesteps. (\ref{ShootPairwiseVis5}) The Pearson correlation coefficient between a teammate's distance to the prey closest from the learning agent and $\bar{Q}_j$ is gathered and visualized. (\ref{FeatureImportance5}) $C_{j,k}(a)$ is computed for observations preceding a capture of a prey by the learning agent. Analysis is further limited towards the edges associated to pairs of edges connecting agents that hunted the prey together with the learning agent.}
\label{WolfpackUtilAnalysis}
\end{figure*}

The result of our analysis is provided in Figure~\ref{WolfpackUtilAnalysis}. We find that GPL assigns higher individual action values for a teammate as they get closer towards the prey that the learning agent is hunting. This is a reasonable value assignment since teammates can better help the learning agent hunt as they get closer to the target prey. On the other hand, we also see that GPL progressively learns to increase the difference in pairwise action values between pairs of actions that lead to a capture and other alternative pairs of actions. Therefore, GPL helps the learning agent to understand the positive contribution resulting from pairs of actions leading towards a capture.

\section{Baseline action-value analysis}
\label{sec:BaselineActValAnalysis}
    In this section, we analyze the action-value estimates produced by the single-agent RL baselines used in our experiments. Despite recognizing the value of shooting, we show that the baselines do not learn the effects of other agents' action towards the learner as GPL does. This results in the significant performance gap between GPL and the baselines in our experiments.
    
\begin{figure*}[!htb]
        \centering
        \begin{subfigure}[b]{0.475\textwidth}
            \centering
            \includegraphics[width=\textwidth]{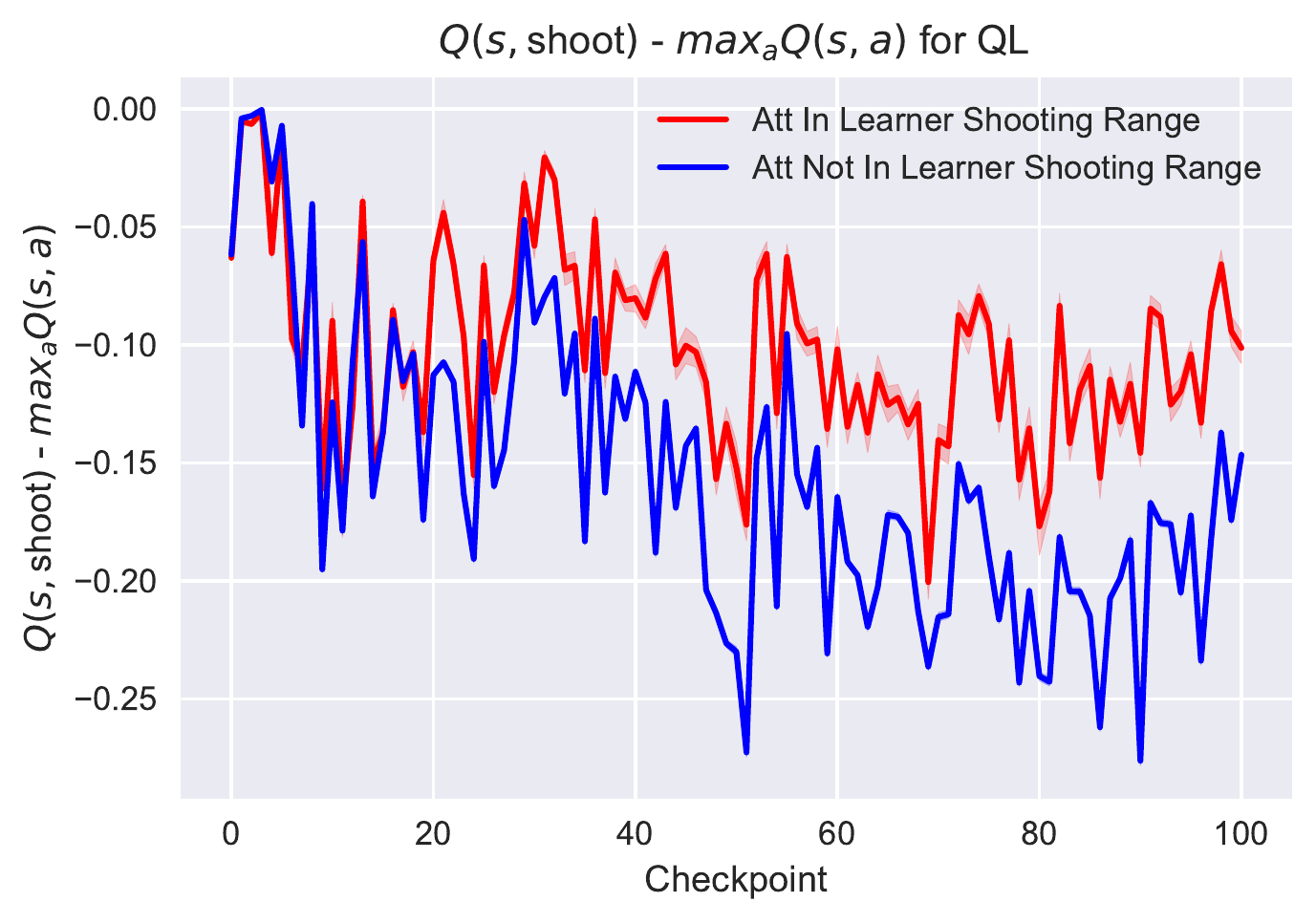}
            \caption[Network2]%
            {{\small $Q(s,\text{shoot}) - \text{max}_{a}Q(s,a)$ for QL.}}    
            \label{fig:DiffQQL}
        \end{subfigure}
        \hfill
        \begin{subfigure}[b]{0.475\textwidth}  
            \centering 
            \includegraphics[width=\textwidth]{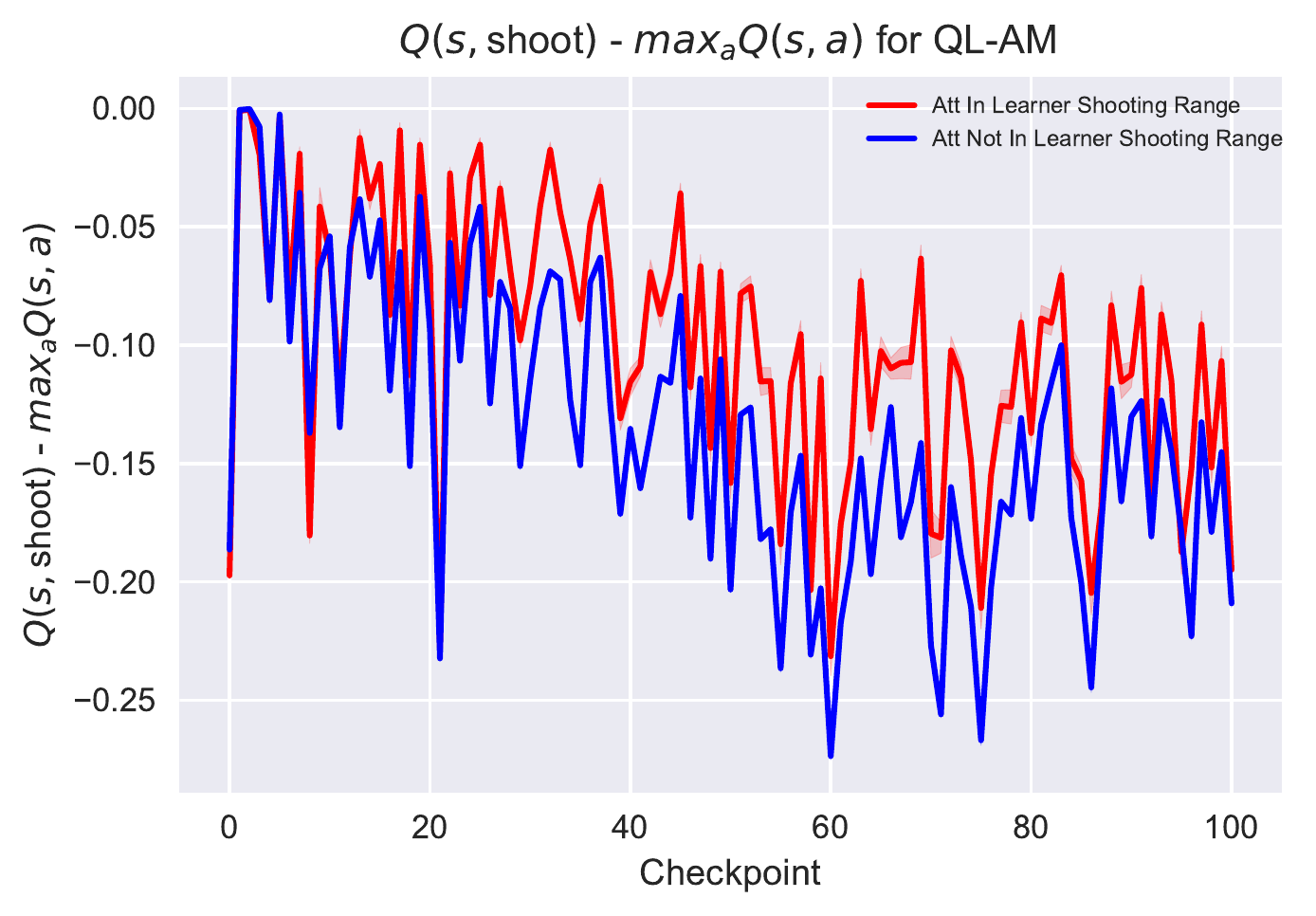}
            \caption[]%
            {{\small $Q(s,\text{shoot}) - \text{max}_{a}Q(s,a)$ for QL-AM.}}    
            \label{fig:DiffQQLAM}
        \end{subfigure}
        \vskip\baselineskip
        \begin{subfigure}[b]{0.475\textwidth}   
            \centering 
            \includegraphics[width=\textwidth]{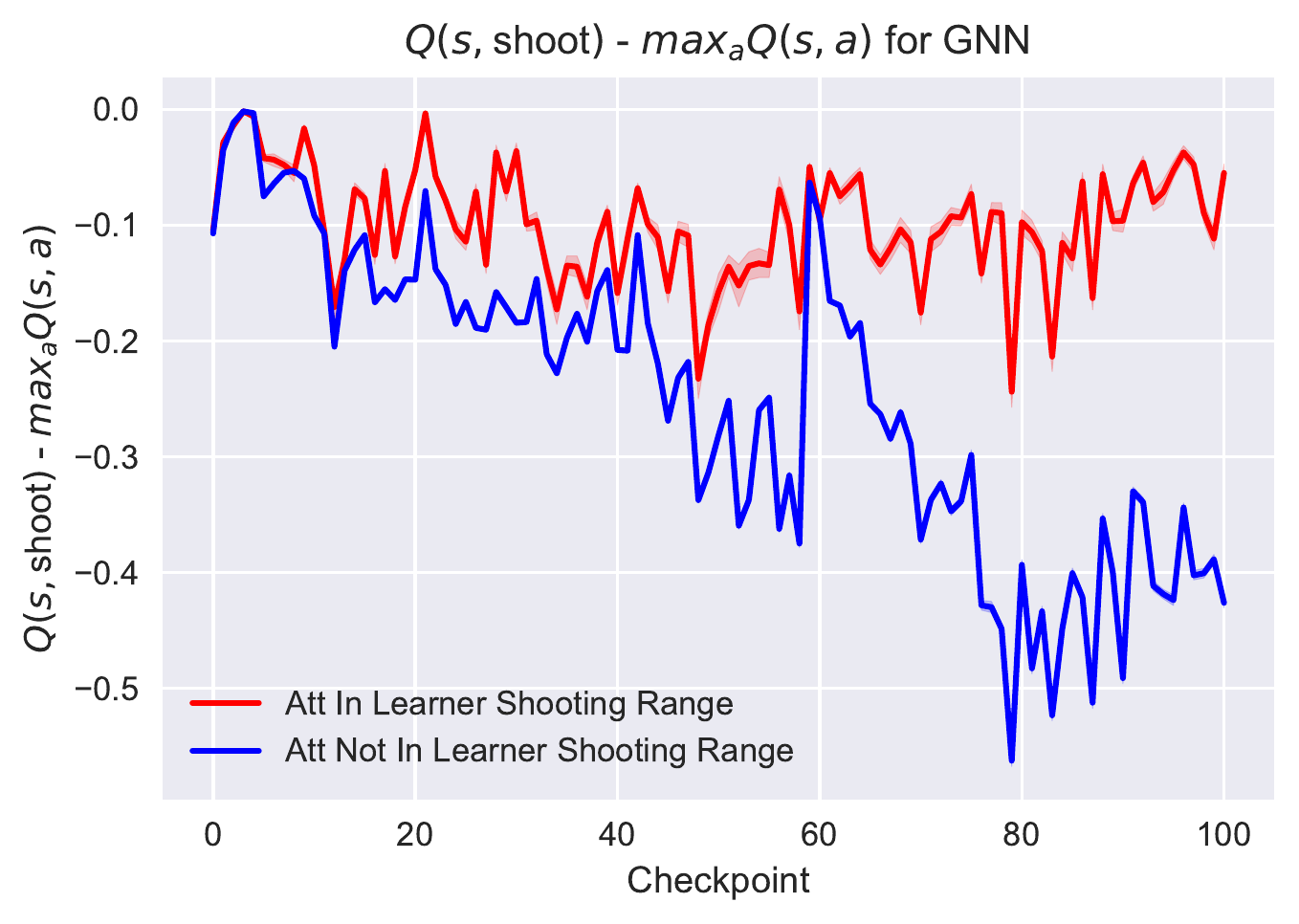}
            \caption[]%
            {{\small $Q(s,\text{shoot}) - \text{max}_{a}Q(s,a)$ for GNN.}}    
            \label{fig:DiffQGNN}
        \end{subfigure}
        \hfill
        \begin{subfigure}[b]{0.475\textwidth}   
            \centering 
            \includegraphics[width=\textwidth]{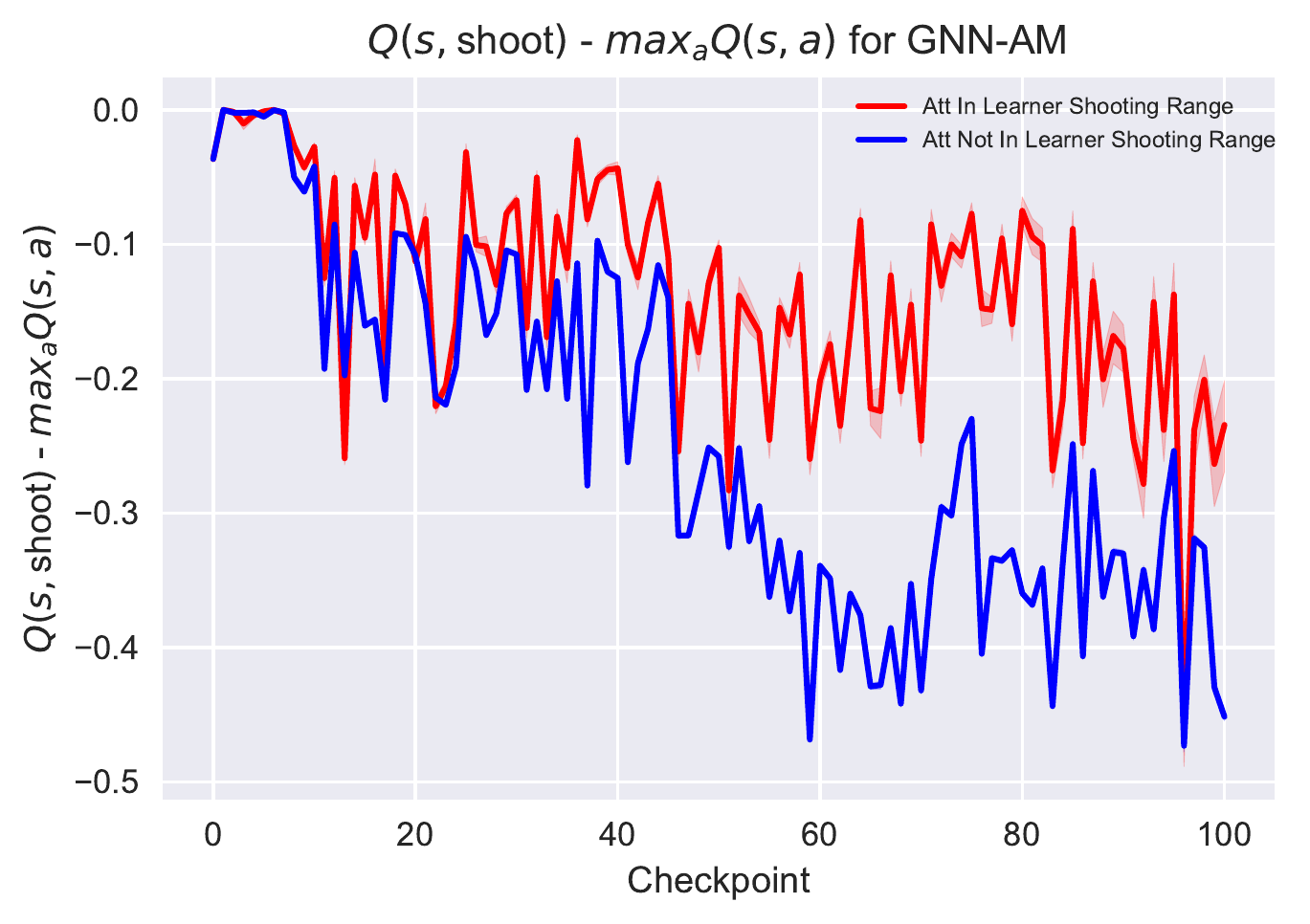}
            \caption[]%
            {{\small $Q(s,\text{shoot}) - \text{max}_{a}Q(s,a)$ for GNN-AM.}}    
            \label{fig:DiffQGNNAM}
        \end{subfigure}
        \caption[ The average and standard deviation of critical parameters ]
        {\small \textbf{$\mathbf{Q(s,}\text{shoot}) - \text{max}_{\mathbf{a}}\mathbf{Q(s,a)}$ for single-agent RL baselines:} This visualization compares the action-value of shooting and the estimated optimal action for (a) QL, (b) QL-AM, (c) GNN, and (d) GNN-AM. We obtain this figure by running a policy entailed by the value-networks that we saved during open ad hoc teamwork training done in Section~\ref{sec:exp_adhoc}. For each checkpoint, we gather data by running the policy for 480000 steps and recording the observed states. The observed states is subsequently organized based on whether there is at least an attacker is in the learner's shooting range or not. We finally measure $Q(s,\text{shoot}) - \text{max}_{a}Q(s,a)$ at these states and visualize its mean and 95\% confidence intervals. The blue line corresponds to states where there is no attacker in the learner's shooting range while the red line represents states where there is at least an attacker in the learner's shooting range. As training commences for all algorithms, this visualization shows that shooting becomes increasingly likely of becoming the optimal action whenever an attacker is inside the learner's shooting range.} 
        \label{fig:DiffQAnalysis}
    \end{figure*}

    Following our analysis in Section~\ref{GPLUtilAnalysis}, we limit the baselines' action-value analysis to the FortAttack environment. Our first analysis is done by comparing the difference between the action-value of shooting, $Q(s,\text{shoot})$, with the maximum action-value of all possible actions, $\text{max}_{a}Q(s, a)$, at different states. When this difference is closer to zero, it means that the agent is more likely to choose to shoot at $s$. As provided by Figure~\ref{fig:DiffQAnalysis}, we subsequently compare this value between states where no attacker is in the learner's shooting range and states where at least one attacker is in the learner's shooting range. 
    
    As learning progresses, our results show that all algorithms results in the learner becoming increasingly aware of the value of shooting attackers. For GNN, the gap between the red and blue lines become increasingly large around the $50^{th}$ checkpoint, which coincides with when GNN and GNN-AM agents start to increase their performance. On the other hand, although there is still a significant difference between the values of the red and blue line, QL and QL-AM does not result in a large difference to these lines compared to GNN/GNN-AM. This inability to further highlight the value of shooting is then the reason why QL/QL-AM does not perform as good as GNN/GNN-AM.
    
    \begin{figure*}[]
        \centering
        \begin{subfigure}[b]{0.475\textwidth}
            \centering
            \includegraphics[width=\textwidth]{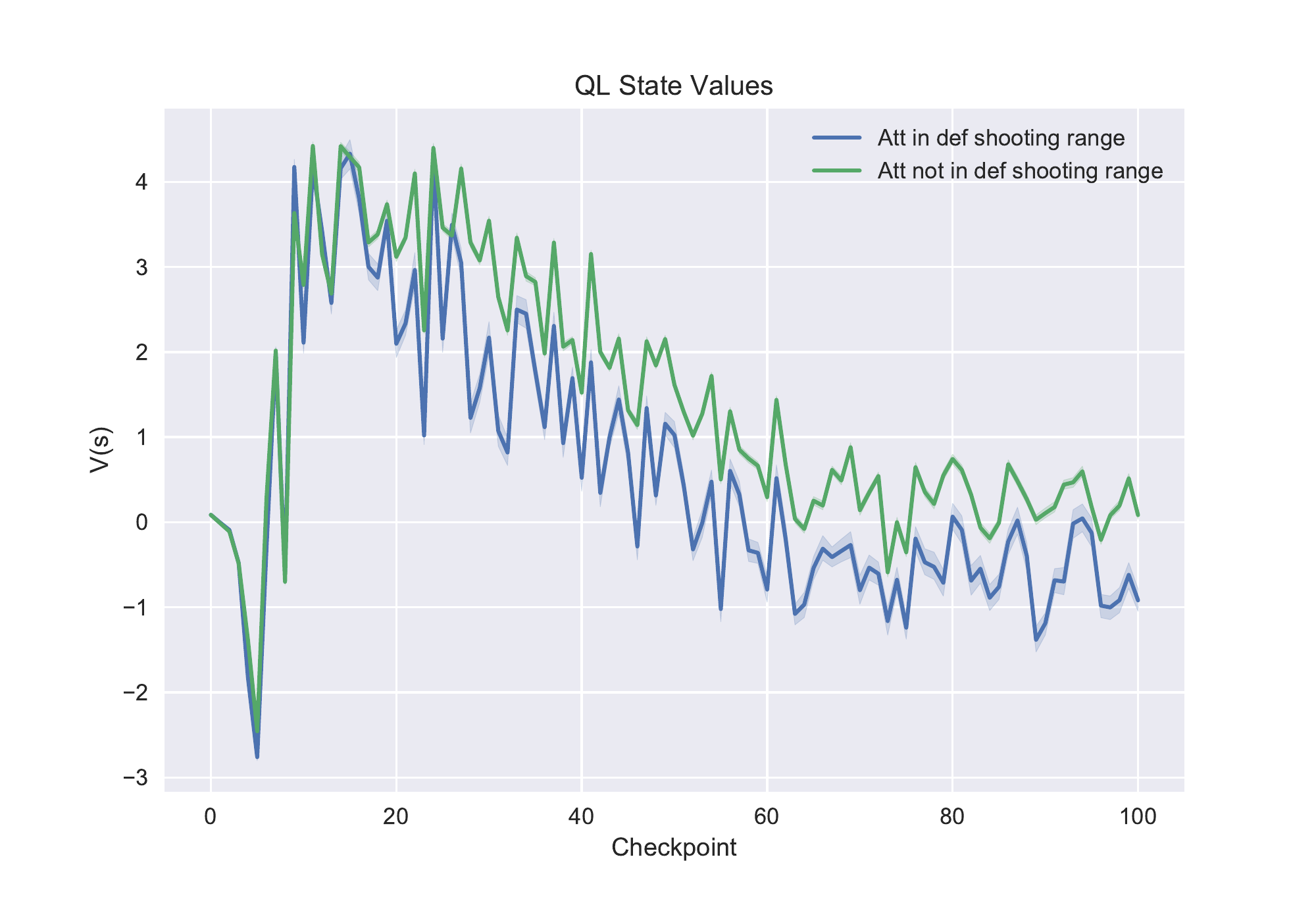}
            \caption[Network2]%
            {{\small State values for QL.}}    
            \label{fig:mean and std of net14}
        \end{subfigure}
        \hfill
        \begin{subfigure}[b]{0.475\textwidth}  
            \centering 
            \includegraphics[width=\textwidth]{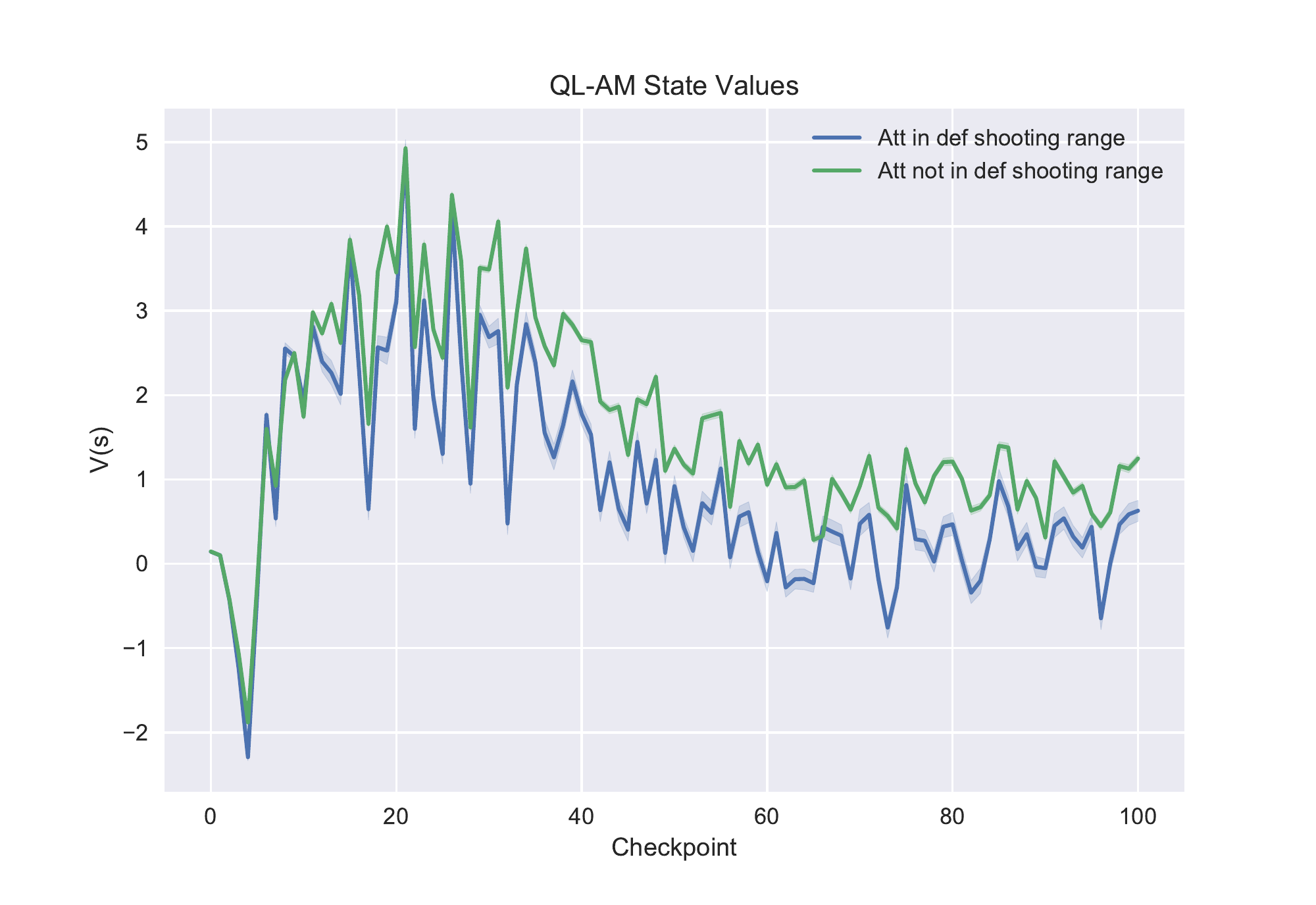}
            \caption[]%
            {{\small State values for QL-AM}}    
            \label{fig:mean and std of net24}
        \end{subfigure}
        \vskip\baselineskip
        \begin{subfigure}[b]{0.475\textwidth}   
            \centering 
            \includegraphics[width=\textwidth]{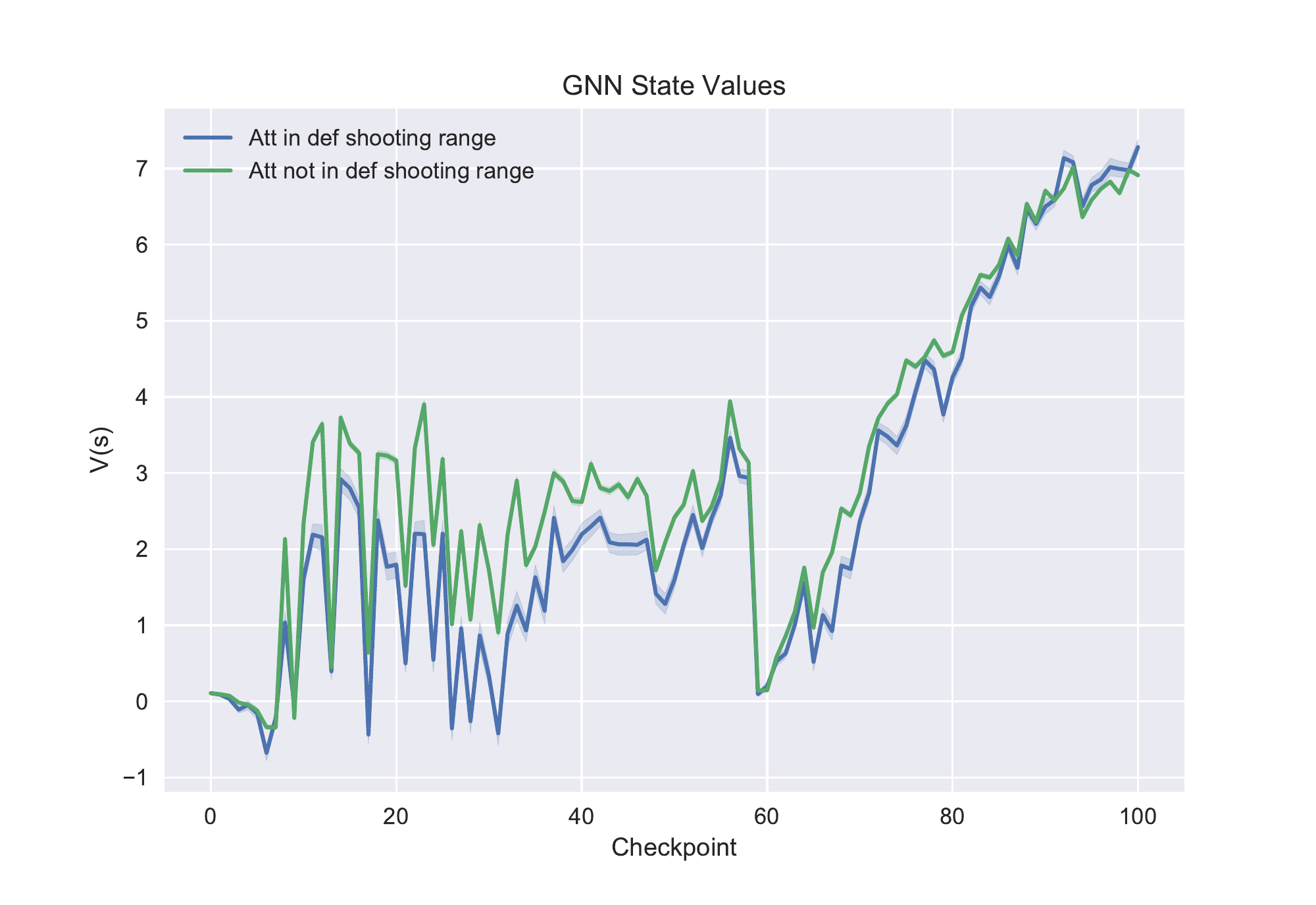}
            \caption[]%
            {{\small State values for GNN}}    
            \label{fig:mean and std of net34}
        \end{subfigure}
        \hfill
        \begin{subfigure}[b]{0.475\textwidth}   
            \centering 
            \includegraphics[width=\textwidth]{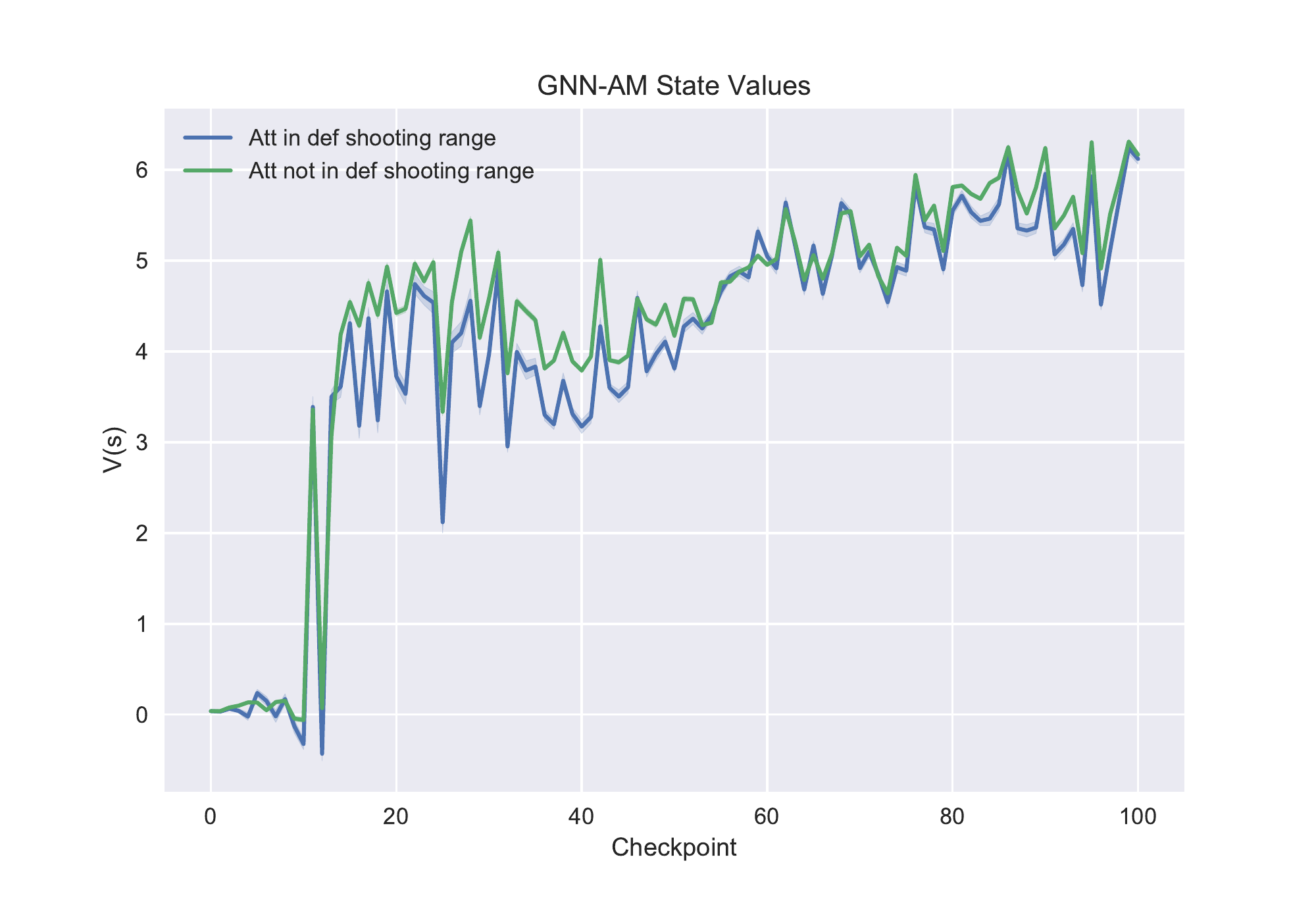}
            \caption[]%
            {{\small State values for GNN-AM}}    
            \label{fig:mean and std of net44}
        \end{subfigure}
        \caption[ The average and standard deviation of critical parameters ]
        {\small \textbf{State values for all single-agent RL baselines: }This visualization compares the state values for (a) QL, (b) QL-AM, (c) GNN, and (d) GNN-AM. This figure is obtained by running the policies that are stored during the open ad hoc teamwork training described in Section~\ref{sec:exp_adhoc}. Data collection is done by running these policies for 480000 steps and recording the observed states and predicted state-values throughout the process. To get a Monte Carlo estimate of the values of states when at least an attacker is in the defender's shooting range and when there are no attackers in any defender's shooting range, we average the recorded state values for these two situations. The blue line shows the average and 95\% confidence bounds of $V(s)$ when at least an attacker is in a defender's shooting range. By contrast, the green line shows the average and 95\% confidence bounds of $V(s)$ when no attacker is in any defender's shooting range. This figure shows that either the baselines assigns higher values to states where no attacker is in any defender's shooting range (QL and QL-AM) or the baselines most often fails to distinguish the value of both types of states (GNN and GNN-AM).}
        \label{fig:BaseVAnalysis}
    \end{figure*}
    
    We now show that despite learning the value of shooting attackers that are inside the learner's range, the single-agent RL baselines do not learn concepts that are learned by GPL's pairwise utility model, $MLP_{\delta}$. Since all baselines do not have a CG-based joint action-value model, it is difficult to measure $\bar{Q}_{j,k}$ as in Section~\ref{GPLUtilAnalysis}. The closest metric to $\bar{Q}_{j,k}$ that is still obtainable from the value networks of the baselines is the state value, $V(s)$, defined as :
    \begin{equation}
        V(s) = \text{max}_a Q(s,a). \nonumber
    \end{equation}
    If we collect $V(s)$ in a reasonably large number of states and average it between states that share a specific trait, we can obtain a Monte Carlo estimate of the value assigned to that specific trait by the value network.
    
    Following a similar data collection process for our analysis in Section~\ref{GPLUtilAnalysis}, we measure $V(s)$ and contrast its average for states where there is at least an attacker in any defender's shooting range and states where there are no attackers in any defender's shooting range. By contrasting the green and blue line in Figure~\ref{fig:BaseVAnalysis}, we see that QL and QL-AM assign higher values to states where no attackers in any defender's shooting range. On the other hand, the blue and green line in plots associated to GNN and GNN-AM has similar values in most checkpoints. This demonstrates the baselines' inability to recognize the value of having an attacker in a defender's shooting range or, even worse, their preference towards states where there are no attackers in any defender's shooting range. Unlike the baselines, GPL learners do learn to assign higher utility values when an attacker is inside a defender's shooting range.
    
    We have then showed that the single-agent RL baselines failed to learn the effects of other agents' action towards the learner as GPL. With GPL, learning these concepts became an important path to enable improved performance in FortAttack. The baselines' inability to learn these concepts eventually lead them to produce subpar performance compared to GPL.
    





\end{document}